\PassOptionsToPackage{dvipsnames}{xcolor}
\documentclass[journal]{IEEEtran}

\usepackage{etoolbox}

\usepackage{xr}
\usepackage[%
style=numeric-comp,%
sortcites=true,%
sorting=none,%
url=false,%
doi=false,%
eprint=false,%
isbn=false,%
]{biblatex}
\usepackage{graphicx}
\graphicspath{{./img/}}
\usepackage[labelformat=simple]{subcaption}
\newcommand{\boximg}[2]{\parbox[c]{#1\subfigsz}{\includegraphics[width=#1\subfigsz]{#2}}}

\usepackage{tikz}
\usepackage{pgfplots}
\usepackage{pgfplotstable}
\pgfplotsset{compat=1.16,
  grid style=dashed,
  ymajorgrids=true,
}

\newlength{\subfigsz}

\usetikzlibrary{
  calc,
  shapes,
  arrows,
  shadows,
  positioning,
  decorations.shapes,
  decorations.markings,
  fit,
  bayesnet,
  matrix,
}

\usepgfplotslibrary{
  groupplots,
  colorbrewer,
  statistics,
  fillbetween,
}

\usepackage{booktabs}           %
\usepackage{multirow}
\usepackage{arydshln}           %

\usepackage{siunitx}
\robustify\bf

\makeatletter
\def\adl@drawiv#1#2#3{%
  \hskip.5\tabcolsep
  \xleaders#3{#2.5\@tempdimb #1{1}#2.5\@tempdimb}%
  #2\z@ plus1fil minus1fil\relax
  \hskip.5\tabcolsep}
\newcommand{\cdashlinelr}[1]{%
  \noalign{\vskip\aboverulesep
    \global\let\@dashdrawstore\adl@draw
    \global\let\adl@draw\adl@drawiv}
  \cdashline{#1}
  \noalign{\global\let\adl@draw\@dashdrawstore
    \vskip\belowrulesep}}
\makeatother

\usepackage{pdflscape}

\usepackage{amsmath}
\usepackage{mathtools}
\usepackage{commath}
\usepackage{amsfonts}           %
\usepackage{nicefrac}           %
\usepackage{microtype}          %

\DeclarePairedDelimiterX{\infdivx}[2]{(}{)}{%
  #1\;\delimsize\|\;#2%
}
\newcommand{\kl}{\operatorname{KL}\infdivx}
\newcommand{\skl}{\operatorname{SKL}\infdivx}

\let\given\givenbase

\DeclareMathOperator*{\argmax}{arg\,max}

\makeatletter
\DeclareRobustCommand\bigop[1]{%
  \mathop{\vphantom{\sum}\mathpalette\bigop@{#1}}\slimits@
}
\newcommand{\bigop@}[2]{%
  \vcenter{%
    \sbox\z@{$#1\sum$}%
    \hbox{\resizebox{\ifx#1\displaystyle.9\fi\dimexpr\ht\z@+\dp\z@}{!}{$\m@th#2$}}%
  }%
}
\makeatother

\DeclareMathOperator*{\E}{\mathbb{E}}

\let\oldtimes\times
\def\times{{\mkern1mu\oldtimes\mkern1mu}}

\makeatletter
\DeclareRobustCommand{\rvdots}{%
  \vbox{
    \baselineskip4\p@\lineskiplimit\z@
    \kern-\p@
    \hbox{.}\hbox{.}\hbox{.}
}}
\makeatother

\usepackage{algorithm}
\usepackage{algorithmic}

\PassOptionsToPackage{hyphens}{url}
\usepackage{hyperref}
\hypersetup{
  colorlinks,
  linkcolor = BrickRed,
  citecolor = NavyBlue,
  urlcolor  = WildStrawberry,
}
\usepackage{url}

\usepackage{xspace}
\makeatletter
\DeclareRobustCommand\onedot{\futurelet\@let@token\@onedot}
\def\@onedot{\ifx\@let@token.\else.\null\fi\xspace}

\def\eg{{e.g}\onedot} \def\Eg{{E.g}\onedot}
\def\ie{{i.e}\onedot} 
\def\cf{{cf}\onedot} 
 \def\vs{{vs}\onedot}
\def\wrt{w.r.t\onedot} 
  \def\aka{a.k.a\onedot}

\makeatother

\begin{document}

\title{Video Reenactment as Inductive Bias for Content-Motion Disentanglement}

\author{J.~F. Hernández~Albarracín,
        and A. Ramírez~Rivera,~\IEEEmembership{Senior Member,~IEEE}%
\thanks{%
Juan F. Hernández Albarracín is with Institute of Computing, University of Campinas, SP, Brazil, e-mail \texttt{juan.albarracin@ic.unicamp.br}.  Adín Ramírez Rivera is with Department of Informatics, University of Oslo, Norway, e-mail \texttt{adinr@uio.no}; and part of this work was done at the Institute of Computing, University of Campinas.}%
\thanks{Juan F. Hernández Albarracín was funded by the São Paulo Research Foundation (FAPESP) under grant No.~2017/16144-2.  A. Ramírez Rivera was funded by the Brazilian National Council for Scientific and Technological Development (CNPq) under grant No.~307425/2017-7; and in part by FAPESP under grant No.~2019/07257-3. Juan F. Hernández Albarracín and A. Ramírez Rivera were funded by the Coordena\c{c}\~ao de Aperfei\c{c}oamento de Pessoal de N\'ivel Superior---Brasil (CAPES)---Finance Code 001.}%
\thanks{The source code is available at \url{https://gitlab.com/mipl/mtc-vae}.}
\thanks{Pre-print to appear in IEEE Trans.\ on Image Processing.}
\thanks{Digital Object Identifier \href{https://doi.org/10.1109/TIP.2022.3153140}{10.1109/TIP.2022.3153140}.}
}

\markboth{IEEE Transactions on Image Processing}%
{Hern\'andez~Albarrac\'{i}n \& Ram\'{i}rez~Rivera: Video Reenactment as Inductive Bias for Content-Motion Disentanglement}

\maketitle

\begin{abstract}
Independent components within low-dimensional representations are essential inputs in several downstream tasks, and provide explanations over the observed data.
Video-based disentangled factors of variation provide low-dimensional representations that can be identified and used to feed task-specific models.
We introduce MTC-VAE, a self-supervised motion-transfer VAE model to disentangle motion and content from videos.
Unlike previous work on video content-motion disentanglement, we adopt a chunk-wise modeling approach and take advantage of the motion information contained in spatiotemporal neighborhoods.
Our model yields independent per-chunk representations that preserve temporal consistency.
Hence, we reconstruct whole videos in a single forward-pass.
We extend the ELBO's log-likelihood term and include a Blind Reenactment Loss as an inductive bias to leverage motion disentanglement, under the assumption that swapping motion features yields reenactment between two videos.
We evaluate our model with recently-proposed disentanglement metrics and show that it outperforms a variety of methods for video motion-content disentanglement.
Experiments on video reenactment show the effectiveness of our disentanglement in the input space where our model outperforms the baselines in reconstruction quality and motion alignment.
\end{abstract}


\IEEEpeerreviewmaketitle

\section{Introduction}
\label{sec:introduction}

\IEEEPARstart{W}{hile} the goal of representation learning is to obtain low-dimensional vectors useful for a diverse set of tasks, Disentangled Representation Learning (DRL) captures independent factors of variation within the observed data.
These disentangled representations are robust and interpretable, simplify several downstream tasks like classification and Visual Question Answering~\cite{Locatello2020aaai}, and support diverse content generation tasks~\cite{Chen2020tip,Ramesh2021}.
DRL shifted from unsupervised to weakly- and self-supervised methods, as inductive biases have shown to be fundamental in Deep Generative Models (DGM)~\cite{Locatello2019, Shu2020}.
DRL methods from video separate \emph{time independent} (\aka content) from \emph{dependent} (\aka motion) factors of variation.
While content features must be forced to have a low variance throughout the sequence, motion ones are expected to change.

Disentangling information from videos is of major importance since it can ease tasks that depend on the spatiotemporal data.
For instance, prediction tasks could rely on the independent representations of the objects or only on their temporal information.
These independence could not only ease the load on the downstream tasks but also enforce fairness and privacy over the data.
DRL from videos has been approached as a sequential learning process forcing temporal consistency among frames.
This problem is commonly addressed with Recurrent Neural Networks (RNN), due to their capacity of modeling temporal data of variable length.
Although architectures based exclusively on 3D Convolutional Neural Networks (3D-CNN) have been used in general representation learning from videos for downstream tasks~\cite{Carreira2017, Feichtenhofer2019}, few works rely only on convolutional architectures for DRL and posterior video generation~\cite{Wang2020, Aich2020}, despite their capacity of modeling whole videos, as they are constrained to fixed-length sequences.

Taking into account the great suitability of Variational Autoencoders (VAE) for unsupervised tasks~\cite{Su2020,Shi2018},
we propose a self-supervised DRL model that takes advantage of local spatio-temporal regularity to reconstruct videos by disentangling their content and motion while learning a robust representation space.
Motion-Transfer Chunk Variational Autoencoder (MTC-VAE) is a Variational Autoencoder that models temporal segments (\aka chunks) as independent random variables, maps them into a disentangled latent distribution, and maps them back consistently.
When modeling chunks as independent, the reconstructed videos may not be temporally consistent.
Hence, we preserve the temporal dependency that naturally exists among the chunks by assuming a Markovian relation between consecutive chunks at inference time.
To enforce it, we incorporate two inductive biases in our model:
(i)~We assume content features as stationary and motion ones as non-stationary in our model's log-likelihood.
(ii)~Video Reenactment (VR) is equivalent to swapping the motion representation of two videos and mapping them to the input space.
We show that this duality (independence at generation time, and dependence at inference time) is successful at representing video sequences for both disentanglement and reconstruction.

Our contributions are: (i)~A self-supervised DGM for VR and content-motion disentanglement from arbitrary-length videos through a simple 3D-CNN architecture in a single forward pass, improving over existing methods.
(ii)~Even assuming chunk independence, we significantly ease the disentangled motion-content feature inference and consistent video reconstruction, due to our inductive biases, and the self-supervised representation learning scheme.
(iii)~We show, that chunk-wise is better suited for DRL and video synthesis than frame-wise modeling for long videos.
Moreover, we highlight that, unlike SotA VR models, MTC-VAE is suited to learn disentangled low-dimensional representations.
VR models rely on entangled high-dimensional features and bypass information through the architecture to achieve better reconstruction at the cost of bloated features.
In contrast, our objective is to obtain independent factors of variation that are expressive enough for simple generators to create natural videos.

\section{Related Work}
\label{sec:soa}

\subsection{General Disentangled Representation Learning}
Seminal works on DRL are mostly unsupervised, and the majority rely on VAEs.
InfoGAN~\cite{Chen2016}, however, is the most relevant exception.
It uses control variables (categorical, discrete, or continuous) in the latent representation as inductive biases while penalizing mutual information among the latent units in an adversarial framework.
$\beta$-VAE~\cite{Higgins2017} includes the $\beta$ hyper-parameter into the VAE's ELBO to leverage independence among the latent scalars, leading to a higher-quality disentanglement.
Later approaches (\eg, $\beta$-TCVAE~\cite{Chen2018dr} and FactorVAE~\cite{Kim2018}) penalize Total Correlation among the latent scalars, yielding a better trade-off between disentanglement and reconstruction quality.
The ground-breaking work by \textcite{Locatello2019} showed that unsupervised methods for DRL are extremely weak.
Posterior works have shifted to weakly- and self-supervised approaches.
Hence, our proposed MTC-VAE introduces inductive biases in the latent space, such as explicit latent factors to represent content and motion features, with sufficient encoded information to guarantee VR from them.

\subsection{Disentangled Representations from Video}
\label{sec:soa_vdrl}
These works focus on disentangling time-dependent from time-independent features for each frame of the video and then enforcing inter-frame consistency.
Common setups of these approaches perform pose-content disentanglement while achieving consistency using RNNs and GANs~\cite{Denton2017, Villegas2017, Hsieh2018, Ge2018}.
Instead of pose-content disentanglement, some works separate deterministic from stochastic features~\cite{Denton2018, Lee2019savp}.
Most of the works in this area are applied to video prediction, but recent ones have started to be tested on VR tasks~\cite{Li2018,Wang2020,Aich2020,Zhu2020}.
Few of them \cite{Wang2020, Aich2020} rely on 3D-convolutional generators, but are constrained to fixed-length videos.
The rest use RNNs to capture the temporal relation between frames or segments at generation time, to perform either video reconstruction, prediction, or sequence-to-sequence translation.
Although MTC-VAE models dependent chunks at inference time, it assumes independence at generation time.
These assumptions simplify the tasks of reconstruction and VR since, to reconstruct a chunk of a video, it does not need to reconstruct the previous ones.
Therefore, the chunkwise approach takes the best of both worlds at not being constrained either to fixed-length-sequences or sequential generation.

\subsection{Video Reenactment}
\label{sec:soa_vr}
Recent methods on VR work in the domain of human faces~\cite{Zakharov2019, Nirkin2019, Chen2018, Zhou2019aaai}, human poses~\cite{Chan2019, Zhou2019, Liu2019tg, Yang2020}, or objects in general~\cite{Bansal2018, Siarohin2019, Siarohin2021, Zhao2018, Xie2020}.
Their main objective is to generate realistic videos, while the representation is either irrelevant or a secondary objective.
Instead, DRL models hold this objective as primary.
Most of these methods rely on warping techniques assisted by spatial transformer networks~\cite{Jaderberg2015} for frame-wise conditional video generation.
To apply such transformations, the generator requires high-dimensional spatial information that would normally be lost in a low-dimensional latent representation.
Hence, they either map to latent spaces that are larger than the original input space, to preserve spatial information, or bypass this information through skip connections from the encoder to the decoder.
Thus, a low-dimensional latent representation is not enough to represent the whole video.
In contrast, our proposal reconstructs videos while learning low-dimensional and factorized representations.
We highlight that our method reconstructs videos exclusively from low-dimensional representations.
Due to this restriction, we expect the perceptual quality and motion complexity of rendered videos to be higher in VR methods in comparison to DRL ones.
Despite this limitation, we consider our work as a step towards bridging these two areas.

\section{Proposed Approach: MTC-VAE}
\label{sec:method}

Given that content changes at a much slower rate than motion in a video, we propose to extract disentangled representations from local spatiotemporal neighborhoods (\aka chunks).
Content information of neighboring chunks changes so slowly that we may assume that it remains constant throughout a scene, while motion presents rapid changes.
Unlike existing frame-wise approaches, we use chunks to better capture the temporal characteristics of the video (\cf Section~\ref{sec:ablation} for the impact of the temporal windows), and their relations to obtain a self-supervised learning signal.

MTC-VAE contains only 3D-convolutional streams and, unlike recurrent approaches, models chunks as independent random variables for the generative pass, yet Markovian-dependent for the inference one.
Our formulation starts diverging from a standard two-latent-priors VAE when we extend our $\log p(x)$ to leverage inter-chunk consistency, which helps to reconstruct realistic videos, even though chunks are independently generated.
We go further and introduce the self-supervised \textit{blind reenactment loss} (BRL): another inductive bias that blindly simulates VR between two videos.

\subsection{Chunk-wise Video Modeling}
\label{sec:modeling}

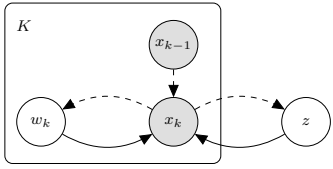
\begin{figure}%
  \centering%
  \resizebox{0.5\linewidth}{!}{
\begin{tikzpicture}[
  latent/.append style={
    minimum width=25pt,
    font=\footnotesize
  },
  plate caption/.append style={below right=0pt and 0pt of #1.north west},
]
\node[obs] (xk) {$x_k$};
\node[obs, above=.5cm of xk] (xkp) {$x_{k-1}$};
\node[latent, right=1.5cm of xk] (z) {$z$};
\node[latent, left=1.5cm of xk] (wk) {$w_k$};

\plate [inner sep=.3cm, xshift=0.1cm] {xw} {(wk)(xk)(xkp)} {$K$};

\path (wk) edge[bend right, ->] (xk);
\path (z) edge[bend left, ->] (xk);

\path (xk) edge[dashed, bend right, ->]  (wk);
\path (xk) edge[dashed, bend left, ->]  (z);
\path (xkp) edge[dashed, ->]  (xk);

\end{tikzpicture}}
  \caption{%
  In the generative model (solid arrows), $K$ chunks $\{x_{k}\}$ (observed) share the same content~$z$, while having their own motion~$w_k$.
During inference (dashed arrows), the latent variables $z$ and~$w_k$ are inferred from each chunk, while each chunk~$x_k$ also depends on the previous one.
  }
  \label{fig:graphical_model}%
\end{figure}

We represent the video $x = (x_{k})_{k=1}^K$ as a sequence of $K$ non-overlapping and equally-sized chunks~$x_{k}$ of length $c$.%
\footnote{%
  For brevity, we assume that $c$ divides the length of the video.
  However, we can model arbitrary-length videos by padding incomplete chunks to match $c$.}
Similarly, we define $w = \left(w_k\right)_{k=1}^K$ as the sequence of motion representations of each~$x_k$.
For the $k$-th chunk, we model the content and motion as independent latent variables $z$ and $w_k$, respectively.
We assume $z$ to be unique and shared across the chunks, as content remains constant through time.
Fig.~\ref{fig:graphical_model} depicts the graphical model for a video~$x$.

Different from common frame-wise approaches, where~$w$ normally depend on previous frames, in the generative phase, we model all the motion representations $\{w_k\}$ as independent random variables.
This assumption simplifies the generation process since it lets us generate a particular chunk without having to consider the previous ones in the video.
A unique $z$ for all the chunks sets an implicit dependence of each chunk to the whole video in the inference phase of the model.

\begin{figure}[tb]%
  \centering%
  \resizebox{\linewidth}{!}{\input{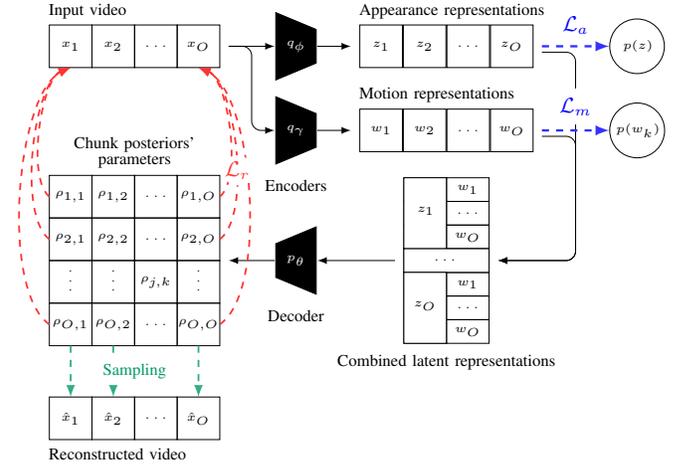}}
 \caption{%
  We feed consecutive chunks $\{x_k\}_{k=1}^O$ to the encoders $q_\phi$ and $q_\gamma$, yielding their representations, $\{w_k\}_{k=1}^O$ and $\{z_j\}_{j=1}^O$.
  We concatenate all combinations of $z_j$'s and $w_k$'s, and decode them to obtain the p.d.f.\ parameters $\rho_{j,k}$ for the $k$-th chunk posteriors $p_\theta(x_k \given w_k, z_j)$.
  Every posterior from $w_k$ must generate $x_k$.
  We maximize the log-likelihood of each chunk under the corresponding set of posteriors.
  Chunk posteriors \textcolor{red}{relate} with the original chunks through $\mathcal{L}_r$.
  The latent prior distributions \textcolor{blue}{relate} through $\mathcal{L}_a$ and $\mathcal{L}_m$.
  We \textcolor{blue!40!green}{sample} from the chunk posterior by applying the Sigmoid function to the output of the decoder.
  }
  \label{fig:logp}%
\end{figure}

Being the chunks independent, the joint probability of the model is the product of the conditionals of each chunk and their latent variables, \ie,
\begin{equation}
p(x, w, z) = p(z)\prod_k p(x_{k} \given w_k, z)p(w_k).
\end{equation}
We model the generative process of a single chunk through a VAE~\cite{Kingma2013}, with content encoder $q_\phi(z \given x_{k})$, motion encoder $q_\gamma(w_k \given x_{k})$, and decoder $p_\theta(x_{k} \given w_k, z)$ with parameters ($\phi$, $\gamma$, $\theta$), updated to maximize of the evidence lower bound (ELBO) of the expected log-likelihood
\begin{align}
\argmax_{\phi, \gamma, \theta} \E_{\tilde{q}(x_{1:k})} \sum_k \Big\lbrace & \E_{q_\phi}\E_{q_\gamma} \left[\log p_\theta(x_k \given w_k, z) \right] \nonumber \\
& - \kl{q_\gamma(w_k \given x_k)}{p(w_k)} \nonumber \\
& - \kl{q_\phi(z \given x_k)}{p(z)} \Big\rbrace.
\label{eq:elbo_whole}
\end{align}
Fig.~\ref{fig:logp} shows the pipeline to calculate the ELBO~(\ref{eq:elbo_whole}).
We maximize the expected reconstruction loss over the two latent variables \wrt their distributions $q_\phi(z \given x_{k})$ and $q_\gamma(w_k \given x_{k})$ (first term), and minimize the Kullback-Leibler divergence between these distributions \wrt their priors.
We compute their expected value \wrt the empirical distribution of the chunks $\tilde{q}(x_{1:k}) = \prod_k q(x_{k} \given x_{k-1})$ that models a Markovian temporal relation between them.\footnote{We assume the first chunk to be distributed through $q(x_1 \given x_0) \equiv q(x_1)$ to simplify the notation.}
We approximate the chunk distribution through a sampling process on the videos, and model all prior distributions as standard Gaussians.
To generate a new video from the chunk posterior, we concatenate the expected values of the chunk posteriors, directly provided by the decoder.
See Appendix~\ref{sec:elbo} for further detail and proof of our formulation.

Our architecture consists of two encoders $q_\phi(z \given x_{k})$ and $q_\gamma(w_k \given x_{k})$, and one decoder $p_\theta(x_{k})$\@.
All of them have five 3D-convolutional layers, with Batchnorm and ReLU activations.
The number of filters in the hidden layers of the decoder is double the number of filters in the encoders.

\subsection{Inter-Chunk Consistency}
\label{sec:inter-chunk-consistency}

As shown in Equation~\ref{eq:elbo_whole}, we can train a VAE to independently reconstruct chunks.
However, the independence assumption at generation time may cause the videos to not be smoothly rendered between chunks.
To solve this issue, we force our model to yield a unique content representation~$z$, regardless of the chunk from which it is inferred.

We part from the assumption that content is constant throughout the video, and so does its latent representation $z \sim q_\phi(z \given x_{k})$---\cf Section~\ref{sec:modeling}.
To force our model to learn this constraint, we train it to maximize $\log p_\theta(x_{k} \given w_k,z_j)$ for every~$j$, \ie, maximize the log-likelihood of a chunk $x_{k}$ given its own motion~$w_k$ and any~$z_j$ content representation---\cf Fig.~\ref{fig:logp}.
We extend the $\log p(x_{k} \given  w_k, z)$ term~(\ref{eq:elbo_whole}) to fulfill this constraint.
So our final reconstruction loss is
\begin{equation}
\label{eq:logp_x}
\mathcal{L}_{r}(\theta, \phi,\gamma) = \sum_{k=1}^O\sum_{j=1}^O \left[\log p_\theta(x_{k} \given w_k,z_j)\right],
\end{equation}
where $z_j \sim q_\phi(z \given x_{j})$, $w_k \sim q_\gamma(w_k \given x_{k})$, and $O$ is defined as the \textit{order of the model} that restricts the number of chunks used to calculate the loss.
As Fig.~\ref{fig:logp} shows, the decoder outputs the distribution parameters $\rho_{j,k}$ of each chunk likelihood $p_\theta(x_{k} \given w_k,z_j)$, used in $\mathcal{L}_{r}$.
Due to its combinatory nature, it is impractical to apply $\mathcal{L}_r$ to all the chunks.
Hence, for each forward pass, we consider only a sequence of $O \le K$ consecutive chunks of $x$, starting at a random frame.

The second and third terms of the expected log-likelihood~(\ref{eq:elbo_whole}) correspond to the regularization terms of the motion and content distributions, respectively.
That is, we compute
\begin{align}
  \label{eq:lm}
  \mathcal{L}_{m}(\gamma)&= -\sum_{k=1}^O \kl{q_\gamma(w_k \given x_k)}{p(w_k)}, \text{ and} \\
  \label{eq:la}
  \mathcal{L}_{a}(\phi)&= -\sum_{k=1}^O \kl{q_\phi(z \given x_k)}{p(z)},
\end{align}
on $O$ consecutive chunks instead of the whole video---\cf Fig.~\ref{fig:logp}.

Unlike other variational inference methods of grouped observations~\cite{Locatello2020,Mathieu2016,Bouchacourt2018,Hosoya2019}, we opted for the extended log-probability term~(\ref{eq:logp_x}), considering different combinations of appearance features, to yield stronger gradients for chunk-consistency, instead of averaging the shared representations in the group.

\subsection{Blind Reenactment Loss}
\label{sec:brl}

Our proposed Blind Reenactment Loss (BRL) loss increases the likelihood $\log p(x_{k} \given w_k, z)$ of our ELBO given any encoded chunks.
It aims at leveraging content-motion disentanglement by doing VR between a source video~$S$ and a driving video~$D$.
The motion representation of $S$ is replaced by the one of $D$, to reconstruct a reenacted video with the object of interest from $S$ moving like the one in $D$.
This translation can be achieved uniquely if the content and motion representations of both videos are disentangled.
The main difficulty is that, in principle, we would need to train our model with ground-truth reenacted videos.
However, we opt for self-supervised training and take advantage of our chunk-based approach.

Consider two chunks $s_i$ and~$s_j$ from $S$, and one chunk $d_l$ from $D$.
Assuming constant content throughout the video, if we independently reenact $s_i$ and~$s_j$ \wrt $d_l$, the two reconstructed chunks must be the same since $s_i$ and $s_j$ have the same content.
To achieve this objective, we force the corresponding chunk posteriors $p(x_{k} \given w_k, z)$ to be equivalent, \ie, $p(x \given w_l^d, z_j^s) \equiv p(x \given w_l^d, z_i^s)$, where $z_i^s \sim q(z \given s_i)$, $z_j^s \sim q(z \given s_j)$, and $w_l^d \sim q(w_k \given d_l)$, by minimizing the KL divergence between every two posteriors that fit the described case.
Let
\begin{align}
  \label{eq:brl}
\mathcal{L}_b & (\theta,\phi,\gamma) = \nonumber \\
& -\sum_{l=1}^O\sum_{j=1}^O\sum_{i=1}^O \operatorname{SKL} \Big(p_\theta(x \given  w_l^d,z_j^s) \;\big\| p_\theta(x \given w_l^d,z_i^s) \Big).
\end{align}
be our BRL, where $\skl{P}{Q}= \frac{1}{2}(\kl{P}{Q} + \kl{Q}{P})$ is a symmetrical operator.
This loss involves two empirical distributions of unobservable samples, so we are not aware, at training time, of whether the sampled videos are correctly reenacted.
If there is disentanglement, posteriors sharing the same motion of $D$ and \textit{any} content of $S$ must be equivalent, regardless of their samples.

The BRL must be optimized along with $\mathcal{L}_r$~(\ref{eq:logp_x}) to prevent posterior collapse.
Notice that, if $O = 1$, then $j=i=1$ and $\mathcal{L}_b=0$, so this objective can only be optimized for $O \geq 2$.

\subsection{General Loss Function}

We define the general objective to be maximized as
\begin{equation}
\mathcal{L} = \mathcal{L}_r + \lambda\mathcal{L}_b + \beta(\mathcal{L}_a + \mathcal{L}_m),
\end{equation}
where $\beta$ comes from $\beta$-VAE by \cite{Higgins2017}, and $\lambda$ weights $\mathcal{L}_b$\@.
Each element in the batch is conformed by a sequence of $O$ chunks, so $\mathcal{L}$ can be calculated independently for every element.

\section{Experiments}
\label{sec:experiments}

\begin{table*}[tb]
  \caption{Performance for content-motion disentanglement and data realism. (*\,$c=1$)}
  \label{tab:baselines}
  \scriptsize
  \centering
  \resizebox{\linewidth}{!}{%
  \begin{tabular}{cSSSSS[table-format=3.2(4)]cSSSSS[table-format=3.2(4)]}
    \toprule
    &   {FVAE $\uparrow$} & {MIG $\uparrow$} & {SAP $\uparrow$} & {SSIM $\uparrow$} & {FID $\downarrow$}
    & & {FVAE $\uparrow$} & {MIG $\uparrow$} & {SAP $\uparrow$} & {SSIM $\uparrow$} & {FID $\downarrow$} \\
    \cmidrule{2-6} \cmidrule{8-12}
    & \multicolumn{5}{c}{3dShapes} & & \multicolumn{5}{c}{LPC} \\
    \cmidrule{2-6} \cmidrule{8-12}
    $\beta$-TCVAE & .50(2) & \bf .01(1) &     .11(8)  &     .53(10) &    140.25(5113) & &     .81(3) &     .02(02) &     .00(00) &     .64(1) &     80.27(317) \\
    dis-VAE       & .50(0) &     .00(0) &     .08(6)  &     .40(3)  & \bf 71.24(1235) & &     .92(2) &     .02(02) &     .01(01) &     .78(1) &     71.70(176) \\
    SVG-LP        & .50(0) & \bf .01(0) &     .03(02) &     .54(5)  &    136.00(8112) & &     .63(2) &     .00(00) &     .00(00) & \bf .79(2) &     62.75(987) \\
    MTC-VAE       & .50(2) & \bf .01(0) & \bf .41(14) &     .67(6)  &    119.47(5100) & & \bf .93(6) & \bf .11(11) & \bf .60(40) &     .67(1) & \bf 41.72(331) \\
    MTC-VAE*      & .50(1) & \bf .01(0) &     .39(11) & \bf .73(2)  &    100.80(4682) & &     .86(1) &     .00(00) &     .11(03) &     .67(1) &     42.59(409) \\
    \cmidrule{2-6} \cmidrule{8-12}
    & \multicolumn{5}{c}{CK+} & & \multicolumn{5}{c}{MMNIST} \\
    \cmidrule{2-6} \cmidrule{8-12}
    $\beta$-TCVAE &     .79(5) & \bf .03(2) &     .06(4)  &     .50(7)  &    116.74(2470) & &     .66(7) &     .04(4) &     .04(4)  & \bf .71(3) &     152.56(1693) \\
    dis-VAE       &     .71(2) &     .01(1) &     .04(2)  &     .61(5)  &     71.48(309)  & &     .64(4) &     .02(2) &     .03(2)  &     .70(2) &     149.43(973) \\
    SVG-LP        &     .70(6) &     .02(1) &     .04(2)  &     .02(0)  & \bf 38.79(1763) & &     .52(1) &     .01(0) &     .02(2)  &     .58(2) &     179.08(5049) \\
    MTC-VAE       & \bf .86(4) &     .02(1) & \bf .13(5)  &     .66(12) &     63.13(2250) & & \bf .95(4) & \bf .11(7) & \bf .10(5)  &     .68(1) & \bf 102.11(99) \\
    MTC-VAE*      &     .85(2) & \bf .03(1) &     .05(2)  & \bf .68(13) &     76.16(1937) & &     .91(4) &     .09(5) &     .09(4)  &     .69(1) &     186.25(2355) \\
    \cmidrule{2-6} \cmidrule{8-12}
    & \multicolumn{5}{c}{dSprites} & & \multicolumn{5}{c}{MUG} \\
    \cmidrule{2-6} \cmidrule{8-12}
    $\beta$-TCVAE &     .57(3) &     .00(0) &     .04(1) & \bf .79(3) &     79.34(618) & &     .74(5) & \bf .05(4) &     .23(03) &     .51(01) &     44.78(332) \\
    dis-VAE       &     .70(2) &     .01(0) &     .01(0) & \bf .79(0) &     97.07(173) & & \bf .76(3) &     .01(1) &     .11(03) & \bf .78(01) &     62.84(345) \\
    SVG-LP        &     .61(5) &     .00(0) &     .00(0) & \bf .79(2) &     98.14(920) & &     .64(4) &     .02(1) &     .38(04) &     .50(15) &    101.59(3700) \\
    MTC-VAE       &     .91(2) & \bf .04(1) & \bf .10(1) &     .78(0) & \bf 57.18(643) & &     .72(4) &     .01(1) &     .73(05) &     .63(02) & \bf 28.79(115) \\
    MTC-VAE*      & \bf .92(1) &     .02(2) &     .01(0) &     .77(1) &    105.79(586) & &     .70(9) &     .04(2) & \bf .76(10) &     .66(06) &     43.86(1315) \\
    \bottomrule
  \end{tabular}}

\end{table*}

We evaluated MTC-VAE in DRL, VR, and downstream tasks.
Although MTC-VAE does not require labels in training time, we used labels to asses disentanglement, and to split the training and testing datasets.
We detail the implementation of the model and the experimental setup in Appendix~\ref{sec:implementation}.

\textbf{Datasets.}
(i)~Cohn-Kanade (CK+) facial dataset~\cite{Kanade2000, Lucey2010}, (ii)~Liberated Pixel Cup (LPC) sprites, (iii)~Moving MNIST (MMNIST)~\cite{Srivastava2015}, (iv)~Deepmind's dSprites, (v)~Deepmind's 3dShapes, and (vi)~Multimedia Understanding Group (MUG) facial dataset~\cite{Aifanti2010}.
We generated videos from the images of dSprites and 3dShapes, forming sequences of objects moving in linear and curved trajectories, or changing their perspective.
Each dataset contains \num{10000} videos, except for CK+ (\num{320}), LPC (\num{200000}), and MUG (\num{700}).
We report the average model performance in a $5$-fold cross-validation setup ($80\%$ for training and $20\%$ for testing).
Appendix~\ref{sec:data} provides further detail about the datasets, as well as the factors of variation.

\textbf{Baselines.}
We compared our method against the Disentangled Sequential Autoencoder (dis-VAE) \cite{Li2018}, SVG-LP \cite{Denton2018}, and $\beta$-TCVAE \cite{Chen2018dr}.
The first two are frame-wise approaches that disentangle time-dependent from time-independent factors.
Although SVG-LP namely disentangles deterministic from stochastic features, they force the deterministic features to remain constant, while the stochastic ones change from frame to frame, like a content-motion modeling.
$\beta$-TCVAE is an unsupervised disentanglement model, tested so far on images, so we extended it to 3D-CNNs to support chunks.

\textbf{Hyper-parameters.}
After a hyper-parameter search in the models (see details in Appendix~\ref{sec:implementation}), we tuned the $\beta$ parameter and the latent space size.
For dSprites, LPC and MMNIST, $\beta = 1$, and $\beta = 5$ for the other datasets.
Regarding the latent space dimensionality (where each dimension is a \textit{latent unit}), $\text{dim}(z)=14$, $\text{dim}(w_k)=7$ for CK+, LPC, and MUG, $\text{dim}(z)=12$, $\text{dim}(w_k)=6$ for 3dShapes, $\text{dim}(z)=12$, $\text{dim}(w_k)=4$ for dSprites, and $\text{dim}(z)=8$, $\text{dim}(w_k)=4$ for MMNIST.
We performed ablation studies on $\lambda$, $c$, $O$, and $\beta$ (\cf Section~\ref{sec:ablation} and Appendix~\ref{sec:detailed_results}).

\subsection{Content-Motion Disentanglement}
\label{sec:experiments_disentanglement}

We obtained the latent representations from the trained models for the test set and, using ground-truth labels, we calculated the Mutual Information Gap (MIG)~\cite{Chen2018dr}, the FactorVAE (FVAE) disentanglement metric~\cite{Kim2018}, and the Separated Attribute Predictability Score (SAP)~\cite{Kumar2018}.

Assessing disentanglement quality is narrowly application-related~\cite{Eastwood2018,Ridgeway2018}.
We adhere to the criteria defined by \textcite{Ridgeway2018}, by which we may evaluate disentanglement based on either \emph{modularity} (\ie, each unit contains information of at most one factor), \emph{compactness} (\ie, each factor is ideally encoded by at most one unit) or \emph{explicitness} (\ie, each factor is easily recovered from its code).

Since our objective is to encode two factors of variation (content and motion) in various latent units, our main interest is modularity.
Compactness, although desirable, is expected to not be fulfilled, as content and motion are complex factors that can barely be represented in few latent units.
Explicitness is important to estimate the effectiveness of disentangled representations for downstream tasks, like classification.

MIG and SAP heavily penalize representations that are not compact, by depending on the mean difference between the first and second most predictive/informative units.
Hence, FVAE is the metric that interests us the most, as it measures both modularity and explicitness.
We report results on MIG and SAP for completeness since, besides assessing compactness, to some extent, MIG also assesses modularity, and SAP, explicitness.

For $\beta$-TCVAE and MTC-VAE, we split every test video into chunks and calculated one latent vector per chunk.
For dis-VAE and SVG-LP, we obtained one vector per frame.
We aggregated the multiple factors, provided in 3dShapes, dSprites, and LPC, into two categories: time-dependent and time-independent, yielding two factors, to reduce the risk of over-estimation of disentanglement performance, due to pairs of disentangled factors while the rest are entangled.

Table~\ref{tab:baselines} shows the performance of the models on the content-motion disentanglement.
We included the results obtained for the frame-wise version of MTC-VAE (\ie, $c=1$) to compare against dis-VAE and SVG-LP\@.
Both the chunk and frame versions of MTC-VAE are the ones with the best disentanglement performance, followed by $\beta$-TCVAE\@ and dis-VAE.
It is remarkable that SVG-LP uses skip connections from the encoder to the decoder, so most of the appearance information is not in the latent representation.
This is reflected in the fact that it attained the poorest performance.
In general, the chunk version of MTC-VAE outperforms the frame version.

\begin{table}[tb]
\caption{Multi-factor disentanglement. (*\,$c=1$)}
\label{tab:mf}
\scriptsize
\centering
\begin{tabular}{crSSS}
\toprule
 & & {FVAE $\uparrow$} & {MIG $\uparrow$} & {SAP $\uparrow$} \\
\midrule
\multirow{5}{*}{3dShapes}
& $\beta$-TCVAE &     .21(3) &     .07(4) &     .03(2) \\
& dis-VAE       &     .19(1) &     .03(1) &     .01(1) \\
& SVG-LP        &     .18(1) &     .02(1) &     .01(0) \\
& MTC-VAE       &     .27(5) & \bf .19(7) & \bf .08(3) \\
& MTC-VAE*      & \bf .31(2) &     .14(5) &     .05(2) \\
\midrule
\multirow{5}{*}{dSprites}
& $\beta$-TCVAE &     .28(1) &     .02(1) &     .01(0) \\
& dis-VAE       &     .28(0) &     .02(1) &     .01(0) \\
& SVG-LP        &     .29(0) &     .00(0) &     .00(0) \\
& MTC-VAE       & \bf .33(2) & \bf .11(1) & \bf .02(0) \\
& MTC-VAE*      &     .29(1) &     .07(2) &     .01(0) \\
\midrule
\multirow{5}{*}{LPC}
& $\beta$-TCVAE &     .32(7) &     .16(9) &     .03(1) \\
& dis-VAE       &     .22(1) &     .04(1) &     .06(5) \\
& SVG-LP        &     .17(0) &     .01(0) &     .01(1) \\
& MTC-VAE       &     .41(7) & \bf .21(5) & \bf .89(1) \\
& MTC-VAE*      & \bf .43(6) &     .18(3) & \bf .89(0) \\
\bottomrule
\end{tabular}
\end{table}

\begin{figure*}[tb]
\centering
  \scriptsize
  \setlength{\subfigsz}{.32\linewidth}
  \setlength\tabcolsep{1.5pt}
  \begin{tabular}{cccccc}
    Source & Driving & Source & Driving & Source & Driving \\
    \boximg{.15}{MUG_s.png} &
    \boximg{.75}{MUG_d.png} &
    \boximg{.15}{LPC_s.png} &
    \boximg{.9}{LPC_d.png} &
    \boximg{.15}{dSprites_s.png} &
    \boximg{.75}{dSprites_d.png} \\
    $\beta$-TCVAE & \boximg{.75}{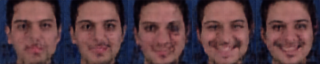} &
    $\beta$-TCVAE & \boximg{.9}{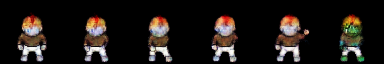} &
    $\beta$-TCVAE & \boximg{.75}{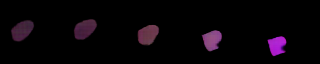} \\
    dis-VAE & \boximg{.75}{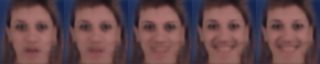} &
    dis-VAE & \boximg{.9}{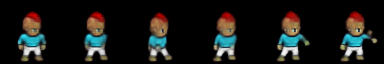} &
    dis-VAE & \boximg{.75}{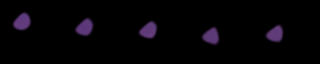} \\
    SVG-LP  & \boximg{.75}{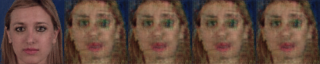} &
    SVG-LP  & \boximg{.9}{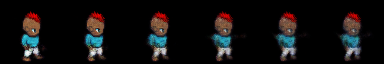} &
    SVG-LP  & \boximg{.75}{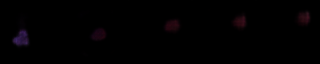} \\
    MTC-VAE & \boximg{.75}{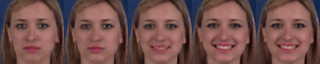} &
    MTC-VAE & \boximg{.9}{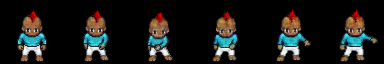} &
    MTC-VAE & \boximg{.75}{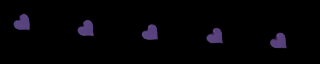}
  \end{tabular}
\caption{
  Reenactment results.  Each set shows the reenacted video of each method with the appearance of \textit{source} and the motion of \textit{driving}.
}
\label{fig:comparison}
\end{figure*}

Although MTC-VAE is trained for motion-content disentanglement, we can argue that this task can be used as a step towards multi-factor disentanglement.
To show our point, we calculated MIG, FVAE, and SAP considering all the factors of variation provided in the datasets' metadata.
Table~\ref{tab:mf} shows the results for 3dShapes, dSprites, and LPC since the others only provide motion-content labels.
In all cases, MTC-VAE (both frame and chunk versions) significantly outperforms the baselines.
The second best method was $\beta$-TCVAE, which is expected since it has been already tested on multi-factor disentanglement for images.
Table~\ref{tab:mf} demonstrates that multi-factor disentanglement is a significantly harder task, but it is remarkable that MTC-VAE features are more disentangled than the others, even when the model was not trained for this specific task.
We provide a list and a description of the factors of variation considered for each dataset in Appendix~\ref{sec:implementation}.

\subsection{Video Reenactment}
\label{sec:reenactment}

\begin{figure*}[tb]
\resizebox{\linewidth}{!}{
\pgfplotstableread{
c 3dShapesFID 3dShapesFVAE 3dShapesMIG 3dShapesSAP 3dShapesSSIM CKFID CKFVAE CKMIG CKSAP CKSSIM dSpritesFID dSpritesFVAE dSpritesMIG dSpritesSAP dSpritesSSIM LPCFID LPCFVAE LPCMIG LPCSAP LPCSSIM MMNISTFID MMNISTFVAE MMNISTMIG MMNISTSAP MMNISTSSIM MUGFID MUGFVAE MUGMIG MUGSAP MUGSSIM AvgFID AvgFVAE AvgMIG AvgSAP AvgSSIM StdFID StdFVAE StdMIG StdSAP StdSSIM
1 100.8 .5 .01 .39 .73 76.16 .85 .03 .05 .68 105.79 .92 .02 .01 .77 42.59 .86 .00 .11 .67 186.25 .91 .09 .09 .69 43.86 .7 .04 .76 .66 92.58 .79 .03 .24 .70 53.21 .16 .03 .29 .04 
3 114.23 .51 .01 .41 .73 70.56 .87 .03 .11 .67 70.41 .87 .02 .06 .78 41.72 .93 .11 .60 .67 96.04 .95 .08 .09 .68 37.62 .73 .03 .7 .62 71.76 .81 .05 .33 .69 29.88 .17 .04 .28 .06 
5 119.47 .5 .01 .41 .67 63.13 .86 .02 .13 .66 68.48 .82 .03 .06 .78 84.87 .87 .00 .17 .66 102.11 .95 .07 .1 .68 37.08 .73 .01 .69 .62 79.19 .79 .02 .26 .68 29.41 .16 .03 .24 .05 
7 117.88 .48 .01 .32 .62 63.03 .85 .02 .12 .62 71.17 .79 .02 .05 .78 76.92 .87 .00 .16 .67 95.69 .92 .06 .09 .68 37.36 .72 .01 .72 .62 77.01 .77 .02 .24 .67 27.64 .16 .02 .25 .06 
9 124.43 .51 .01 .33 .58 60.56 .81 .03 .11 .61 72.77 .76 .03 .10 .78 53.57 .89 .01 .19 .67 116.05 .91 .1 .12 .68 37.56 .72 .02 .7 .61 77.49 .77 .03 .26 .66 35.12 .15 .03 .23 .07 
}\datatable

\pgfplotsset{cycle list/Dark2-6}

\begin{tikzpicture}[
  curve/.style={
    mark=*, mark size=1pt, smooth, thick
  },
  avg/.style = {
    curve, 
    draw=black!80, 
    opacity=0.75,
  },
  std/.style={
    fill=black!80, 
    opacity=0.15,
  },
  y double precision/.style = {
    /pgfplots/y tick label style={
      /pgf/number format/fixed,
      /pgf/number format/fixed zerofill,
      /pgf/number format/precision=2,
    },
  },
]
\begin{groupplot}[
  group style={
    group size=5 by 1, 
    horizontal sep=1cm,
  }, 
  cycle list/Dark2-6,
  xtick={1, 3, 5, 7, 9}, 
  xmin=0, xmax=10, 
  height=6cm, width=6cm,
]

\nextgroupplot[
    title={FVAE $\uparrow$},
    ytick={0.4,0.5,...,1},
]

\foreach \c in {3dShapes, CK, dSprites, LPC, MMNIST, MUG} 
    \addplot+[curve] table[x=c,y=\c FVAE] from \datatable;

\addplot [avg] table[x=c,y=AvgFVAE] from \datatable;
\addplot [name path=upper, draw=none] table[x=c,y expr=\thisrow{AvgFVAE}+\thisrow{StdFVAE}] from \datatable;
\addplot [name path=lower, draw=none] table[x=c,y expr=\thisrow{AvgFVAE}-\thisrow{StdFVAE}] from \datatable;
\addplot [std] fill between[of=upper and lower];

\nextgroupplot[
    title={MIG $\uparrow$},
    y double precision,
    ytick={0, 0.03,..., 0.12}, 
]

\foreach \c in {3dShapes, CK, dSprites, LPC, MMNIST, MUG} 
    \addplot+[curve] table[x=c,y=\c MIG] from \datatable;

\addplot [avg] table[x=c,y=AvgMIG] from \datatable;
\addplot [name path=upper, draw=none] table[x=c,y expr=\thisrow{AvgMIG}+\thisrow{StdMIG}] from \datatable;
\addplot [name path=lower, draw=none] table[x=c,y expr=\thisrow{AvgMIG}-\thisrow{StdMIG}] from \datatable;
\addplot [std] fill between[of=upper and lower];

\nextgroupplot[
    title={SAP $\uparrow$},
    legend style={at={(.5,-.2)},anchor=center},
    legend style={
      legend columns=7,
      draw=gray,
    },
]

\legend{3dShapes, CK, dSprites, LPC, MMNIST, MUG, Avg.}

\foreach \c in {3dShapes, CK, dSprites, LPC, MMNIST, MUG} 
    \addplot+[curve] table[x=c,y=\c SAP] from \datatable;

\addplot [avg] table[x=c,y=AvgSAP] from \datatable;
\addplot [name path=upper, draw=none] table[x=c,y expr=\thisrow{AvgSAP}+\thisrow{StdSAP}] from \datatable;
\addplot [name path=lower, draw=none] table[x=c,y expr=\thisrow{AvgSAP}-\thisrow{StdSAP}] from \datatable;
\addplot [std] fill between[of=upper and lower];

\nextgroupplot[
    title={SSIM $\uparrow$},
    y double precision,
    ytick={0.6,.65,...,0.8},
]

\foreach \c in {3dShapes, CK, dSprites, LPC, MMNIST, MUG} 
  \addplot+[curve] table[x=c,y=\c SSIM] from \datatable;

\addplot [avg] table[x=c,y=AvgSSIM] from \datatable;
\addplot [name path=upper, draw=none] table[x=c,y expr=\thisrow{AvgSSIM}+\thisrow{StdSSIM}] from \datatable;
\addplot [name path=lower, draw=none] table[x=c,y expr=\thisrow{AvgSSIM}-\thisrow{StdSSIM}] from \datatable;
\addplot [std] fill between[of=upper and lower];

\nextgroupplot[
    title={FID $\downarrow$},
]

\foreach \c in {3dShapes, CK, dSprites, LPC, MMNIST, MUG} 
    \addplot+[curve] table[x=c,y=\c FID] from \datatable;

\addplot [avg] table[x=c,y=AvgFID] from \datatable;
\addplot [name path=upper, draw=none] table[x=c,y expr=\thisrow{AvgFID}+\thisrow{StdFID}] from \datatable;
\addplot [name path=lower, draw=none] table[x=c,y expr=\thisrow{AvgFID}-\thisrow{StdFID}] from \datatable;
\addplot [std] fill between[of=upper and lower];

\end{groupplot}
\end{tikzpicture}}
\caption{Study of the chunk size \vs several metrics. The gray area shows one standard deviation away from the average plot.}
\label{fig:cs}
\end{figure*}
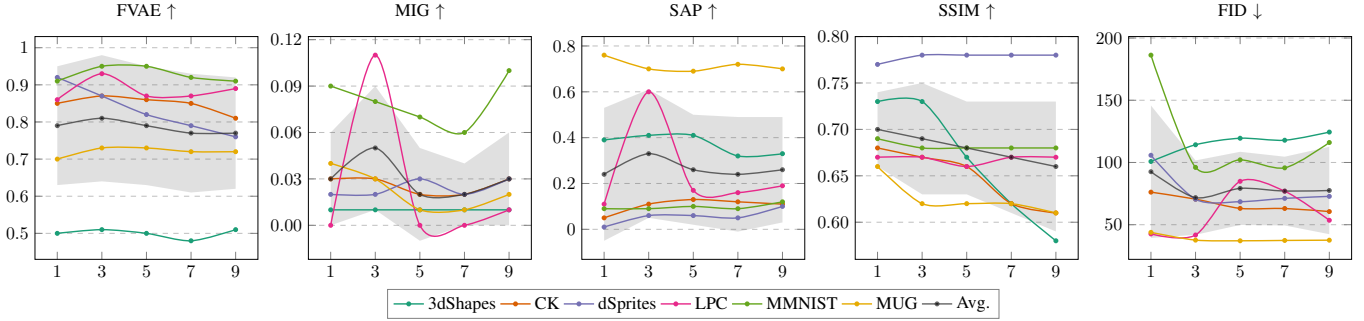

\begin{figure*}[tb]
  \resizebox{\linewidth}{!}{
\pgfplotstableread{
0.52 0.52 0.78 0.90 0.87 0.86 0.98 0.99 0.79 0.98 .68 .71 0.02 0.01 0.01 0.01 0.01 0.04 0.02 0.04 0.02 0.11 .03 .00 0.40 0.40 0.03 0.12 0.00 0.01 0.04 0.01 0.09 0.07 .66 .67 0.59 0.60 0.68 0.69 0.81 0.82 0.63 0.68 0.68 0.68 .62 .64 58.52 55.62 67.62 59.10 69.15 96.80 141.35 146.10 91.51 94.79 29.61 28.02
0.47 0.47 0.76 0.77 0.88 0.89 0.99 0.99 0.90 0.98 .67 .80 0.01 0.01 0.03 0.03 0.01 0.02 0.01 0.02 0.06 0.13 .00 .02 0.53 0.52 0.10 0.08 0.00 0.01 0.09 0.04 0.16 0.08 .72 .75 0.59 0.59 0.68 0.75 0.82 0.82 0.63 0.65 0.67 0.67 .63 .65 53.69 65.30 61.89 46.48 75.06 109.76 129.05 143.22 102.95 107.85 30.29 29.32
0.48 0.48 0.85 0.86 0.82 0.85 0.99 1.00 0.90 0.99 .74 .76 0.01 0.02 0.02 0.01 0.02 0.01 0.05 0.03 0.02 0.12 .00 .02 0.19 0.58 0.08 0.02 0.00 0.01 0.05 0.03 0.09 0.03 .71 .74 0.59 0.60 0.63 0.72 0.82 0.82 0.64 0.64 0.68 0.68 .65 .66 55.19 59.69 80.40 52.05 72.09 96.16 133.78 130.62 101.95 101.36 29.40 28.62
0.52 0.52 0.81 0.83 0.87 0.87 0.98 0.99 0.90 0.88 .67 .75 0.01 0.00 0.01 0.03 0.01 0.01 0.02 0.03 0.06 0.11 .00 .02 0.19 0.43 0.01 0.10 0.01 0.01 0.06 0.10 0.03 0.15 .79 .80 0.59 0.59 0.69 0.69 0.82 0.82 0.63 0.66 0.69 0.69 .61 .61 58.40 54.29 57.50 66.35 70.56 106.19 133.06 138.70 109.04 107.56 30.51 30.47
0.52 0.52 0.80 0.86 0.82 0.87 0.95 1.00 0.90 0.98 .71 .73 0.00 0.01 0.02 0.01 0.03 0.02 0.03 0.02 0.02 0.11 .00 .01 0.50 0.34 0.12 0.07 0.01 0.01 0.03 0.17 0.03 0.04 .75 .75 0.59 0.60 0.70 0.72 0.81 0.81 0.66 0.65 0.68 0.69 .61 .61 55.98 55.13 70.67 71.69 70.04 102.41 141.01 140.35 103.16 106.41 28.61 27.54
}\csvdata

\pgfplotsset{
  cycle list/Paired-12,
  cycle multiindex* list={
    mark=*\nextlist
    Paired-12\nextlist
  },
  /pgf/declare function={
    Floor(\x) = round(\x-0.49);
  },
}
\begin{tikzpicture}[
  y double precision/.style = {
    /pgfplots/y tick label style={
      /pgf/number format/fixed,
      /pgf/number format/fixed zerofill,
      /pgf/number format/precision=2,
    },
  },
]

\begin{groupplot}[
  group style={
    group size=5 by 1, 
    horizontal sep=1cm
  },
  width=6cm, height=6cm, 
  boxplot/draw direction = y,
  enlarge x limits=.01, 
  xtick style = {draw=none},
  xticklabels = {}, 
  xtick={1,2,3,4}, 
  boxplot={
    draw position={1/3 + Floor(\plotnumofactualtype/2) + 1/3*mod(\plotnumofactualtype,2)},
    %
    box extend=0.28,
  },
  every boxplot/.style={mark=*,every mark/.append style={mark size=1pt}},
]

  \nextgroupplot[title={FVAE $\uparrow$},
    legend to name={gl},
    legend entries = {\strut, 3dShapes, \strut, CK, \strut, dSprites, \strut, LPC, \strut, MMNIST, \strut, MUG},
    legend style={
      legend columns=12,
      draw=gray,
    },
    ytick={0.5,0.6,...,1},
  ]
  \foreach \n in {0,...,11} {
    \addplot+[boxplot, fill, draw=black] table[y index=\n] {\csvdata};
  }

  \nextgroupplot[title={MIG $\uparrow$}, y double precision, ytick={0, 0.02,..., 0.11}] 
  \foreach \n [count=\i from 0] in {12,...,23} {
    \addplot+[boxplot, fill, draw=black] table[y index=\n] {\csvdata};
  }

  \nextgroupplot[title={SAP $\uparrow$}, ytick={0,0.1,...,0.8}]
  \foreach \n [count=\i from 0] in {24,...,35} {
    \addplot+[boxplot, fill, draw=black] table[y index=\n] {\csvdata};
  }

  \nextgroupplot[title={SSIM $\uparrow$}, y double precision, ytick={0.6,0.65,...,0.85}]
  \foreach \n [count=\i from 0] in {36,...,47} {
    \addplot+[boxplot, fill, draw=black] table[y index=\n] {\csvdata};
  }

  \nextgroupplot[title={FID $\downarrow$}, ytick={25,50, 75,..., 150}]
  \foreach \n [count=\i from 0] in {48,...,59} {
    \addplot+[boxplot, fill, draw=black] table[y index=\n] {\csvdata};
  }

\end{groupplot}
\path (group c1r1.south east) -- node[below, yshift=-.1cm, xshift=5.5cm]{\pgfplotslegendfromname{gl}} (group c2r1.south east);
\end{tikzpicture}}
  \caption{Ablation on BRL\@. Light colors indicate the absence of $\mathcal{L}_b$ ($\lambda=0$), while dark colors indicate its presence ($\lambda=1$).}
  \label{fig:brl}
\end{figure*}
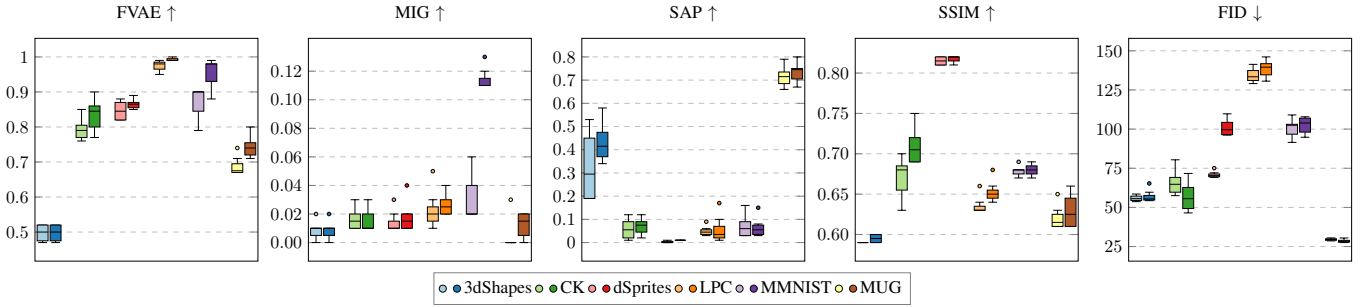

\begin{figure}[tb]
\centering
\scriptsize
  \setlength{\subfigsz}{\linewidth}
  \setlength\tabcolsep{1.5pt}
  \begin{tabular}{cc}
    Source  & Driving \\
    \boximg{.14}{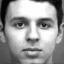} & \boximg{.84}{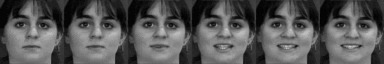} \\
    $c=1$ & \boximg{.84}{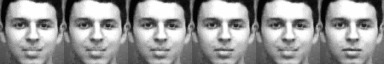} \\
    $c=3$ & \boximg{.84}{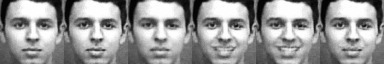} \\
    \boximg{.14}{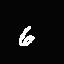} & \boximg{.84}{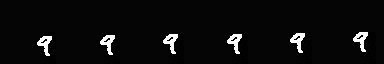} \\
    $c=1$ & \boximg{.84}{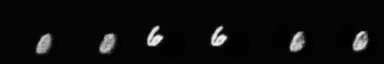} \\
    $c=3$ & \boximg{.84}{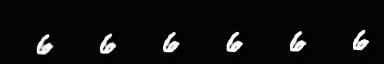}
  \end{tabular}
\caption{Qualitative comparison of performances for the frame version of MTC-VAE ($c=1$) and the chunk version with a temporal neighborhood of $c=3$ in reenactment quality and inter-chunk consistency.}
\label{fig:ablaton_c}
\end{figure}

We generated \num{10000} videos, each one from a source video $S$ and driving video $D$.
For $\beta$-TCVAE and MTC-VAE, we fixed the content representation of the first chunk of $S$, replicated it, and concatenated each replica to the motion representation of each chunk in $D$\@.
Due to the assumption of appearance preservation throughout the video, our model must be able to reconstruct the video from the appearance representation of any of their chunks.
We decided to use the first chunk of each video for easinesses in the implementation.
The reenacted video was obtained by decoding the resulting vectors.
For dis-VAE, we obtained the content representation from the mean of the frames' appearances and sequentially calculated the motion representations.
For SVG-LP, we obtained the representation from the inference model of the first frame of $S$ and concatenated it with each representation yielded by the learned prior on each frame of $D$.
For $\beta$-TCVAE, since we do not know which units correspond to content and which ones to motion, we considered the classification scheme used to calculate the FVAE metric, which returns an estimate of the units that are more likely to represent either content and motion.
Based on these criteria, we swapped the units that are more likely to represent motion from $D$ to $S$.

Our metrics are frame-wise Structural Similarity (SSIM)~\cite{Wang2004} to quantify identity preservation after reenactment (\ie, whether the reenacted video contains the content of $S$ and no leaked content of $D$), and frame-wise Fréchet Inception Distance (FID)~\cite{Heusel2017} to assess the realism of the reenacted videos.
Table~\ref{tab:baselines} shows the performance of the models for SSIM and FID\@.
In half of the cases, MTC-VAE outperforms the baselines, but its superiority is not as significant as it is in disentanglement.

Due to the lack of metrics to assess that the reenacted video mimics $D$, we provide a qualitative assessment between videos reenacted by the models and their corresponding source videos.
Fig.~\ref{fig:comparison} shows some examples.
It can be seen that MTC-VAE yields reenacted videos that are better synchronized \wrt $D$ than the baselines.
Also, in terms of sharpness, identity preservation, and inter-chunk consistency, MTC-VAE shows a clear advantage.
In general, dis-VAE was more successful in representing time-dependent features than $\beta$-TCVAE\@.
Qualitatively, SVG-LP yielded the poorest reenactment.

Additional results are in Appendix~\ref{sec:detailed_pictures}.
We explored the limits of our model on high-resolution videos (Appendix~\ref{sec:hq}) and on a real-world human-action dataset (Appendix~\ref{sec:taichi}).
Although it has shown to be robust in high-resolution videos, our experiments on human-action datasets make evident the fact that exclusively-CNN-based architectures fall short in reconstructing large motions~\cite{Siarohin2018, Balakrishnan2018}, like the ones done by the human body.
We show that the yielded representations are successful in capturing the semantics of the content and motion of the videos, which suggests that our model obtains meaningful representations of any kind of data.
However, its effectiveness for reconstruction and reenactment is restricted to motions with fewer degrees of freedom (like simple trajectories, facial expressions, and a reduced set of human actions).
These experiments reveal that the bottleneck of the model is the decoder.

\subsection{Ablation Studies}
\label{sec:ablation}

We conducted ablation studies to determine the impact of the chunk size ($c$), the order of the model ($O$), the hyperparameter $\beta$, and the presence/absence of the Blind Reenactment Loss ($\lambda$).
Figs.~\ref{fig:cs} and~\ref{fig:brl} show, respectively, charts on the ablative study on $c \in \{1,3,5,7,9\}$ and $\lambda \in \{0,1\}$.
In Appendix~\ref{sec:detailed_results}, we present complete examples with all the cases on the ablation study, tables with the detailed scores, and the ablation on $O$.

In Fig.~\ref{fig:cs}, we plotted the curves of the metrics as a function of~$c$.
Most of them peaked in \num{3} or \num{5} for FVAE and SAP, meaning that middle-sized chunks are preferable.
For SSIM, when $c > 5$, there is a slight decrease on performance and, although for $c \le 5$ performance is similar, it reaches is lowest variability at $c = 5$ (\cf gray curve).
FID shows a heterogeneous behavior among the datasets.
For CK+ and LPC, the greater the chunk size, the better the performance while the opposite stands for 3dShapes.
For MMNIST, middle values attain the best performance, while LPC shows its worst performance at the same values.
Table~\ref{tab:cs} presents more detailed results.

Although there is a pattern in most of the metrics pointing to a better performance with middle-sized chunks, numerically, the impact on the chunk size may be little significant for the metrics considered.
A more explicit impact on the performance of using chunks ($c > 1$) instead of frames ($c = 1$) is qualitatively evidenced in both reenactment quality and inter-chunk consistency.
As we do not count on metrics to quantify such properties, we depict in Fig.~\ref{fig:ablaton_c} the perceptual difference of performance between the frame and the chunk version of MTC-VAE.
Both CK+ and MMNIST show poor reenactment performance for $c=1$.
This suggests that wider temporal neighborhoods eases motion encoding, to be transferred between videos more accurately, as well as it also eases smoothness.
We show a thorough comparison in Appendix~\ref{sec:detailed_pictures}.

Fig.~\ref{fig:brl} shows the impact of BRL on the performance metrics.
The boxes correspond to the distribution of the five experiments associated with each configuration, due to the 5-fold cross-validation scheme.
Boxes with light colors indicate the performance when $\lambda = 0$, and the ones with dark colors when $\lambda = 1$.
Regarding disentanglement, it can be seen that the positive impact of the BRL is significant in general for FVAE, except for the 3dShapes datasets.
For MIG and SAP, the impact is not that significant, however, this is expected, since both metrics measure compactness, and the BRL loss is not designed for this objective.
Regarding reconstruction metrics (SSIM and FID), its impact was not significant and, in the case of FID, it showed to decrease the performance in dSprites, LPC and MMNIST\@.
Regarding the order of the model, we concluded that optimal values of $O$ are $2$ or $3$, depending on the length of the videos in the dataset (\cf Appendix~\ref{sec:detailed_results}).
Since the complexity of the model is quadratic \wrt to $O$, higher values are not worth considering.

\subsection{Performance on Downstream Tasks}
\label{sec:dt}

\begin{table}[tb]
  \caption{Content (C)/Motion (M) classification accuracy. (*\,$c=1$)}
  \label{tab:2fdt}
  \scriptsize
  \centering
  \setlength\tabcolsep{2.3pt}

  \resizebox{\linewidth}{!}{%
  \begin{tabular}{@{}lSScSScSS@{}}
    \toprule
    & \multicolumn{2}{c}{3dShapes} &
    & \multicolumn{2}{c}{CK+} &
    & \multicolumn{2}{c}{dSprites} \\
    \cmidrule{2-3} \cmidrule{5-6} \cmidrule{8-9}
    & {C} & {M} &
    & {C} & {M} &
    & {C} & {M} \\
    \cmidrule{2-3} \cmidrule{5-6} \cmidrule{8-9}
    $\beta$-TCVAE &     .53(5) &     .44(5) & &     0.90(2) &     .52(5) & &     .22(1) &     .62(2) \\
    dis-VAE       &     .48(1) &     .42(1) & & \bf 1.00(0) &     .62(4) & &     .54(6) &     .59(1) \\
    SVG-LP        &     .11(2) &     .11(1) & &     0.87(4) &     .60(2) & &     .00(0) &     .60(1) \\
    MTC-VAE       & \bf .95(1) & \bf .59(1) & &     0.97(1) & \bf .68(7) & & \bf .61(1) & \bf .63(3) \\
    MTC-VAE*      &     .46(1) &     .41(2) & &     0.94(1) &     .63(8) & &     .20(8) &     .60(3) \\
    \cmidrule{2-3} \cmidrule{5-6} \cmidrule{8-9}
    & \multicolumn{2}{c}{LPC} &
    & \multicolumn{2}{c}{MMNIST} &
    & \multicolumn{2}{c}{MUG} \\
    \cmidrule{2-3} \cmidrule{5-6} \cmidrule{8-9}
    $\beta$-TCVAE &     .11(1) & \bf .99(1) & &     .32(03) & \bf .26(5) & & \bf 1.00(0) &     .54(6) \\
    dis-VAE       &     .43(2) &     .95(1) & & \bf .54(08) &     .16(1) & & \bf 1.00(0) &     .55(2) \\
    SVG-LP        &     .00(0) &     .54(3) & &     .14(01) &     .14(1) & &     0.48(4) &     .34(1) \\
    MTC-VAE       &     .65(1) &     .93(4) & &     .48(10) &     .20(6) & & \bf 1.00(0) & \bf .79(5) \\
    MTC-VAE*      & \bf .68(3) &     .97(1) & &     .45(07) &     .19(2) & & \bf 1.00(0) &     .70(8) \\
    \bottomrule
  \end{tabular}}
\end{table}

\begin{table}[tb]
  \caption{Classification accuracy in multiple factors. (*\,$c=1$)}
  \label{tab:nfdt}
  \scriptsize
  \centering
  \setlength\tabcolsep{4pt}
  \begin{tabular}{@{}rSSSSS@{}}
    \toprule

    Factor & {$\beta$-TCVAE} & {dis-VAE} & {SVG-LP} & {MTC-VAE} & {MTC-VAE*} \\
    \midrule & \multicolumn{5}{c}{3dShapes} \\
    \cmidrule{2-6}
    Floor hue & .94(04) & \bf 1.00(0) & .27(05) &     .99(1) & .99(1) \\
    Wall hue & .95(07) & \bf 1.00(0) & .53(11) &     .97(3) & .97(3) \\
    Obj. hue & .82(10) & \bf 1.00(0) & .17(02) &     .95(4) & .95(4) \\
    Init. size & .66(25) &     0.97(4) & .66(18) & \bf .98(3) & .97(4) \\
    Final size & .30(09) &     0.25(2) & .46(06) & \bf .51(3) & .50(3) \\
    Shape & .25(06) &     0.26(2) & .25(02) & \bf .38(2) & .37(3) \\
    Init. persp. & .20(06) &     0.17(0) & .25(01) & \bf .28(3) & .27(2) \\
    Final persp. & .17(05) &     0.18(1) & .15(01) & \bf .25(5) & .24(2) \\
    \midrule & \multicolumn{5}{c}{dSprites} \\
    \cmidrule{2-6}
    R & .02(00) & .03(0) & .01(00) & \bf .07(00) & .04(1) \\
    G & .02(00) & .03(0) & .01(00) & \bf .08(01) & .04(1) \\
    B & .03(01) & .03(0) & .01(00) & \bf .07(01) & .03(0) \\
    Shape & .46(02) & .45(1) & .34(01) & \bf .51(05) & .50(4) \\
    Scale & .44(03) & .50(1) & .18(01) & \bf .56(03) & .47(3) \\
    Rot. & .12(01) & .10(1) & .03(00) & \bf .49(04) & .10(4) \\
    Traj. & .62(02) & .59(1) & .60(01) & \bf .63(03) & .60(3) \\
    \midrule & \multicolumn{5}{c}{LPC} \\
    \cmidrule{2-6}
    Body &     .18(01) & 0.54(4) & .13(00) & \bf 0.97(01) & \bf .97(1) \\
    Gender &     .60(04) & 0.91(3) & .50(00) & \bf 1.00(00) &     .98(2) \\
    Shirt &     .72(03) & 0.96(3) & .66(04) & \bf 1.00(00) &     .85(8) \\
    Pants &     .68(04) & 0.89(2) & .18(01) & \bf 0.99(00) &     .87(7) \\
    Hair & \bf .97(02) & 0.89(4) & .28(01) &     0.91(04) &     .92(1) \\
    Hat &     .69(01) & 0.88(2) & .56(00) & \bf 1.00(00) &     .99(1) \\
    Action &     .62(04) & 0.64(3) & .61(04) &     0.64(03) & \bf .69(3) \\
    Perspective &    .94(03) & \bf 1.00(0) & .58(03) &     0.72(14) &   .98(2) \\
    \bottomrule
  \end{tabular}
\end{table}

\begin{figure*}[tb]
\centering
  \begin{subfigure}{.32\textwidth}
    \includegraphics[width=\textwidth]{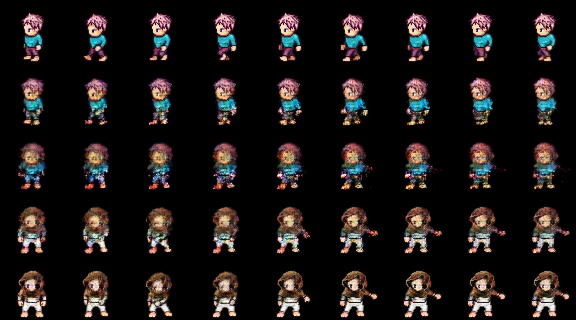}
    \caption{Full latent space}
  \end{subfigure}
  \begin{subfigure}{.32\textwidth}
    \includegraphics[width=\textwidth]{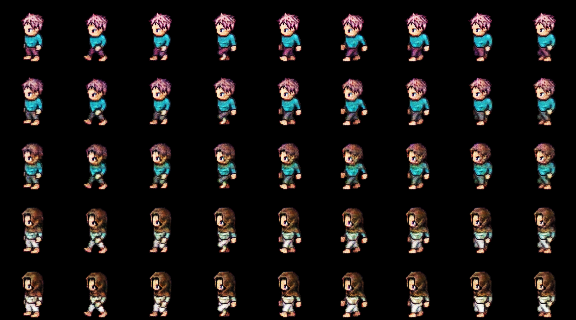}
    \caption{Content subspace}
  \end{subfigure}
  \begin{subfigure}{.32\textwidth}
    \includegraphics[width=\textwidth]{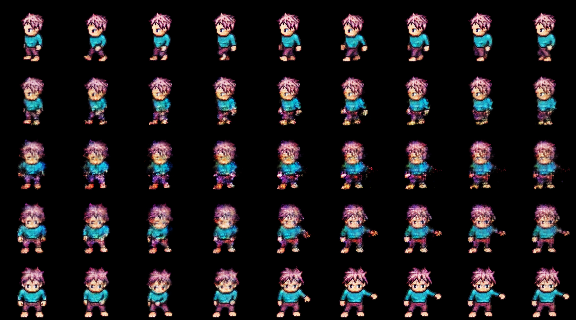}
    \caption{Motion subspace}
  \end{subfigure}
\caption{%
  Latent-space traversals on LPC.
  The upper and lower sequences are, respectively, the start and endpoints of the traversals.}
\label{fig:traversals}
\end{figure*}

\begin{figure}[tb]
\centering
  \setlength{\subfigsz}{\linewidth}
  \setlength\tabcolsep{1.5pt}
  \begin{tabular}{rc}
    Original & \boximg{.8}{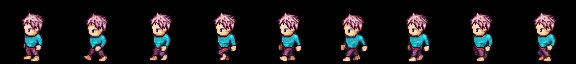} \\
    Hair color & \boximg{.8}{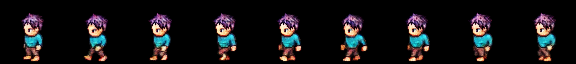} \\
    Hairstyle & \boximg{.8}{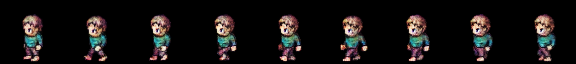} \\
    Shirt & \boximg{.8}{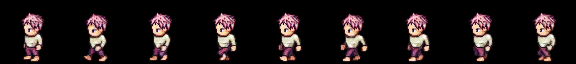} \\
    Pants & \boximg{.8}{traverse_pants_color_13.png} \\
    Perspective & \boximg{.8}{traverse_perspective_14_18_20.png}
  \end{tabular}
\caption{Some controllable visual traits by traversing specific latent units.}
\label{fig:controllable}
\end{figure}

To evaluate the robustness of the learned disentangled representations, we extracted them from the datasets, and trained a Linear Support Vector Machine to assess whether they are linearly separable.
We chose a simple classifier, as more sophisticated ones are prone to work around weaker representations, hindering the comparison between our model and the baselines.
We tested the models in (i)~content-motion and (ii)~multi-factor classification.

For the first scenario, we used the same ground-truth labels to calculate appearance/motion disentanglement, and report the obtained accuracies in Table~\ref{tab:2fdt}, showing that recognizing content is easier than actions.
In most of the datasets, our model outperforms the baselines in both content and motion.

For the second scenario, we used the same ground-truth labels to calculate multi-factor disentanglement.
This scenario was harder for all the models (\cf Table~\ref{tab:nfdt}).
However, ours outperformed the rest in most of the cases.
This is expected since none of them was trained for multi-factor disentanglement.
Notice that each row in Table~\ref{tab:nfdt} is a classification scheme on different sets of classes.
\Eg, for dSprites, factor \textit{R} represents the red RGB contribution of the shape, so it is a 256-class problem, while factor \textit{Shape} is a 4-class problem, as there are only four different shapes in the dataset (\cf Table~\ref{tab:factors}).
In both scenarios, the chunk-wise version of our model outperformed the frame-wise version (MTC-VAE*) most of the times.

\subsection{Latent-Space Traversals}
\label{sec:traversals}

We include some examples of latent-space traversals on the LPC dataset, to show how MTC-VAE could be used for conditional video generation.
Fig.~\ref{fig:traversals} shows three trajectories, between two videos~$x_0$ and~$x_1$, separated by 5 steps.
The leftmost trajectory traverses the whole latent space, so it is possible to see the complete transformation from~$x_0$ to~$x_1$.
The central trajectory is done in the content subspace while remaining stationary in the motion space, so it can be seen how the endpoint is a video with the appearance of $x_1$ and the motion of $x_0$.
The opposite can be observed in the rightmost trajectory, which only traverses the motion subspace.

All the trajectories are linear, so it is expected that examples in the middle do not look plausible, due to a high probability of sampling outside either~$q_\phi(z\given x_k)$ or~$q_\gamma(w_k\given x_k)$.
To correctly traverse the latent space requires awareness of its topology.
We leave as future work to explore more sophisticated methods to traverse the space of our model~\cite{Ye2019, Song2020cyb}.

Fig.~\ref{fig:controllable} shows some examples of controllable video generation.
We highlight that we do not expect to perform this task perfectly, as we focus exclusively on content-motion disentanglement, so it is normal that visual traits that should be independent (\eg hair color and skin color) happen to be entangled in the representation.
However, it is possible to independently traverse each latent unit of the space and manually check which visual traits were affected.
The sequences of Fig.~\ref{fig:controllable} are the endpoints of the trajectories (Appendix~\ref{sec:detailed_pictures} shows the complete trajectories), and each one shows a visual trait that was affected by traversing latent units.
Most of them were affected by only one unit: hair color ($z[5]$), hairstyle ($z[7]$), shirt color ($z[11]$), and pants color ($z[13]$).
Motion-related units were more difficult to traverse, since independent motion traits of the video remain more entangled than the appearance ones, as shown by our results on multi-factor classification (Table~\ref{tab:nfdt}).
This means that traversals have a high risk of sampling outside the support of $q_\gamma(w_k\given x_k)$.
The last example in Fig.~\ref{fig:controllable} was constructed by traversing~$w[0]$, $w[4]$, and~$w[6]$, and it is clear that we sampled outside $q_\gamma(w_k\given x_k)$.
This set of experiments show that it is possible to interpret, to some extent, the meaning of the components of the latent representations.

\section{Conclusion}
\label{sec:conclusions}

Our proposed MTC-VAE for content-motion disentanglement learns to represent videos as a consistent sequence of chunks that are independent at generation time, but dependent at inference time.
It considers two extensions to the VAE formulation: (i)~training the model such that each chunk implicitly contains information about the whole video under the assumption of content invariability, while separating motion per chunk, and (ii)~using the task of video reenactment as an inductive bias to leverage the learning of independent content and motion representations.
MTC-VAE yields less latent vectors to represent a video (one per chunk, instead of per frame).
To reconstruct one video, it is trained with chunks modeled as independent random variables at generation time.
Given that a chunk does not depend on the reconstruction of the previous one, all chunks in a video can be reconstructed in a single forward-pass.
The experiments show the capacity of our chunk-wise approach in learning time-dependent and -independent representations from videos as well as the positive impact of video reenactment as an inductive bias to improve such representations.
Our ablative study on the size of the chunks shows a better disentanglement and VR performance of middle-sized chunks, over the frame-wise approach.
We also showed the superiority of MTC-VAE for multiple-factor disentanglement, even though it was not explicitly trained for more than two factors.
We explored the limits of our model in additional experiments on high-resolution videos (Appendix~\ref{sec:hq}) and on a real-world human-action dataset (Appendix~\ref{sec:taichi}).
These experiments reveal that the bottleneck of the model is the decoder, whose enhancement we leave for future work as well as exploring different latent and data priors, and devising fusion strategies for the chunks to yield more informative gradients and a better reconstruction, as well as disentanglement quality.

\renewcommand{\algorithmiccomment}[1]{\hfill$\triangleright$\,#1}

\ifdef{\boximg}{%
\renewcommand{\boximg}[3]{\parbox[c]{#1\subfigsz}{\includegraphics[width=#2\subfigsz]{#3}}}%
}{%
\newcommand{\boximg}[3]{\parbox[c]{#1\subfigsz}{\includegraphics[width=#2\subfigsz]{#3}}}%
}

\appendices
\counterwithin{figure}{section}
\counterwithin{table}{section}

\section{Derivation of the ELBO}
\label{sec:elbo}

In this section, we present the derivations of the statements introduced in Section~\ref{sec:experiments} to construct the loss functions of our model based on the Evidence Lower Bound (ELBO) of the expected log-likelihood of our model.

Let  the video~$x$ be a sequence of $K$~chunks, $x = (x_k)_{k=1}^K$.
Similarly, let $w = \left(w_k\right)_{k=1}^K$ be the sequence of motion representations for all the chunks on the video~$x$. 
For the $k$-th chunk, we model the content and motion as independent latent variables $z$ and $w_k$.

We are interested in maximizing the expected log-likelihood of the videos \wrt the data empirical distribution~$q(x)$.  
First, let's consider it based on the sets~$x$ and~$w$ such that
\begin{align}
\E_{q(x)} \log p(x) =& \E_{q(x)} \log\iint p(x, w, z) \dif w \dif z, \\
=& \E_{q(x)} \log\iint \frac{q(w, z \given x)}{q(w, z \given x)}p(x,w,z) \dif w \dif z, \\
\ge& \E_{q(x)}\E_{q(w, z \given x)} \left[\log\frac{p(x,w,z)}{q(w, z \given x)}\right], \\
=& \E_{q(x)}\E_{q(w, z \given x)} \left[\log\frac{p(x \given w,z)p(w)p(z)}{q(w \given x)q(z \given x)}\right], \\\nonumber
=& \E_{q(x)}\E_{q(w, z \given x)} \left[\log p(x \given w,z) + \log\frac{p(w)}{q(w \given x)} \right.\\
&\left.+ \log\frac{p(z)}{q(z \given x)}\right].
\end{align}
However, we are interested in modeling the chunks and their respective latent variables.
Hence, we need a change in the variable.
First, let's consider the video distribution based on its chunks as $q(x) = \prod_{k=1}^K q(x_k \given x_{k-1})$, such that $q(x_1 \given x_0) \equiv q(x_1)$, \ie, we consider the video as a Markov chain of chunks.
Then, by plugging into the sequence representations of the video~$x$ and the motion latent variable~$w$, we get
\begin{align}
\E_{q(x)} \log p(x) \geq& \E_{\prod_k q(x_k \given x_{k-1})} \E_{\prod_k q(w_k, z \given x_k)} \Bigg[ \nonumber\\
& \log  \prod_k p(x_k \given w_k, z) + \nonumber \log\frac{\prod_k p(w_k)}{\prod_k q(w_k \given x_k)} \nonumber\\
&+ \log\frac{\prod_k p_k(z)}{\prod_k q(z \given x_k)} \Bigg].
\end{align}
We denote the content prior distribution over the sequence of chunks as $p(z) = \prod_k p_k(z)$.
(Abusing notation, we will refer to these priors as $p_k(z) = p(z)$ since they are all the same over the sequence.)
Then, we can simplify the expected log-likelihood as
\begin{align}
\E_{q(x)} \log p(x) & \geq \E_{\prod_k q(x_k \given x_{k-1})} \sum_k \E_{ q(w_k, z \given x_k)} \Bigg[\log  p(x_k \given w_k, z) \nonumber\\
&+ \log\frac{p(w_k)}{q(w_k \given x_k)} + \log\frac{p(z)}{q(z \given x_k)} \Bigg], \\
& = \E_{\prod_k q(x_k \given x_{k-1})} \sum_k \Bigg\{ \E_{ q(w_k, z \given x_k) } \left[\log  p(x_k \given w_k, z) \right] \nonumber\\
&  - \kl{q(w_k \given x_k)}{p(w_k)} \nonumber\\
&- \kl{q(z \given x_k)}{p(z)} \Bigg\}.
\end{align}

Notice that the final function corresponds to the expectation over the empirical chain of chunks.
In our experiments, we simulate this process by sampling throughout the video to obtain the chunks and then compute the summation over the losses.

\section{Implementation Details}
\label{sec:implementation}

\subsection{Architecture}

Our model consists of two encoding streams, corresponding to $q_\phi(z \given x_{k})$ and  $q_\gamma(w_k \given x_{k})$, and one decoding stream, corresponding to $p_\theta(x_{k})$, defined in Section~\ref{sec:method}).
All the streams have five 3D-convolutional layers, with batchnorm and ReLU activations. 
The number of filters in the hidden layers of the decoder is double the number of filters in the encoders.

As previous works on DRL from video and VR~\cite{Siarohin2019,Siarohin2019nips,Aberman2019}, we used an appearance-suppressed input to the motion encoding stream. 
In our case, we added a layer that calculates the optical flow of the chunk with the Lucas-Kanade method~\cite{Lucas1981}.

We use a \textit{Bernoulli observation VAE} where the observed samples of the decoder are used as logits of a Bernoulli distribution, in contrast with the traditional Gaussian observations. 
We observed a remarkable superiority at reconstruction time of the Bernoulli observations, particularly for videos where the proportion of the object of interest \wrt the background is reasonably low.
We consider standard Normal priors for both the content and motion latent representations, \ie, $p(z) = \mathcal{N}(0,1)$, and  $p(w_k) = \mathcal{N}(0,1)$ for all $k$.

\begin{algorithm}[tb]
   \caption{Chunk Sequence Learning training procedure algorithm.}
   \label{alg:sampling}
\begin{algorithmic}
\small
\STATE {\bf Input:} batch $\mathcal{X}$, video length $T_x$, chunk sequence length $O$, chunk size $c$
\FOR{\textbf{each} $x \in \mathcal{X}$}
  \STATE $h \sim U(1, T_x - cO)$ \COMMENT{Random starting position}
  \STATE Sample $\{x_k\}_{k=1}^O$, $O$ consecutive size $c$ chunks starting at $h$
  \STATE \COMMENT{Approx. of $x_k \sim q(x_k \given x_{k-1}), \forall\,k$}
  \FOR{$k=1$ {\bf to} $O$}
    \STATE $w_k \sim q(w_k \given x_k)$
    \FOR{$j=1$ {\bf to} $O$}
      \STATE $z_j \sim q(z \given x_j)$
      \STATE $\mathcal{L}_r \mathrel{{+}{=}} \log p_\theta(x_k \given w_k, z_j)$
    \ENDFOR
    \STATE $\mathcal{L}_a \mathrel{{-}{=}} \kl{q(z \given x_k)}{p(z)}$
    \STATE $\mathcal{L}_m \mathrel{{-}{=}} \kl{q(w_k \given x_k)}{p(w_k)}$
  \ENDFOR
\ENDFOR
\FOR{\textbf{each} $s,d \in \mathcal{X} \times \mathcal{X}, s \ne d$}
  \STATE $h_s \sim U(1, T_s - cO)$
  \STATE $h_d \sim U(1, T_d - cO)$
  \STATE Sample $\{s_k\}_{k=1}^O$, $O$ consecutive size $c$ chunks starting at $h_s$
  \STATE Sample $\{d_k\}_{k=1}^O$, $O$ consecutive size $c$ chunks starting at $h_d$
  \FOR{$l=1$ {\bf to} $O$}
    \STATE $w_l^d \sim q(w_k \given d_l)$
    \FOR{$j=1$ {\bf to} $O$}
      \STATE $z_j \sim q(z \given s_j)$
      \FOR{$i=1$ {\bf to} $O$}
        \STATE $z_i \sim q(z \given s_i)$
        \STATE $\mathcal{L}_b \mathrel{{-}{=}} \skl{p_\theta(x_l \given w_l^d, z_j^s)}{p_\theta(x_l \given w_l^d, z_i^s)}$
        \STATE \COMMENT{SKL as defined in Equation~\ref{eq:brl}}
      \ENDFOR
    \ENDFOR
  \ENDFOR
\ENDFOR
\STATE $\mathcal{L} = \mathcal{L}_r + \lambda\mathcal{L}_b + \beta(\mathcal{L}_a + \mathcal{L}_m)$
\STATE $(\phi, \gamma, \theta) \mathrel{{-}{=}} \nabla_{(\phi, \gamma, \theta)} \mathcal{L}$ \COMMENT{Backprop the loss over the parameters}
\end{algorithmic}
\end{algorithm}

The loss functions $\mathcal{L}_r$, $\mathcal{L}_a$, and $\mathcal{L}_m$ require consecutive chunks of a unique video, while $\mathcal{L}_b$ requires chunks of the source and the driving video.
In order to train all the losses in the same forward pass, we feed the model with batches of $O$-tuples of consecutive chunks, as shown in the Algorithm~\ref{alg:sampling}, which describes in detail the training procedure of our model.
For $\mathcal{L}_b$, we create a reversed copy of the batch to be used as the batch of driving videos, while the original batch corresponds to the source videos.
This gives a sense of completeness for training because it ensures that source videos will also act as driving videos, and vice-versa, in the same forward pass.
Although Algorithm~\ref{alg:sampling} is expressed so, for each batch, all the possible pairs of videos are used as source and driving, in practice, it is unfeasible because the calculation of $\mathcal{L}_b$ takes cubic time \wrt $O$.
We can argue that, by means of the stochastic batched training, most of the possible pairs of videos can be covered for our model, if trained for enough time.

\subsection{Model Training}

Making use of labels describing the factors of variation in a video, such as the identity of the object of interest or its motion, we split the datasets in training-test and tested our model in two generalization scenarios.
We will refer to this as a \emph{soft generalization} scenario, in which the model is requested to reconstruct novel videos from contents and motions seen in training time.
We included two \emph{hard generalization} scenarios: the \emph{appearance holdout} scenario, in which the model is requested to reconstruct novel videos with appearances that were not seen in training time, and the \emph{motion holdout} scenario, in which the model is requested to reconstruct novel videos with motions that were not seen in training time.
The quantitative results presented in the main text of this paper (Table~\ref{tab:baselines}, and Figs.~\ref{fig:comparison} and~\ref{fig:cs}) correspond to the soft generalization scenario.
We show in Appendix~\ref{sec:detailed_results} the quantitative performance of MTC-VAE and the baselines in the three generalization scenarios, as well as detailed results on the ablation studies, corresponding to the soft generalization scenario.

To run the complete set of experiments, including the baselines, the hyper-parameter search, and the ablation study of our model, we used a total of $12$ GPUs Titan X, Titan Xp, RTX 2080 Ti, RTX 5000, GTX 1080 Ti, and Tesla P100.
However, our model can be executed in a single GPU of 12 GB memory, and the training time varies from $20$ minutes to $12$ hours, depending on the length of the videos, the chunk size, and more importantly, the order of the model.
Given two videos, the chunk-wise reenactment process takes no more than $2$ seconds.

\subsection{Data}
\label{sec:data}

\begin{table}[tb]
\caption{Factors of variation for each multi-factor dataset.}
\label{tab:factors}
\centering
\scriptsize
\setlength{\tabcolsep}{2pt}
\resizebox{\linewidth}{!}{%
\begin{tabular}{>{\ttfamily}l>{\ttfamily}l>{\ttfamily}r}
\toprule
\multicolumn{1}{l}{\textbf{Factor (size)}} & \multicolumn{1}{l}{\textbf{Labels}} & \multicolumn{1}{r}{\textbf{S}} \\
\midrule
\multicolumn{3}{l}{3dShapes} \\
\midrule
floor\_hue (10) & [0-9] & C \\ %
wall\_hue (10) & [0-9]  \\ %
object\_hue (10) & [0-9]  \\ %
shape (4) & [0-3]  \\ \midrule
init\_size (8) & [0-7] & M \\ %
final\_size (8) & [0-7] \\ %
init\_angle (15) & [0-14] \\ %
final\_angle (15) & [0-14] \\ %
speed (2) & [1-2] \\
\midrule
\multicolumn{3}{l}{dSprites} \\
\midrule
R (256) & [0-255] & C \\ %
G (256) & [0-255] \\ %
B (256) & [0-255] \\ %
orientation (40) & [0-39] \\ %
shape (3) & [0-2] \\ %
scale (6) & [0-5] \\ \midrule
x\_init (32) & [0-31] & M \\ %
y\_init (32) & [0-31] \\ %
x\_final (32) & [0-31] \\ %
y\_final (32) & [0-31] \\ %
speed (3) & [1-3] \\ %
trajectory (2) & linear arc \\
\midrule
\multicolumn{3}{l}{LPC} \\
\midrule
Sex (2) & male female & C \\ %
Body(10) & light dark dark2 darkelf orc redorc \\
         & darkelf2 tanned tanned2 skeleton \\ %
Shirt (8) & longsleeve\_maroon sleeveless\_teal \\
            & sleeveless\_maroon longsleeve\_teal \\
            & longsleeve\_brown sleeveless\_white \\
            & longsleeve\_white sleeveless\_brown \\ %
Pants (4) & magenta red teal white \\ %
Hat (8)   & none bandana\_red cap\_leather chain \\
          & helmet\_golden helmet\_metal \\
          & hood\_chain hood\_cloth \\ %
Hair (17) & none bangslong2\_brunette bedhead\_brunette \\
          & bangslong2\_green swoop\_red mohawk\_red \\
          & shoulderl\_raven plain\_blue loose\_blue \\
          & shoulderl\_pink messy2\_raven \\
          & bedhead\_green messy2\_pink swoop\_white \\
          & mohawk\_white loose\_blonde plain\_blonde \\ \midrule
Action (6) & spellcast thrust walk slash shoot hurt & M \\ %
Perspective (4) & back front left right \\
\bottomrule
\end{tabular}}
\end{table}

\textbf{Cohn-Kanade (CK+) facial expressions dataset.}~\cite{Kanade2000, Lucey2010} $326$ gray-scale videos of $64 \times 64$ pixels of $118$ characters performing six actions: anger, disgust, fear, happy, sad, and surprise.
This dataset only provides two-factor labels: identity and expression.

\textbf{Multimedia Understanding Group (MUG) facial expressions dataset.}~\cite{Aifanti2010} $931$ RGB videos of $64 \times 64$ pixels of $52$ characters performing six actions: anger, disgust, fear, happy, sad, and surprise.
This dataset only provides two-factor labels: identity and expression.

\textbf{Liberated Pixel Cup (LPC).}\footnote{\url{https://github.com/jrconway3/Universal-LPC-spritesheet} as as September, 2019. \newline\url{http://lpc.opengameart.org/} as September, 2019.} We generated \num{10000} RGB videos of $64\times 64$ pixels creating $24$ motions classes performed by the characters, which correspond to six actions (walk, spellcast, thrust, shoot, hurt, and slash) times four perspectives (front, back, left, and right).
For content, we combined different genders, body types, hairstyles, and clothes, creating a large number of different identities.
In total, we generated \num{10000} videos for training.
Table~\ref{tab:factors} shows the factors of variation used to evaluate multi-factor disentanglement (Table~\ref{tab:mf}).
For content-motion disentanglement (Table~\ref{tab:baselines}), we joined these factors in two supersets, as pointed in the \textbf{S} column in Table~\ref{tab:factors}.

\textbf{Moving MNIST (MMNIST).}~\cite{Srivastava2015}
We generated \num{10000} binary videos of $64\times 64$ pixels with ten identities, corresponding to the digits from $0$ to~$9$.
All the videos have $32$ frames.
The digits follow linear trajectories from random starting points. 
We created $14$ motion classes that distinguish the direction of the trajectory (\eg, down, diagonal up, right-left, left-right).
This dataset only provides two-factor labels: identity and motion.

\textbf{dSprites.}
We took the data provided in Deepmind's project\footnote{\url{https://github.com/deepmind/dsprites-dataset} as September, 2019.} and generated \num{10000} videos of $64\times 64$ pixels from the images provided.
The moving sprites have all possible sizes and shape types, yielding a large number of different identities. 
We can tweak the starting position, the final position, the velocity, and the type of trajectory (either linear or curved) of the sprite.
This yields an explosive number of motion classes so, when taking the disentanglement metrics, we decided to label the videos with either linear or curved trajectory.
Table~\ref{tab:factors} shows the factors of variation used to evaluate multi-factor disentanglement (Table~\ref{tab:mf}).
For content-motion disentanglement (Table~\ref{tab:baselines}), we joined these factors in two supersets, as pointed in the \textbf{S} column in Table~\ref{tab:factors}.

\textbf{3dShapes.}
We took the data provided in the Deepmind's project\footnote{\url{https://github.com/deepmind/3d-shapes} as  September, 2019.} and generated \num{10000} videos of $64\times 64$ pixels from the images provided.
We can take the hue of the floor, the shape, and the walls, as well as the type of shape, yielding different identities.
Regarding motion, we teak the size of the shape (yielding a heart-beat-like motion) and the perspective (yielding a camera-motion effect), attaining a large (but not explosive) number of motion classes.
Table~\ref{tab:factors} shows the factors of variation used to evaluate multi-factor disentanglement (Table~\ref{tab:mf}).
For content-motion disentanglement (Table~\ref{tab:baselines}), we joined these factors in two supersets, as pointed in the \textbf{S} column in Table~\ref{tab:factors}.

\subsection{Baselines}

As said in Section~\ref{sec:experiments}, we compared our method against dis-VAE by \textcite{Li2018}, SVG-LP by \textcite{Denton2018}, and $\beta$-TCVAE by \textcite{Chen2018dr}.
We executed code already available for the three models. 
In the case of $\beta$-TCVAE, we extended the code made available by its authors,\footnote{\url{https://github.com/rtqichen/beta-tcvae} as  December, 2019.} so their convolutional streams become 3D ones, in order to support chunks of videos. 
For SVG-LP, we used the official code provided by the authors\footnote{\url{https://github.com/edenton/svg} as  May, 2020.}
For dis-VAE, we used a public reproduction of the method\footnote{\url{https://github.com/mazzzystar/Disentangled-Sequential-Autoencoder} as  December, 2019.}
whose results on the LPC dataset seem to match with the ones presented in the paper. 
In particular, we used the encoder referred as ``full $q$'' by the authors.

We tuned the hyper-parameters of the three models, by testing a small set of variations, as described below, on all the datasets, in the soft generalization scenario, and extracted the five evaluation metrics (MIG, FVAE, SAP, SSIM, and FID).
For dis-VAE, we contrasted the ``factored q'' against the ``full q'' in order to determine which model had the best disentanglement and reconstruction performance.
We determined that the latter had the best performance.
For $\beta$-TCVAE we tunned $\beta$ and $\lambda$, and the effect of annealing each one of them while training.
For SVG-LP we tested between the VGG and the DC-GAN architectures, concluding that the latter attained the best results, so we used it for comparison.
We determined that annealing $\lambda$ while keeping $\beta$ fixed ($1.0$ for MMNIST and dSprites and $5.0$ for the rest of datasets) obtained the best results.
The best baseline configurations for each dataset were compared against our method, as shown in Table~\ref{tab:detailed_comparison}.

\subsection{Metrics Calculation}

In order to calculate the disentanglement metrics, we took all the videos of the test set ($20\%$, according to the $5$-fold cross-validation setup mentioned in Section~\ref{sec:experiments}), divided them into chunks, and calculated the latent representations of each one of the chunks.
In the case of dis-VAE, the representations were per frame.
In total, for CK+, approximately $64$ videos were used to calculate MIG, FVAE, and SAP while, for the rest of datasets, approximately \num{2000} videos were used.

We evaluated content-motion disentanglement for the five datasets (\cf Table~\ref{tab:baselines}), by considering only two factors of variation.
The 3dShapes, dSprites, and LPC datasets contain more than two factors, so we composed them, as noted in the \textbf{S} column in Table~\ref{tab:factors} to attain only the content-motion factors.
As stated in Section~\ref{sec:experiments_disentanglement}, when the number of factors is not equal to the number of units (in our case, the number of units is significantly higher than $2$), the MIG and SAP metrics are expected to be low.

Although MIG is a relatively popular metric, it penalizes dispersed representations, by considering the information gap between the first and second units that best represent a factor.
Thus, when one factor of variation is equally represented by more than one unit, that gap is expected to be low, and so does the metric.

SAP is also thought to be low when there is a mismatch between the number of units and the number of factors since this metric is based on the classification accuracy estimation (using a Linear SVM classifier) when each 1d unit is used to classify examples under each factor.

We consider the FVAE metric to be the most suitable for the objective of motion disentanglement since it only penalizes the undesirable case in which one latent unit represents more than one factor of variation, and we are only considering two factors that we expect to be fully disentangled.

For the reconstructions metrics, in theory, we can generate $n(n-1)$ reenacted videos, where $n$ is the number of videos in the test set.
It was straightforward to generate \num{10000} reenacted videos for all datasets, except for CK+, which had approximately \num{4000} videos.
We used all the generated videos to calculate SSIM and FID\@.

\section{Performance of Training with Partial Representations}
\label{sec:partial}

Aiming at reducing the computational cost of training MYC-VAE when $O$ is high, without reducing its performance, we conducted an experiment to assess the effect of subsampling the number of combinations to calculate the extended log-likelihood (\ref{eq:logp_x}) and the Blind Reenactment Loss (\ref{eq:brl}).
We set $O=4$, but instead of reconstructing $O$ times the input sequence, we reconstruct only two, by randomly sampling two of the $O$ appearance representations.
Notice that, if we sampled only one appearance representation, the BRL calculation would not be possible (see Eq.~\ref*{eq:brl}).
Due to time and computer restrictions, we performed those experiments only in the MUG dataset.

\begin{table*}[tb]
  \caption{Comparison of the model performances with and without subsampling when $O=4$ in the MUG dataset.}
  \label{tab:partial}
  \centering
  \footnotesize
  \setlength\tabcolsep{4pt}
  \begin{tabular}{lSSScSS[table-format=2.2(3)]cSS}
    \toprule
    & \multicolumn{3}{c}{Disentanglement} & & \multicolumn{2}{c}{Reconstruction} & & \multicolumn{2}{c}{Accuracy} \\
    \cmidrule{2-4} \cmidrule{6-7} \cmidrule{9-10}
    & {FVAE $\uparrow$} & {MIG $\uparrow$} & {SAP $\uparrow$} & & {SSIM $\uparrow$} & {FID $\downarrow$} & & {C $\uparrow$} & {M $\uparrow$} \\
    \cmidrule{2-4} \cmidrule{6-7} \cmidrule{9-10}
    Full    & \bf .78(4) & \bf .01(1) &     .82(4) & &     .45(1) &     39.44(355) & & 1.(0) & \bf .40(9) \\
    Partial &     .77(6) &     .00(0) & \bf .87(2) & & \bf .47(1) & \bf 40.40(205) & & 1.(0) &     .39(4) \\
    \bottomrule
  \end{tabular}
\end{table*}

Table~\ref{tab:partial} shows the results of our experiments.
These results suggest that a full representation may increase modularity (higher FVAE), while a partial representation seems to deal better with explicitness (higher SAP) and reconstruction quality (SSIM and FID).
However, the difference between those methods is not big enough to say one is better than the other, reinforcing our hypothesis that higher orders may just add too much redundancy to the training, without improving performance.
This in part may explain why our experiments showed that optimal values of $O$ are~\num{2} or~\num{3} in terms of cost/benefit, even for long sequences like the ones in MUG\@.
We also noticed that using the partial representations yielded smaller architectures (about \num{25}\% less trainable parameters), less GPU memory (about \num{50}\%) and a lower execution time (about \num{50}\%).

\section{Experiments on High-Resolution MUG}
\label{sec:hq}

We tested the effectiveness of MTC-VAE on high-resolution inputs by training it on a $256 \times 256$ version of the MUG dataset, which we will call it as MUG-HQ.

The architecture to process the $256 \times 256$ input contains two more convolutional layers in the encoders and the decoder than the $64 \times 64$ version.
Also, we doubled the size of the content and motion latent representations, and trained our model for $60$ hours, while the $64 \times 64$ model took $24$ hours to converge.

In table \ref{tab:hq} we compare the performance of MTC-VAE between MUG and MUG-HQ, in order to better analyze how the model was affected with a high-resolution input.
The SSIM and FID metrics behaved as expected: the larger the input, the harder to reconstruct it, and the harder to yielding samples that belong to the data distribution.
The performance on the downstream classification task was practically unaffected in content classification, while it had a slight drop in Motion classification.

In general, we expected all the metrics to worsen for high-resolution videos.
For that reason, the increase on the FVAE and SAP metrics is somehow surprising for us.
Our conclusion is that the increase in spatial resolution enhanced the quality of the representations, in particular, the content one, by providing mode discriminant information.
On the other side, the motion representation presented a lower action classification performance, suggesting that it contains less discriminating information.

\begin{table}[tb]
\caption{Results on MUG-HQ compared with its low-quality version. The lower part indicates the performance on downstream tasks.}
\label{tab:hq}
\centering
\begin{tabular}{lS[table-format=3.2(3)]S[table-format=3.2(3)]}
\toprule
Metrics & {Values HQ} & {Values LQ} \\
\midrule
FVAE $\uparrow$    & \bf .77(004) &       .72(004) \\
MIG $\uparrow$     & \bf .02(001) &       .01(001) \\
SAP $\uparrow$     & \bf .83(003) &       .73(005) \\
SSIM $\uparrow$    &     .61(001) & \bf   .63(002) \\
FID $\downarrow$   &   41.12(107) & \bf 28.79(115) \\
\midrule
Content $\uparrow$ &     .99(001) & \bf  1.00(000) \\
Motion $\uparrow$  &     .63(004) & \bf   .79(005) \\
\bottomrule
\end{tabular}
\end{table}

\begin{figure*}
\includegraphics[]{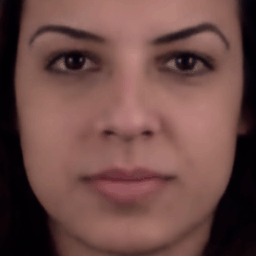}
\includegraphics[]{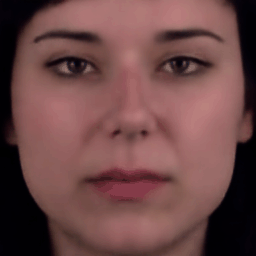}

\vspace{.8mm}
\includegraphics[]{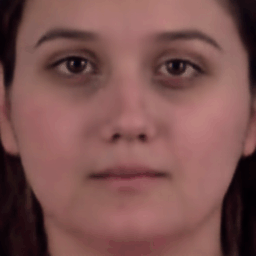}
\includegraphics[]{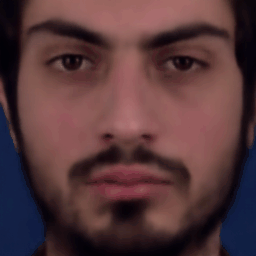}

\centering
Driving video

\includegraphics[width=.8\linewidth]{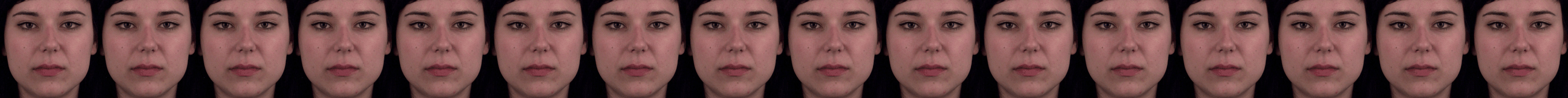}

Reenacted videos

\includegraphics[width=.8\linewidth]{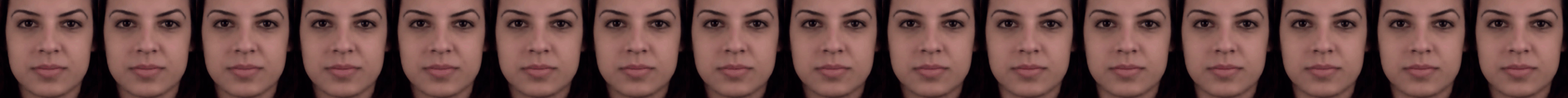}
\includegraphics[width=.8\linewidth]{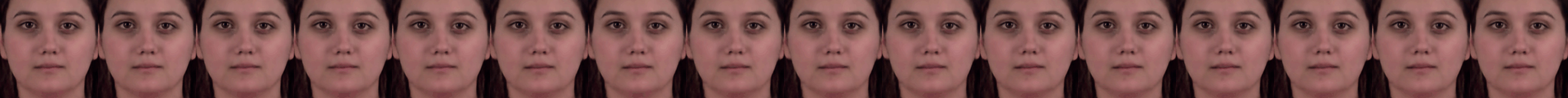}
\includegraphics[width=.8\linewidth]{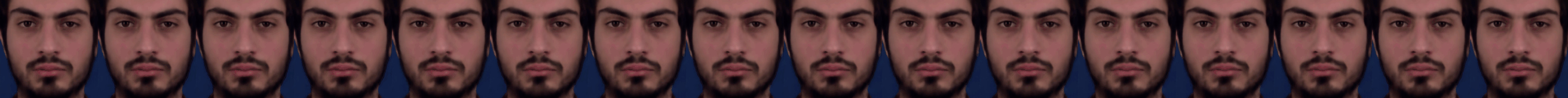}
\caption{MUG-HQ: Reenactment examples. Above: selected frames at full resolution. Below: complete sequences.}
\label{fig:hq1}
\end{figure*}

\begin{figure*}
\includegraphics[]{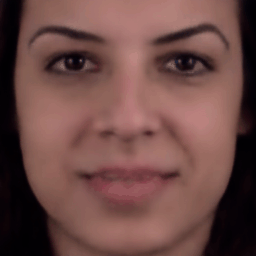}
\includegraphics[]{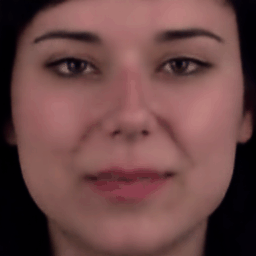}

\vspace{.8mm}
\includegraphics[]{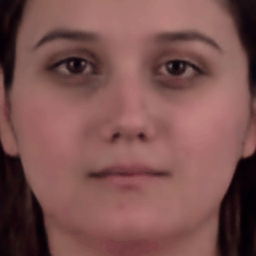}
\includegraphics[]{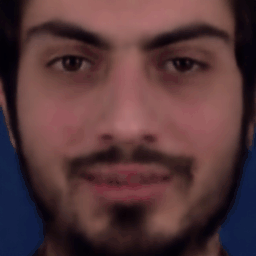}

\centering
Driving video

\includegraphics[width=.8\linewidth]{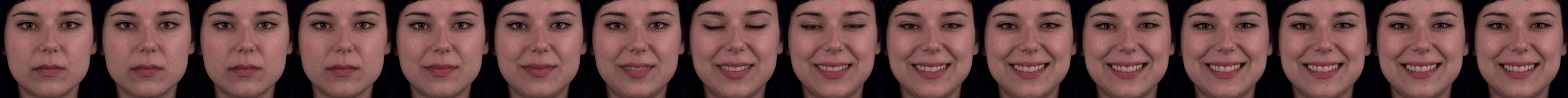}

Reenacted videos

\includegraphics[width=.8\linewidth]{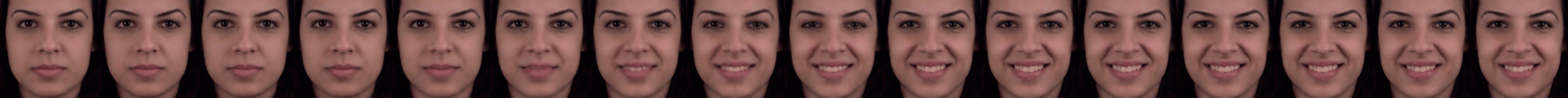}
\includegraphics[width=.8\linewidth]{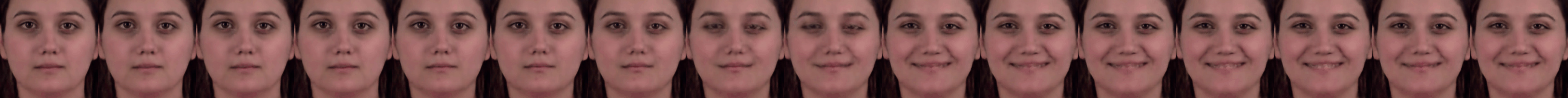}
\includegraphics[width=.8\linewidth]{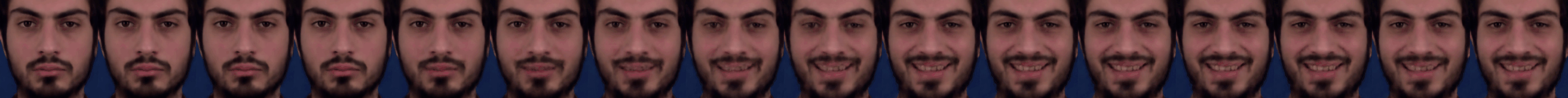}
\caption{MUG-HQ: Reenactment examples. Above: selected frames at full resolution. Below: complete sequences.}
\label{fig:hq2}
\end{figure*}

\begin{figure*}
\includegraphics[]{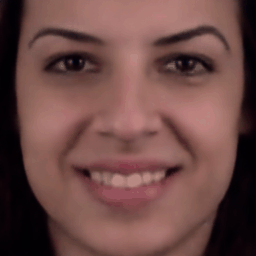}
\includegraphics[]{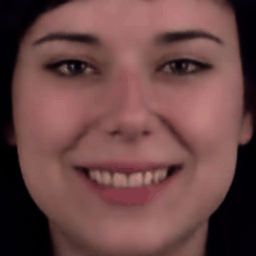}

\vspace{.8mm}
\includegraphics[]{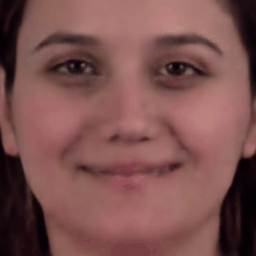}
\includegraphics[]{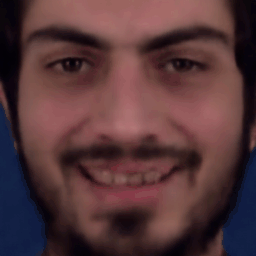}

\centering
Driving video

\includegraphics[width=.8\linewidth]{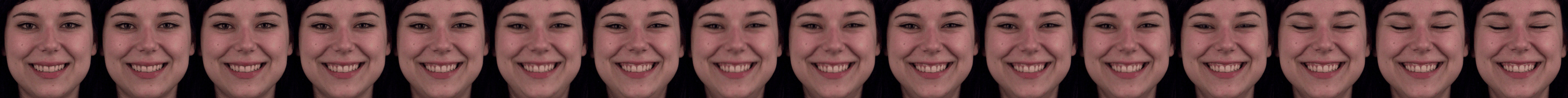}

Reenacted videos

\includegraphics[width=.8\linewidth]{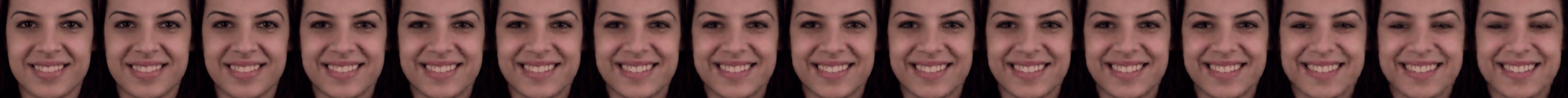}
\includegraphics[width=.8\linewidth]{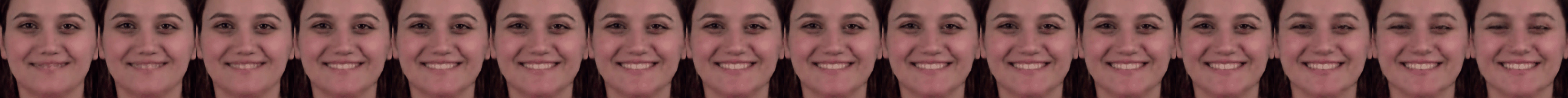}
\includegraphics[width=.8\linewidth]{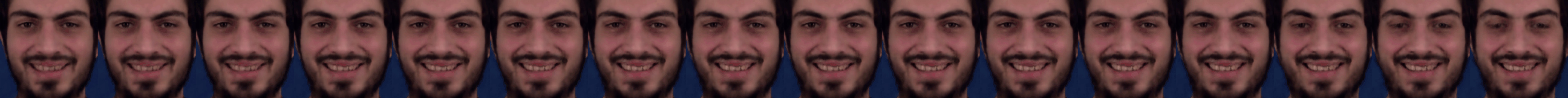}

\caption{MUG-HQ: Reenactment examples. Above: selected frames at full resolution. Below: complete sequences.}
\label{fig:hq3}
\end{figure*}

\begin{figure*}
\includegraphics[]{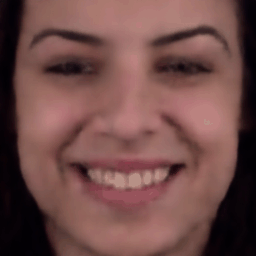}
\includegraphics[]{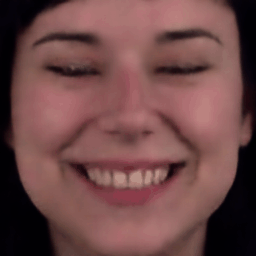}

\vspace{.8mm}
\includegraphics[]{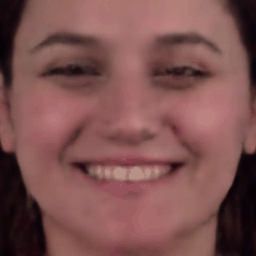}
\includegraphics[]{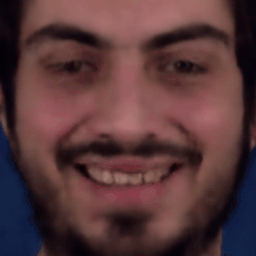}

\centering
Driving video

\includegraphics[width=.8\linewidth]{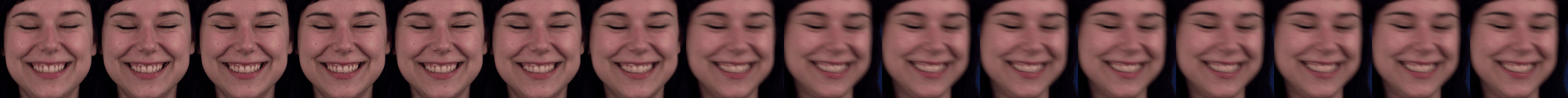}

Reenacted videos

\includegraphics[width=.8\linewidth]{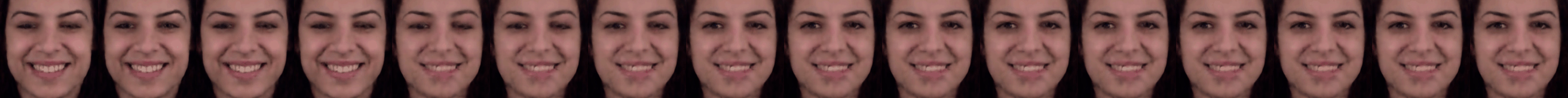}
\includegraphics[width=.8\linewidth]{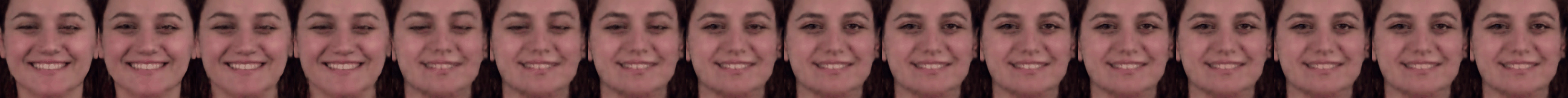}
\includegraphics[width=.8\linewidth]{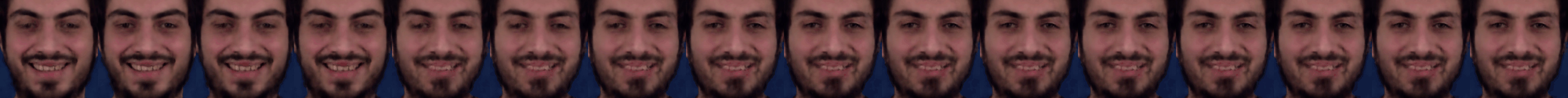}

\caption{MUG-HQ: Reenactment examples. Above: selected frames at full resolution. Below: complete sequences.}
\label{fig:hq4}
\end{figure*}

\begin{figure*}
\includegraphics[]{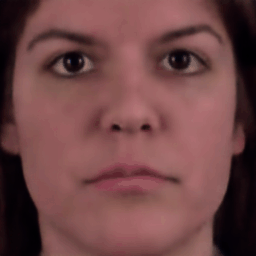}
\includegraphics[]{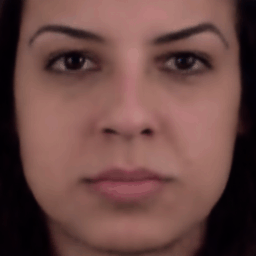}

\vspace{.8mm}
\includegraphics[]{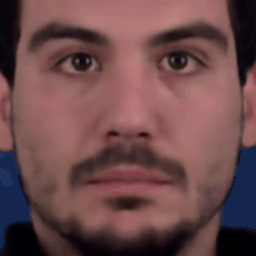}
\includegraphics[]{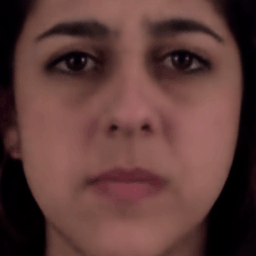}

\centering
Driving video

\includegraphics[width=\linewidth]{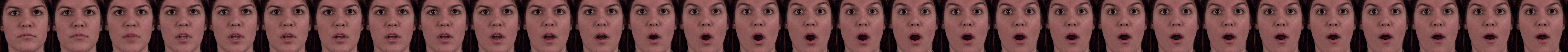}

Reenacted videos

\includegraphics[width=\linewidth]{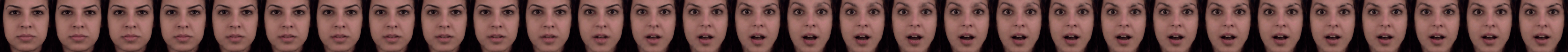}
\includegraphics[width=\linewidth]{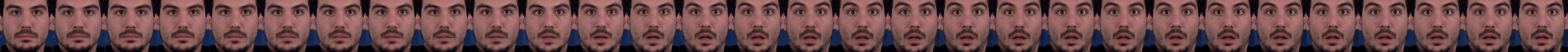}
\includegraphics[width=\linewidth]{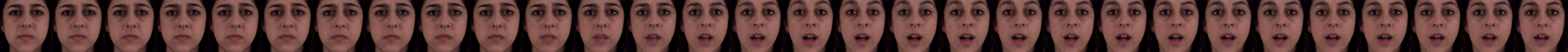}

\caption{MUG-HQ: Reenactment examples. Above: selected frames at full resolution. Below: complete sequences.}
\label{fig:hq5}
\end{figure*}

\begin{figure*}
\includegraphics[]{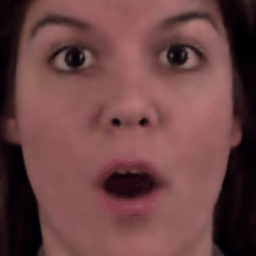}
\includegraphics[]{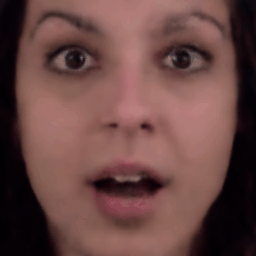}

\vspace{.8mm}
\includegraphics[]{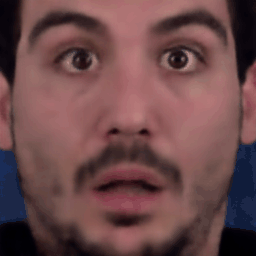}
\includegraphics[]{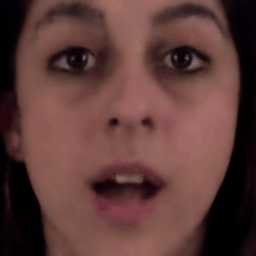}

\centering
Driving video

\includegraphics[width=\linewidth]{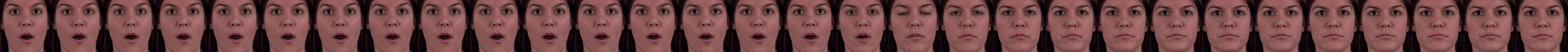}

Reenacted videos

\includegraphics[width=\linewidth]{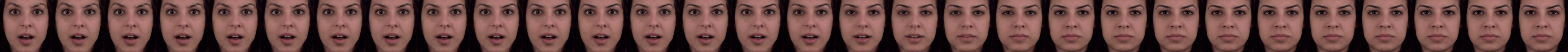}
\includegraphics[width=\linewidth]{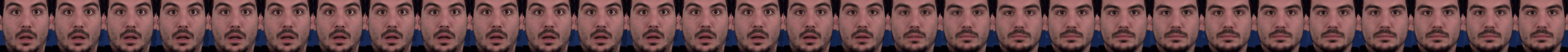}
\includegraphics[width=\linewidth]{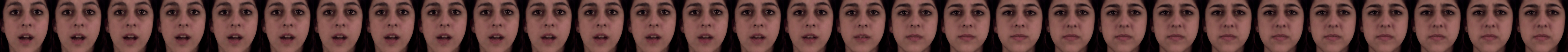}

\caption{MUG-HQ: Reenactment examples. Above: selected frames at full resolution. Below: complete sequences.}
\label{fig:hq6}
\end{figure*}

\begin{figure*}
\includegraphics[]{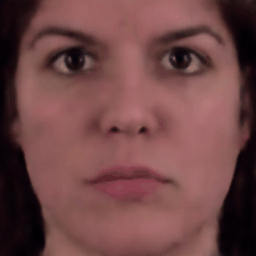}
\includegraphics[]{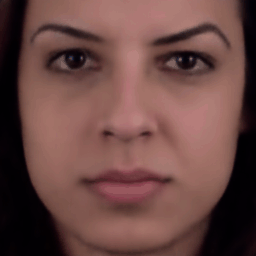}

\vspace{.8mm}
\includegraphics[]{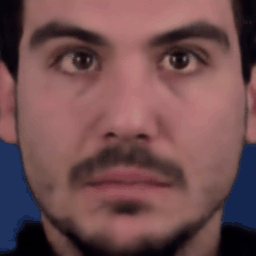}
\includegraphics[]{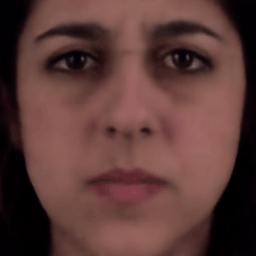}

\centering
Driving video

\includegraphics[width=\linewidth]{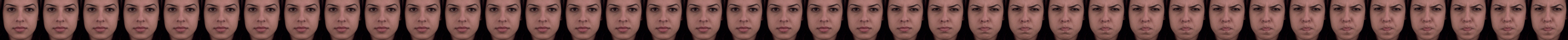}

Reenacted videos

\includegraphics[width=\linewidth]{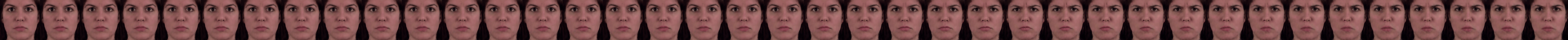}
\includegraphics[width=\linewidth]{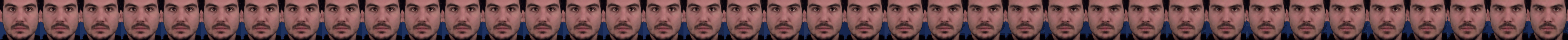}
\includegraphics[width=\linewidth]{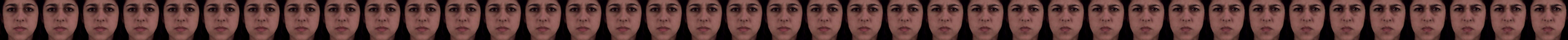}

\caption{MUG-HQ: Reenactment examples. Above: selected frames at full resolution. Below: complete sequences.}
\label{fig:hq7}
\end{figure*}

\begin{figure*}
\includegraphics[]{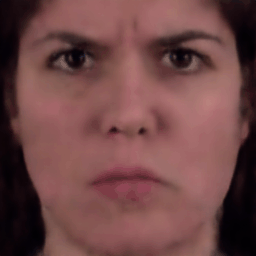}
\includegraphics[]{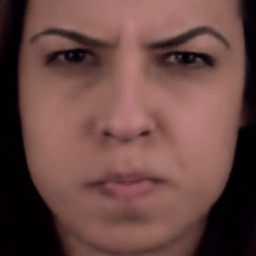}

\vspace{.8mm}
\includegraphics[]{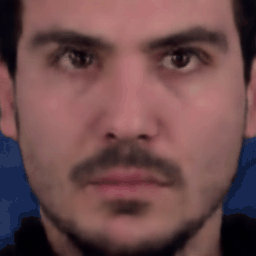}
\includegraphics[]{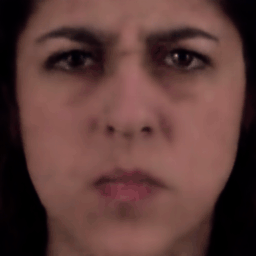}

\centering
Driving video

\includegraphics[width=\linewidth]{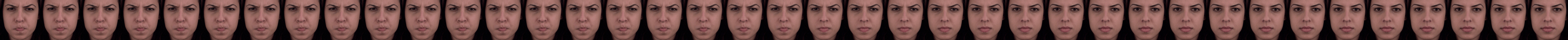}

Reenacted videos

\includegraphics[width=\linewidth]{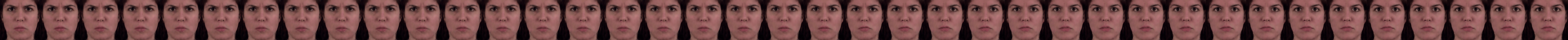}
\includegraphics[width=\linewidth]{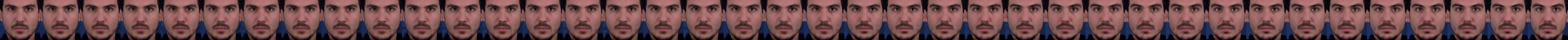}
\includegraphics[width=\linewidth]{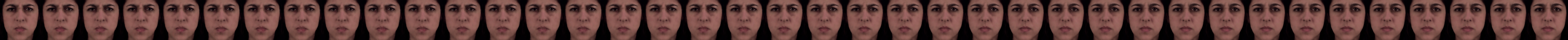}

\caption{MUG-HQ: Reenactment examples. Above: selected frames at full resolution. Below: complete sequences.}
\label{fig:hq8}
\end{figure*}

Figures~\ref{fig:hq1} to~\ref{fig:hq8} show some examples of how successful was the reenactment task in yielding realistic videos with accurate poses.

\section{Experiments on the Tai Chi Dataset}
\label{sec:taichi}

The Tai Chi dataset consists of $1191$ sequences downloaded from YouTube of several Tai Chi movements in diverse scenarios.
The videos are cropped and aligned, in such a way that the character occupies the most of the frame an remains in the center.
We performed a set of experiments on this dataset, in order to test the limitations of MTC-VAE in reconstructing high-complexity real-world scenes.

That being said, we expect the performance of MTC-VAE to fall behind VR-SotA models~\cite{Chan2019, Liu2019, Zhou2019, Liu2019tg, Aberman2019,Yang2020,Bansal2018,Siarohin2019,Siarohin2019nips,Siarohin2021,Zhao2018,Xie2020,Weng2019}, given that such models rely on high-dimensional structured representations that preserve spatial information, while our model, aiming at providing a meaningful and disentangled low-dimensional representation, has an important disadvantage, as it cannot preserve spatial information so accurately.

Table~\ref{tab:taichi} shows the comparison of the performance of MTC-VAE \wrt the baselines.
Given that the ground truth of the dataset only provide identity (\ie, content) labels, it is not possible to calculate the disentanglement metrics (FVAE, MIG and SAP).
Hence, we only report the SSIM and FID metrics, besides the accuracy on content classification.
Our model outperforms the others in realism (FID) and loses to $\beta$-TCVAE on structural similarity (SSIM).
Finally, the features yielded by MTC-VAE significantly outperforms the baselines' when used to classify the identity of the character.

\begin{table}[tb]
  \caption{Performance for content-motion disentanglement and data realism in the Tai Chai dataset}
  \label{tab:taichi}
  \scriptsize
  \centering
  \begin{tabular}{lSS[table-format=3.2(4)]S}
    \toprule
    & {SSIM $\uparrow$} & {FID $\downarrow$} & {Content Class. Acc. $\uparrow$} \\
    \midrule
    $\beta$-TCVAE & \bf .81(4) &     244.77(385) &     .93(2) \\
    dis-VAE       &     .73(5) &     215.51(241) &     .88(4) \\
    SVG-LP        &     .69(7) &     201.82(091) &     .48(4) \\
    MTC-VAE       &     .78(2) & \bf 183.24(117) & \bf .98(2) \\
    \bottomrule
  \end{tabular}

\end{table}

\begin{figure}
\centering
\includegraphics[width=\columnwidth]{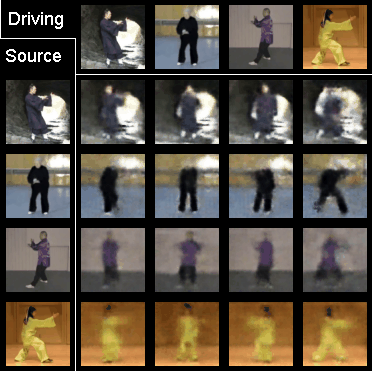}
\caption{Tai Chi: Reenactment Examples.}
\label{fig:taichi1}
\end{figure}

\begin{figure}
\centering
\includegraphics[width=\columnwidth]{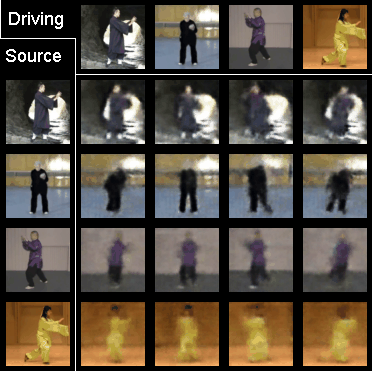}
\caption{Tai Chi: Reenactment Examples.}
\label{fig:taichi2}
\end{figure}

\begin{figure}
\centering
\includegraphics[width=\columnwidth]{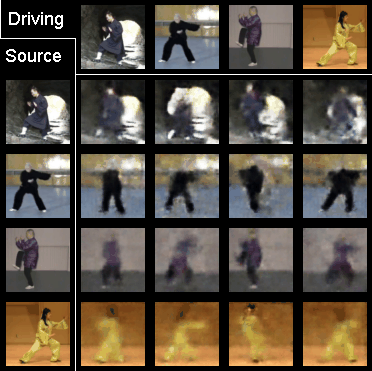}
\caption{Tai Chi: Reenactment Examples.}
\label{fig:taichi3}
\end{figure}

\begin{figure}
\centering
\includegraphics[width=\columnwidth]{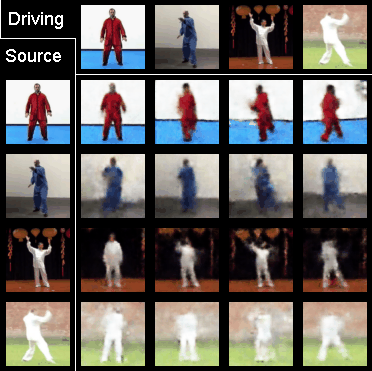}
\caption{Tai Chi: Reenactment Examples.}
\label{fig:taichi4}
\end{figure}

\begin{figure}
\centering
\includegraphics[width=\columnwidth]{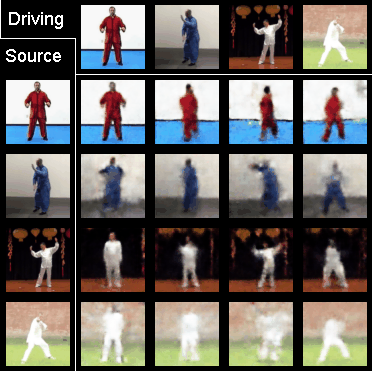}
\caption{Tai Chi: Reenactment Examples.}
\label{fig:taichi5}
\end{figure}

\begin{figure}
\centering
\includegraphics[width=\columnwidth]{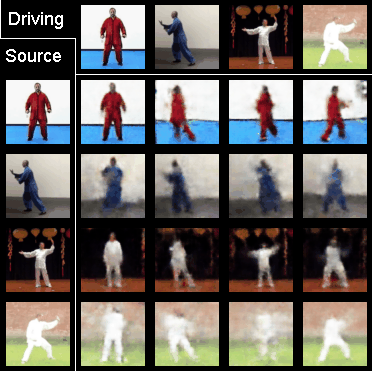}
\caption{Tai Chi: Reenactment Examples.}
\label{fig:taichi6}
\end{figure}

\begin{figure}
\centering
\includegraphics[width=\columnwidth]{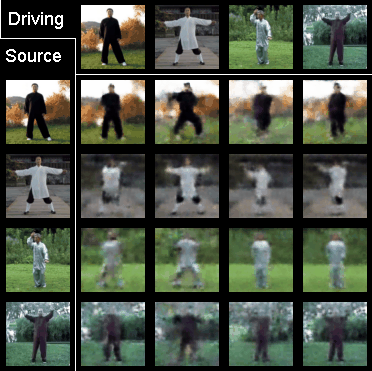}
\caption{Tai Chi: Reenactment Examples.}
\label{fig:taichi7}
\end{figure}

\begin{figure}
\centering
\includegraphics[width=\columnwidth]{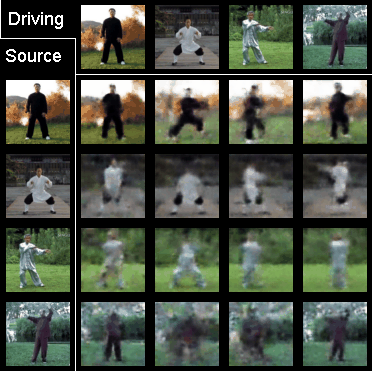}
\caption{Tai Chi: Reenactment Examples.}
\label{fig:taichi8}
\end{figure}

\begin{figure}
\centering
\includegraphics[width=\columnwidth]{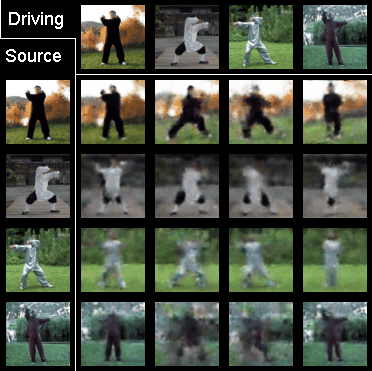}
\caption{Tai Chi: Reenactment Examples.}
\label{fig:taichi9}
\end{figure}

\begin{figure}
\centering
\includegraphics[width=\columnwidth]{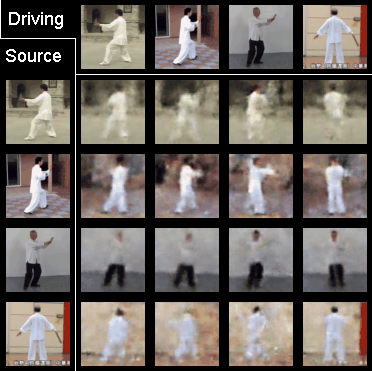}
\caption{Tai Chi: Reenactment Examples.}
\label{fig:taichi10}
\end{figure}

\begin{figure}
\centering
\includegraphics[width=\columnwidth]{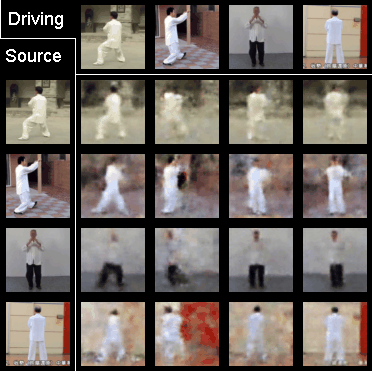}
\caption{Tai Chi: Reenactment Examples.}
\label{fig:taichi11}
\end{figure}

\begin{figure}
\centering
\includegraphics[width=\columnwidth]{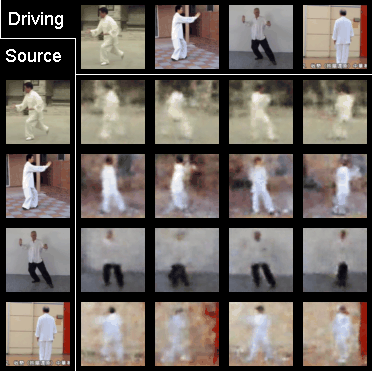}
\caption{Tai Chi: Reenactment Examples.}
\label{fig:taichi12}
\end{figure}

Figures~\ref{fig:taichi1} to~\ref{fig:taichi12} show examples of VR by MTC-VAE, and confirm our expectation of our model not being competitive when compared to SotA methods, due to the reasons presented above.
It is important to mention that, besides the complexity of the motions in the video, the highly heterogeneous backgrounds significantly hinders the reconstruction task.

However, Figures~\ref{fig:taichi1} to~\ref{fig:taichi12} allow us to qualitatively assess the disentanglement performance of our model.
Notice how the appearance is preserved in each row of the matrices of images, while the only trait that changes is the instantaneous pose (\ie, motion).
Although blurry, it is possible to see that the overall deformation of the body to yield a pose is, at some extent, correctly transferred \wrt the driving video, and that the identity of the character as well as the background (\ie, content) is preserved, meaning that both the content and motion representations have the correct meaningful information to reconstruct the video, and the bottleneck in the reconstruction process is in the decoder.

Solutions to handle this problem include explicitly modeling the background (\ie, having identity, motion, and background representations), and using deformations modules based on Spatial Transformer Networks~\cite{Jaderberg2015}.
Such solutions are considered as promising future work, but outside of the scope of our proposal in this manuscript.

\section{Detailed Quantitative Results}
\label{sec:detailed_results}

We present the performance of MTC-VAE and the baselines for each the soft generalization and the two hard generalization scenarios, \wrt the three disentanglement metrics introduced in Section~\ref{sec:experiments_disentanglement} and the two reconstruction metrics introduced in Section~\ref{sec:reenactment}.

\begin{table}[tb]
\caption{Ablation on the chunk size.}
\label{tab:cs}
\centering
\tiny
\begin{tabular}{lrSSSSS[table-format=3.2(3)]}
\toprule
& $c$ & {FVAE $\uparrow$} & {MIG $\uparrow$} & {SAP $\uparrow$} & {SSIM $\uparrow$} & {FID $\downarrow$} \\
\midrule
\multirow{5}{*}{3dShapes}
& 1 &     .50(1) &     .01(00) &     .39(11) & \bf .73(2) & \bf 100.80(4682) \\
& 3 & \bf .51(1) &     .01(00) & \bf .41(10) & \bf .73(3) &     114.23(4763) \\
& 5 &     .50(2) &     .01(00) & \bf .41(10) &     .67(6) &     119.47(5100) \\
& 7 &     .48(0) &     .01(00) &     .32(09) &     .62(7) &     117.88(5690) \\
& 9 & \bf .51(0) &     .01(00) &     .33(07) &     .58(8) &     124.43(5712) \\
\midrule
\multirow{5}{*}{CK+}
& 1 &     .85(2) & \bf .03(01) &     .05(02) & \bf .68(13) &     76.16(1937) \\
& 3 & \bf .87(3) & \bf .03(01) &     .11(03) &     .67(12) &     70.56(1992) \\
& 5 &     .86(4) &     .02(01) & \bf .13(04) &     .66(12) &     63.13(2250) \\
& 7 &     .85(4) &     .02(01) &     .12(05) &     .62(10) & \bf 63.03(1815) \\
& 9 &     .81(5) & \bf .03(01) &     .11(03) &     .61(10) &     60.56(1775) \\
\midrule
\multirow{5}{*}{dSprites}
& 1 & \bf .92(1) &     .02(2) &     .01(00) &     .77(1) &    105.79(586) \\
& 3 &     .89(3) &     .02(2) & \bf .06(05) & \bf .80(4) &     78.40(2319) \\
& 5 &     .82(1) & \bf .03(1) & \bf .06(01) &     .78(0) & \bf 68.48(378) \\
& 7 &     .79(3) &     .02(1) &     .05(01) &     .78(0) &     71.17(345) \\
& 9 &     .76(1) & \bf .03(1) &     .10(01) &     .78(0) &     72.77(282) \\
\midrule
\multirow{5}{*}{LPC}
& 1 &     .86(1) &     .00(0) &     .11(03) & \bf .67(1) & \bf 42.59(409) \\
& 3 &     .88(4) & \bf .02(6) & \bf .24(23) & \bf .67(1) &     47.40(802) \\
& 5 &     .87(1) &     .00(0) &     .17(01) &     .66(1) &     84.87(781) \\
& 7 &     .87(1) &     .00(0) &     .16(02) & \bf .67(1) &     76.92(812) \\
& 9 & \bf .89(1) &     .01(1) &     .19(03) & \bf .67(1) &     53.57(501) \\
\midrule
\multirow{5}{*}{MMNIST}
& 1 &     .91(4) &     .09(05) &     .09(04) & \bf .69(1) &    186.25(2355) \\
& 3 & \bf .95(2) &     .08(04) &     .09(05) &     .68(1) &     96.04(1128) \\
& 5 & \bf .95(4) &     .07(02) &     .10(04) &     .68(1) &    102.11(989) \\
& 7 &     .92(6) &     .06(02) &     .09(03) &     .68(1) & \bf 95.69(1354) \\
& 9 &     .91(5) & \bf .10(06) & \bf .12(06) &     .68(1) &    116.05(1338) \\
\midrule
\multirow{5}{*}{MUG}
& 1 &     .70(9) & \bf .04(2) &     .75(08) & \bf .66(6) &     43.86(1315) \\
& 3 & \bf .73(2) &     .03(2) &     .70(04) &     .62(2) &     37.62(131) \\
& 5 &     .72(3) &     .02(1) & \bf .76(10) &     .63(2) & \bf 31.18(307) \\
& 7 &     .72(5) &     .01(1) &     .72(03) &     .62(2) &     37.36(074) \\
& 9 &     .72(5) &     .02(1) &     .70(04) &     .61(2) &     37.56(106) \\
\bottomrule
\end{tabular}
\end{table}

\begin{table}[tb]
\caption{Ablation on Blind Reenactment Loss.}
\label{tab:brl}
\centering
\tiny
\begin{tabular}{lrSSSSS[table-format=3.2(3)]}
\toprule
& $\lambda$ & {FVAE $\uparrow$} & {MIG $\uparrow$} & {SAP $\uparrow$} & {SSIM $\uparrow$} & {FID $\downarrow$} \\
\midrule
\multirow{6}{*}{3dShapes}
& 0    & .50(2) & .01(1) &     .35(15) & .67(6) &     120.08(5340) \\
& 0.25 & .50(2) & .01(1) &     .36(09) & .67(6) &     120.32(5234) \\
& 0.5  & .50(2) & .01(1) &     .39(18) & .67(6) & \bf 119.84(5563) \\
& 1    & .50(2) & .01(0) & \bf .41(14) & .67(6) & \bf 119.47(5100) \\
& 2    & .50(2) & .01(1) & \bf .41(02) & .67(6) & \bf 119.34(5423) \\
& 5    & .50(2) & .01(1) &     .40(11) & .67(6) &     120.15(5194) \\
\midrule
\multirow{6}{*}{CK+}
& 0    &     .80(3) & .02(1) &     .11(6)  &     .63(11) &     71.33(2235) \\
& 0.25 &     .81(3) & .02(1) &     .11(6)  &     .64(14) &     67.73(2237)\\
& 0.5  &     .78(9) & .02(1) &     .12(8)  &     .64(03) &     64.49(2276)\\
& 1    & \bf .86(4) & .02(1) &     .13(5)  & \bf .66(12) &     63.13(2250) \\
& 2    &     .84(7) & .02(1) &     .15(4)  & \bf .66(10) & \bf 60.26(2242)\\
& 5    &     .80(4) & .02(1) & \bf .17(2)  &     .64(37) &     65.57(2284)\\
\midrule
\multirow{6}{*}{dSprites}
& 0    &     .82(9) &     .02(2) & .01(0) &     .81(0) & \bf 65.83(782) \\
& 0.25 &     .82(6) &     .02(2) & .01(0) &     .81(0) &     69.85(632) \\
& 0.5  &     .84(2) &     .02(2) & .01(0) &     .81(0) &     82.23(684) \\
& 1    & \bf .85(4) & \bf .04(2) & .01(0) & \bf .82(0) &     85.43(926) \\
& 2    & \bf .85(5) &     .04(2) & .01(0) & \bf .82(0) &     87.85(841) \\
& 5    &     .83(2) &     .01(1) & .01(0) & \bf .82(0) &     90.15(756) \\
\midrule
\multirow{6}{*}{LPC}
& 0    &     .88(3) &     .01(1) & \bf .22(04) & .67(1) &     55.81(1559)  \\
& 0.25 &     .86(3) &     .01(1) &     .22(02) & .67(1) &     50.56(373) \\
& 0.5  &     .85(8) & \bf .02(5) &     .22(01) & .67(1) &     45.27(954) \\
& 1    &     .89(1) & \bf .02(5) &     .21(19) & .67(1) & \bf 40.05(426) \\
& 2    &     .87(1) &     .01(1) &     .22(04) & .67(1) &     39.51(542) \\
& 5    & \bf .90(2) &     .01(1) &     .22(04) & .67(1) &     38.29(725) \\
\midrule
\multirow{6}{*}{MMNIST}
& 0    &     .89(4) &     .05(3) &     .08(4)  & .68(1) & \bf 100.61(1303) \\
& 0.25 &     .92(7) &     .05(5) &     .09(7)  & .68(1) &     101.11(963) \\
& 0.5  &     .93(5) & \bf .07(2) &     .09(2)  & .68(1) &     100.97(992) \\
& 1    & \bf .95(4) & \bf .07(2) & \bf .10(5)  & .68(1) &     102.11(989) \\
& 2    &     .94(2) & \bf .07(1) &     .09(1)  & .68(1) &     102.34(975) \\
& 5    &     .94(1) & \bf .07(2) &     .09(1)  & .68(1) &     102.82(951) \\
\midrule
\multirow{6}{*}{MUG}
& 0    &     .72(5) &     .01(1) &     .74(4) &     .62(2) &     34.86(632) \\
& 0.25 &     .72(5) &     .01(1) & \bf .76(2) &     .62(2) &     27.03(055) \\
& 0.5  &     .72(4) &     .01(1) & \bf .76(5) &     .62(1) & \bf 24.14(072) \\
& 1    &     .72(4) & \bf .02(2) &     .74(7) & \bf .63(3) &     29.68(076) \\
& 2    & \bf .75(7) & \bf .02(1) &     .73(5) &     .62(2) &     30.46(113) \\
& 5    &     .74(4) & \bf .02(1) &     .72(2) &     .62(1) &     32.21(098) \\
\bottomrule
\end{tabular}
\end{table}

\begin{table}[tb]
\caption{Ablation on the order of the model.}
\label{tab:o}
\centering
\tiny
\begin{tabular}{lrSSSSS[table-format=3.2(3)]}
\toprule
& $O$ & {FVAE $\uparrow$} & {MIG $\uparrow$} & {SAP $\uparrow$} & {SSIM $\uparrow$} & {FID $\downarrow$} \\
\midrule
\multirow{4}{*}{3dShapes}
& 1 &     .50(2) &     .01(0) &     .30(014) &     .64(5) & \bf 104.14(3820) \\
& 2 &     .50(2) &     .01(0) &     .41(014) & \bf .67(6) &     119.47(5100) \\
& 3 &     .50(2) &     .01(0) & \bf .46(012) & \bf .67(6) &     123.91(5284) \\
& 4 &     .50(2) &     .01(0) &     .37(012) & \bf .67(6) &     127.32(5150) \\
\midrule
\multirow{4}{*}{CK+}
& 1 &     .80(4) &     .02(01) &     .05(03) & \bf .69(12) &     65.79(2473) \\
& 2 &     .86(4) &     .02(01) &     .13(05) &     .66(12) & \bf 63.13(2250) \\
& 3 & \bf .87(3) &     .02(01) &     .13(04) &     .64(12) &     67.57(1931) \\
& 4 &     .86(4) &     .02(01) &     .06(03) &     .68(13) &     65.02(2444) \\
\midrule
\multirow{4}{*}{dSprites}
& 1 &     .86(2) &     .00(0) & \bf .12(02) & \bf .87(1) & \bf 66.29(300) \\
& 2 &     .85(6) & \bf .03(1) &     .06(03) &     .78(1) &     74.30(1584) \\
& 3 &     .89(1) &     .02(1) &     .02(01) &     .78(0) &     81.29(884) \\
& 4 & \bf .91(4) & \bf .03(3) &     .00(01) &     .78(1) &    116.81(1972) \\
\midrule
\multirow{4}{*}{LPC}
& 1 &     .87(1) &     .00(0) &     .19(03) &     .67(1) & \bf 44.70(400) \\
& 2 & \bf .88(3) & \bf .02(5) & \bf .21(22) &     .67(1) &     54.85(1806) \\
& 3 &     .86(1) &     .00(0) &     .20(01) &     .67(1) &     56.50(510) \\
& 4 &     .86(2) &     .01(0) &     .19(01) & \bf .68(1) &     57.14(470) \\
\midrule
\multirow{4}{*}{MMNIST}
& 1 &     .80(2) &     .02(02) &     .03(03) & \bf .72(3) & \bf 96.10(2398) \\
& 2 &     .95(4) &     .07(02) &     .10(05) &     .68(1) &    102.11(989) \\
& 3 & \bf .98(2) &     .07(03) &     .10(06) &     .68(1) &    100.41(1168) \\
& 4 &     .96(2) & \bf .10(05) & \bf .11(07) &     .68(1) &    103.31(965) \\
\midrule
\multirow{4}{*}{MUG}
& 1 &     .71(3) & .02(2) &     .73(08) &     .62(2) & \bf 29.30(067) \\
& 2 &     .72(5) & .02(2) &     .72(06) &     .62(3) &     38.69(605) \\
& 3 & \bf .73(4) & .02(2) &     .77(08) & \bf .64(3) &     30.34(186) \\
& 4 &     .70(2) & .02(1) & \bf .75(07) &     .63(2) & \bf 29.67(065) \\
\bottomrule
\end{tabular}
\end{table}

\begin{table}[tb]
\caption{Ablation on $\beta$.}
\label{tab:beta}
\centering
\tiny
\begin{tabular}{lrSSSSS[table-format=3.2(3)]}
\toprule
& $\lambda$ & {FVAE $\uparrow$} & {MIG $\uparrow$} & {SAP $\uparrow$} & {SSIM $\uparrow$} & {FID $\downarrow$} \\
\midrule
\multirow{2}{*}{3dShapes}
& 1 &     .50(03) &     .03(02) &     .20(02) &     .59(0) &     68.57(1095) \\
& 5 &     .50(03) &     .03(01) & \bf .25(02) & \bf .60(0) & \bf 58.01(458) \\
\midrule
\multirow{2}{*}{CK+}
& 1 &     .76(03) &     .17(03) &     .74(04) &     .64(2) &     85.62(2089) \\
& 5 & \bf .84(05) & \bf .23(03) & \bf .84(09) & \bf .71(2) & \bf 59.13(1025) \\
\midrule
\multirow{2}{*}{dSprites}
& 1 & \bf .91(02) & \bf .04(01) & \bf .10(01) &     .78(0) & \bf 57.18(643) \\
& 5 &     .87(02) &     .02(01) &     .06(01) &     .78(0) &     70.41(521) \\
\midrule
\multirow{2}{*}{LPC}
& 1 & \bf .93(06) & \bf .11(11) & \bf .60(40) &     .67(1) & \bf 41.72(331) \\
& 5 &     .87(01) &     .01(01) &     .03(01) &     .67(1) &     44.27(461) \\
\midrule
\multirow{2}{*}{MMNIST}
& 1 & \bf .96(05) & \bf .28(05) & \bf .86(01) & \bf .68(1) & \bf 103.59(557) \\
& 5 &     .56(03) &     .04(03) &     .17(11) &     .67(1) &     150.93(436) \\
\midrule
\multirow{2}{*}{MUG}
& 1 &     .72(04) &     .01(01) &     .73(05) &     .63(2) & \bf 28.79(115) \\
& 5 & \bf .74(03) & \bf .03(02) & \bf .85(05) & \bf .66(2) &     32.55(071) \\
\bottomrule
\end{tabular}
\end{table}

Tables~\ref{tab:cs}, \ref{tab:brl}, and~\ref{tab:o} shows the performance in the soft generalization scenario for our ablation studies presented in the main text.
In particular, Tables~\ref{tab:cs} and~\ref{tab:brl} have the same data as, respectively, in Figs.~\ref{fig:cs} and~\ref{fig:brl}.
The discussion on these results is provided in Section~\ref{sec:ablation}.

\begin{table}[tb]
\caption{Detailed results for the hard generalization scenarios in multiple factor disentanglement. Comparison between MTC-VAE (ours) and the baselines. (* $c=1$)}
\label{tab:detailed_comparison_mf}
\centering
\tiny
\setlength{\tabcolsep}{4pt}
\begin{tabular}{lrSSScSSS}
\toprule
& & {FVAE $\uparrow$} & {MIG $\uparrow$} & {SAP $\uparrow$} & & {FVAE $\uparrow$} & {MIG $\uparrow$} & {SAP $\uparrow$} \\
\cmidrule{3-5} \cmidrule{7-9}
& & \multicolumn{3}{c}{Appearance Holdout} & & \multicolumn{3}{c}{Motion Holdout} \\
\cmidrule{3-5} \cmidrule{7-9}
\multirow{5}{*}{3dShapes}
& $\beta$-TCVAE &     .23(04) &     .08(5) &     .03(02) & &     .20(2) &     .07(5) &     .03(2) \\
& dis-VAE       &     .19(01) &     .03(2) &     .01(01) & &     .19(0) &     .03(1) &     .01(0) \\
& SVG           &     .18(00) &     .01(0) &     .01(00) & &     .19(1) &     .01(0) &     .01(0) \\
& MTC-VAE       &     .26(02) & \bf .22(7) & \bf .09(02) & &     .27(6) & \bf .16(4) & \bf .07(1) \\
& MTC-VAE*      & \bf .30(02) &     .14(6) &     .05(03) & & \bf .30(3) &     .14(4) &     .06(2) \\
\cmidrule{3-5} \cmidrule{7-9}
\multirow{5}{*}{dSprites}
& $\beta$-TCVAE &     .33(04) &     .03(2) &     .02(02) & &     .29(2) &     .02(2) &     .01(0) \\
& dis-VAE       &     .36(03) &     .02(1) &     .01(00) & &     .39(2) &     .03(0) &     .02(0) \\
& SVG           &     .25(01) &     .00(0) &     .01(01) & &     .25(0) &     .00(0) &     .00(0) \\
& MTC-VAE       & \bf .52(01) & \bf .10(1) & \bf .08(01) & &     .38(1) &     .04(3) &     .02(1) \\
& MTC-VAE*      &     .50(05) &     .07(3) &     .05(02) & & \bf .40(2) & \bf .06(1) & \bf .05(2) \\
\cmidrule{3-5} \cmidrule{7-9}
\multirow{5}{*}{LPC}
& $\beta$-TCVAE &     .49(05) &     .11(3) &     .06(03) & &     .46(8) &     .10(4) &     .05(2) \\
& dis-VAE       &     .31(01) &     .05(1) &     .03(01) & &     .33(6) &     .05(3) &     .03(2) \\
& SVG           &     .24(01) &     .01(1) &     .01(01) & &     .23(1) &     .01(1) &     .01(1) \\
& MTC-VAE       &     .54(19) &     .15(2) &     .08(03) & &     .53(9) & \bf .20(4) &     .10(2) \\
& MTC-VAE*      & \bf .63(04) & \bf .19(2) & \bf .10(02) & & \bf .65(1) & \bf .20(5) & \bf .12(4) \\
\bottomrule
\end{tabular}
\end{table}

Table~\ref{tab:o} shows that performance on disentanglement depends on the dataset, and it can be related to the length of the videos.
\Eg, MMNIST, the dataset with the longest videos, presented better disentanglement performance at the higher orders ($O=4$), while the rest showed better performance in middle-sized orders ($O=2,3$).
On the other hand, for reconstruction, it seems that the best performance was obtained, in general, for $O=1$.
It is important to point that the memory and time required to train the model significantly increase as $O$ grows. 
We consider that having high-order models is not optimal in terms of cost-benefit.
Also, order 1 may achieve better reconstruction results, but present poorer disentanglement results. 
Optimal values of $O$ can be~$2$ or~$3$.

\begin{table}[tb]
\caption{Detailed results for the hard generalization scenarios. Comparison between MTC-VAE (ours) and the baselines, evaluating disentanglement and reconstruction. (* $c=1$)}
\label{tab:detailed_comparison}
\centering
\tiny
\setlength\tabcolsep{5pt}
\begin{tabular}{lrSSSSS[table-format=3.2(3)]}
\toprule
& & {FVAE $\uparrow$} & {MIG $\uparrow$} & {SAP $\uparrow$} & {SSIM $\uparrow$} & {FID $\downarrow$} \\
\cmidrule{3-7}
& & \multicolumn{5}{c}{Appearance Holdout} \\
\cmidrule{3-7}
\multirow{5}{*}{3dShapes}
& $\beta$-TCVAE & \bf .51(2) & \bf .01(1) &     .11(12) &     .54(14) &     183.40(3513) \\
& dis-VAE       &     .50(0) &     .00(0) &     .08(07) &     .38(4)  & \bf  83.38(1211) \\
& SVG           &     .50(1) & \bf .01(0) &     .03(04) &     .57(4)  &     219.82(2297) \\
& MTC-VAE       & \bf .51(2) & \bf .01(1) & \bf .42(15) &     .70(4)  &     174.30(2313) \\
& MTC-VAE*      &     .49(1) & \bf .01(0) &     .32(10) & \bf .75(1)  &     141.65(1833) \\
\cmidrule{3-7}
\multirow{5}{*}{CK}
& $\beta$-TCVAE &     .83(5) & \bf .03(2) &     .09(04) &     .44(1)  &     122.20(2280) \\
& dis-VAE       &     .72(1) &     .01(0) &     .05(01) & \bf .54(2)  &      74.08(405)  \\
& SVG           &     .69(6) &     .02(1) &     .04(03) &     .02(0)  & \bf  61.26(358)  \\
& MTC-VAE       & \bf .86(3) &     .02(1) & \bf .16(06) &     .50(1)  &      89.22(1263) \\
& MTC-VAE*      &     .85(3) & \bf .03(1) &     .07(03) &     .50(1)  &     100.44(598)  \\
\cmidrule{3-7}
\multirow{5}{*}{dSprites}
& $\beta$-TCVAE &     .63(5) &     .04(4) &     .00(00) & \bf .84(2)  &      96.93(974)  \\
& dis-VAE       &     .72(4) &     .01(0) &     .00(00) &     .80(1)  &     123.95(82)   \\
& SVG           &     .52(1) &     .00(0) &     .00(00) &     .78(1)  &      96.97(541)  \\
& MTC-VAE       & \bf .87(1) & \bf .06(1) & \bf .01(01) &     .81(0)  & \bf  93.47(366)  \\
& MTC-VAE*      &     .78(2) &     .03(1) &     .01(00) &     .80(1)  &     117.57(505)  \\
\cmidrule{3-7}
\multirow{5}{*}{LPC}
& $\beta$-TCVAE &     .96(2) & \bf .04(3) & \bf .05(07) &     .59(3)  &     106.71(1006) \\
& dis-VAE       &     .89(4) &     .02(1) &     .01(01) &     .47(0)  &     109.21(534)  \\
& SVG           &     .69(4) &     .01(1) &     .02(02) &     .21(0)  &     118.80(1147) \\
& MTC-VAE       & \bf .99(0) &     .02(1) &     .04(02) & \bf .70(1)  &     151.00(925)  \\
& MTC-VAE*      &     .98(2) &     .02(0) &     .04(03) &     .55(1)  & \bf  54.66(359)  \\
\cmidrule{3-7}
\multirow{5}{*}{MMNIST}
& $\beta$-TCVAE &     .63(5) &     .03(3) &     .04(04) & \bf .69(2)  &     163.29(1178) \\
& dis-VAE       &     .63(3) &     .03(1) &     .04(01) &     .68(1)  &     143.25(1362) \\
& SVG           &     .52(1) &     .01(0) &     .03(01) &     .57(1)  &     195.32(5331) \\
& MTC-VAE       & \bf .94(5) &     .08(3) & \bf .11(07) &     .68(1)  & \bf  99.18(1473) \\
& MTC-VAE*      &     .85(9) & \bf .12(5) &     .11(06) &     .65(5)  &     212.73(2281) \\
\cmidrule{3-7}

& & \multicolumn{5}{c}{Motion Holdout} \\
\cmidrule{3-7}
\multirow{5}{*}{3dShapes}
& $\beta$-TCVAE &     .50(2) & \bf .01(1) &     .11(08) &     .52(12) &    148.64(4391) \\
& dis-VAE       &     .50(0) &     .00(0) &     .06(01) &     .40(1)  & \bf 71.36(363)  \\
& SVG           &     .50(0) & \bf .01(1) &     .03(02) &     .58(1)  &    155.14(1155) \\
& MTC-VAE       &     .50(2) & \bf .01(1) &     .36(18) &     .72(0)  &    126.11(350)  \\
& MTC-VAE*      &     .50(1) & \bf .01(1) & \bf .44(15) & \bf .75(1)  &    120.44(1319) \\
\cmidrule{3-7}
\multirow{5}{*}{CK}
& $\beta$-TCVAE &     .78(6) &     .03(1) &     .04(02) &     .57(6)  &    100.63(1765) \\
& dis-VAE       &     .71(3) &     .01(1) &     .02(01) &     .65(1)  &     69.43(119)  \\
& SVG           &     .71(8) &     .02(1) &     .04(03) &     .02(1)  & \bf 22.08(669)  \\
& MTC-VAE       & \bf .89(2) &     .02(2) & \bf .08(01) &     .77(1)  &     41.03(507)  \\
& MTC-VAE*      &     .87(3) & \bf .04(1) &     .05(02) & \bf .78(2)  &     57.02(548)  \\
\cmidrule{3-7}
\multirow{5}{*}{dSprites}
& $\beta$-TCVAE &     .57(7) &     .00(0) &     .00(00) &     .80(3)  &    111.35(213)  \\
& dis-VAE       & \bf .80(6) &     .02(0) &     .00(00) &     .79(4)  &    129.26(1856) \\
& SVG           &     .52(1) &     .00(0) &     .00(00) &     .78(0)  & \bf 83.04(1778) \\
& MTC-VAE       &     .77(1) & \bf .06(2) &     .00(00) & \bf .82(0)  &    107.17(509)  \\
& MTC-VAE*      &     .75(1) &     .03(0) &     .00(00) &     .81(1)  &    161.05(3461) \\
\cmidrule{3-7}
\multirow{5}{*}{LPC}
& $\beta$-TCVAE &     .94(6) &     .04(1) &     .07(03) &     .65(4)  &    115.16(706)  \\
& dis-VAE       &     .90(3) &     .04(2) &     .02(01) &     .55(16) &    116.17(530)  \\
& SVG           &     .72(4) &     .02(1) &     .03(01) &     .23(3)  & \bf 51.82(1417) \\
& MTC-VAE       & \bf .99(0) & \bf .05(2) & \bf .09(04) & \bf .67(4)  &    150.49(1311) \\
& MTC-VAE*      &     .98(1) &     .03(2) &     .08(05) &     .59(3)  &    137.85(1014) \\
\cmidrule{3-7}
\multirow{5}{*}{MMNIST}
& $\beta$-TCVAE &     .68(7) &     .08(6) &     .06(06) & \bf .73(3)  &     134.94(1545) \\
& dis-VAE       &     .66(6) &     .03(2) &     .02(02) &     .71(1)  &     151.44(724)  \\
& SVG           &     .52(1) &     .01(0) &     .01(01) &     .59(2)  &     164.17(6363) \\
& MTC-VAE       & \bf .96(2) & \bf .16(9) & \bf .11(04) &     .68(1)  & \bf 103.55(884)  \\
& MTC-VAE*      &     .92(6) &     .10(9) &     .07(04) &     .68(2)  &     225.99(1280) \\
\bottomrule
\end{tabular}
\end{table}

Table~\ref{tab:detailed_comparison} presents the comparison of MTC-VAE and the baselines on the two hard generalization scenarios.
In general, the dominance of MTC-VAE over the baselines persists in both scenarios.

Finally, Table~\ref{tab:detailed_comparison_mf} complements Table~\ref{tab:mf}, by showing the multi-factor disentanglement performance for the hard generalization scenarios. The superiority of MTC-VAE, either in its frame and chunk version, is easily spotted.

\section{Detailed Qualitative Results}
\label{sec:detailed_pictures}

\begin{figure}[tb]
  \includegraphics[width=\columnwidth]{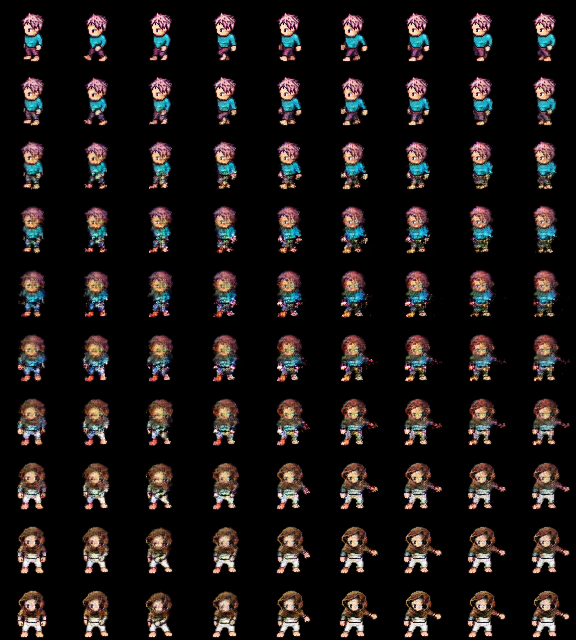}
   \caption{LPC: Latent-space traversal. Whole latent space.}
   \label{fig:traverse_full}
\end{figure}

\begin{figure}[tb]
  \includegraphics[width=\columnwidth]{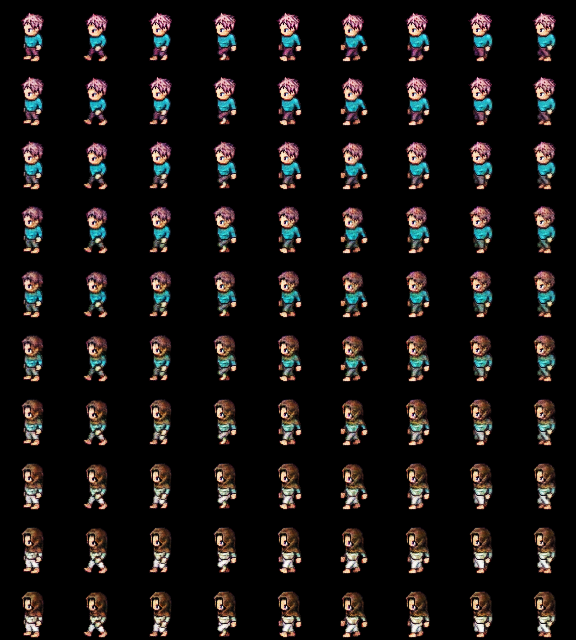}
   \caption{LPC: Latent-space traversal. Appearance latent space.}
   \label{fig:traverse_full_a}
\end{figure}

\begin{figure}[tb]
  \includegraphics[width=\columnwidth]{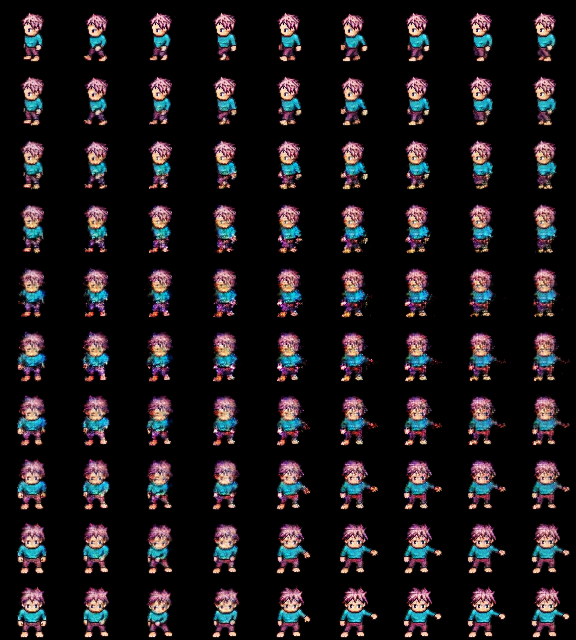}
   \caption{LPC: Latent-space traversal. Motion latent space.}
   \label{fig:traverse_full_m}
\end{figure}

\begin{figure}[tb]
  \includegraphics[width=\columnwidth]{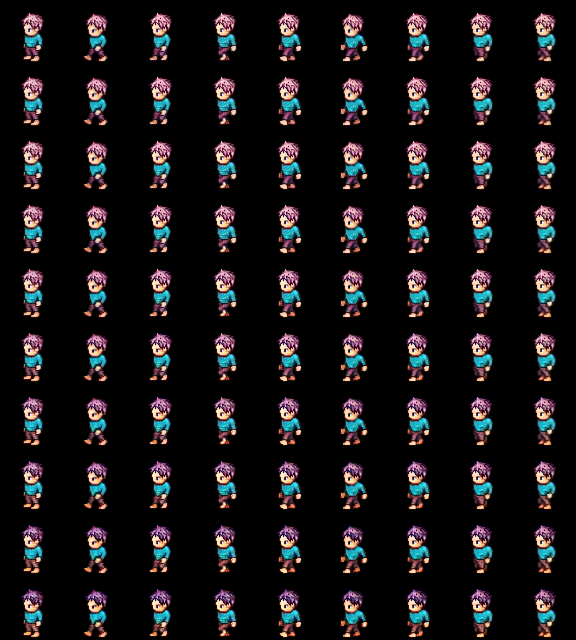}
   \caption{LPC: Latent-space traversal. Hair color controlled by unit $z_5$.}
   \label{fig:traverse_hair_color}
\end{figure}

\begin{figure}[tb]
  \includegraphics[width=\columnwidth]{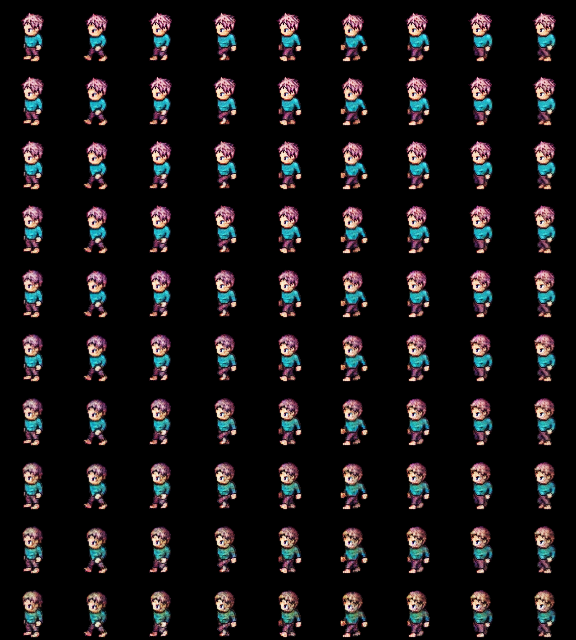}
   \caption{LPC: Latent-space traversal. Hairstyle controlled by unit $z_7$.}
   \label{fig:traverse_hair_style}
\end{figure}

\begin{figure}[tb]
  \includegraphics[width=\columnwidth]{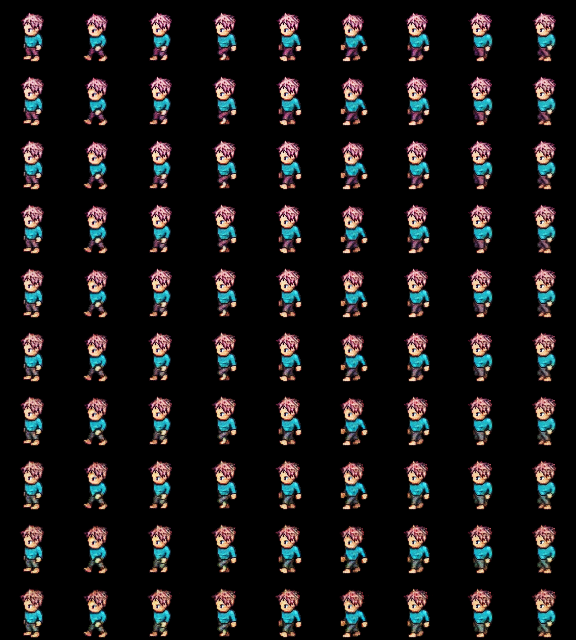}
   \caption{LPC: Latent-space traversal. Pants colors controlled by unit $z_{13}$.}
   \label{fig:traverse_pants_color_13}
\end{figure}

\begin{figure}[tb]
  \includegraphics[width=\columnwidth]{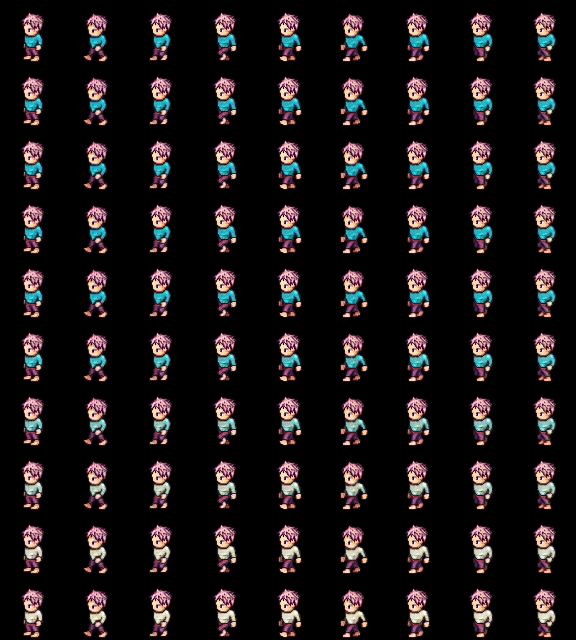}
   \caption{LPC: Latent-space traversal. Shirt color controlled by unit $z_{11}$.}
   \label{fig:traverse_shirt_color11}
\end{figure}

\begin{figure}[tb]
  \includegraphics[width=\columnwidth]{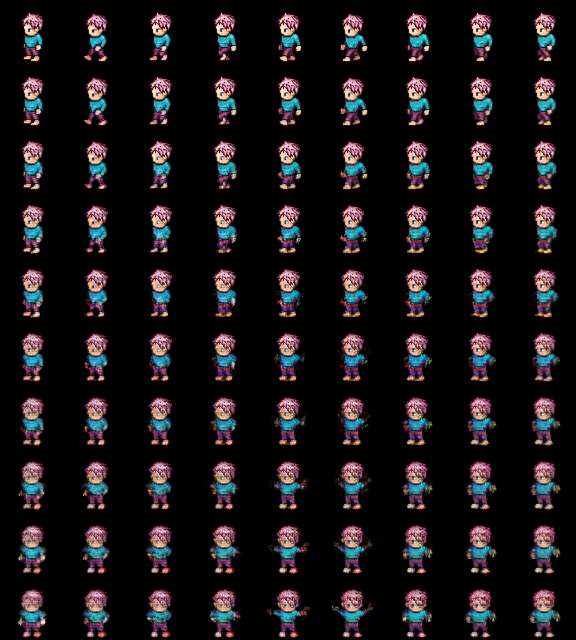}
   \caption{LPC: Latent-space traversal. Perspective controlled by units $w_0,w_4,w_6$.}
   \label{fig:traverse_perspective_14_18_20}
\end{figure}

In this section, we present traversal examples for LPC (Figs.~\ref{fig:traverse_full}, \ref{fig:traverse_full_a}, \ref{fig:traverse_full_m}, \ref{fig:traverse_hair_color}, \ref{fig:traverse_hair_style}, \ref{fig:traverse_pants_color_13}, \ref{fig:traverse_shirt_color11} and~\ref{fig:traverse_perspective_14_18_20}), reenactment examples for LPC (Figs.~\ref{fig:lpc_partial}), 3dShapes (Figs.~\ref{fig:3dShapes_appearance}, \ref{fig:3dShapes_motion}, and~\ref{fig:3dShapes_partial}), dSprites (Figs.~\ref{fig:dSprites_partial}), CK+ (Figs.~\ref{fig:CK_appearance}, \ref{fig:CK_motion}, and~\ref{fig:CK_partial}), and MMNIST (Figs.~\ref{fig:MMNIST_appearance}, \ref{fig:MMNIST_motion}, and~\ref{fig:MMNIST_partial}).
Besides the comparison with the baselines, we included examples for the ablation studies on the chunk size ($c$), impact of the Blind Reenactment Loss ($\lambda$), and the order of the model ($O$).
Recall that the default configuration for MTC-VAE (fifth line in each figure) is $c=5$, $\lambda=1$, and $O=2$.

\newcommand{\boximgc}[2]{\boximg{}{#1}{uncrop/#2}}

\begin{figure*}[tb]
\centering
  \tiny
  \setlength{\subfigsz}{.89\linewidth}
  \setlength\tabcolsep{1.5pt}
  \begin{tabular}{rc}
    Source & \boximgc{.08}{LPC_s} \\
    Driving & \boximgc{}{LPC_d} \\
    $\beta$-TCVAE & \boximgc{}{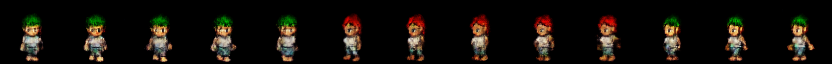} \\
    dis-VAE & \boximgc{}{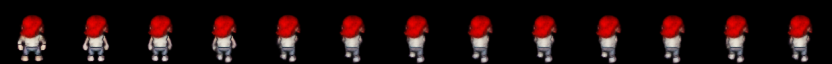} \\
    SVG & \boximgc{}{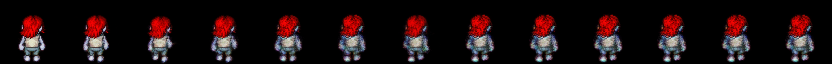} \\
    MTC-VAE & \boximgc{}{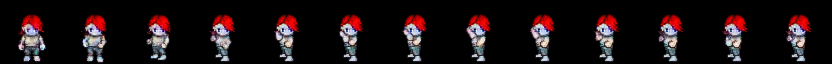} \\
    $\lambda=0$ & \boximgc{}{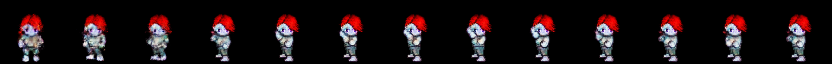} \\
    $O=1$ & \boximgc{}{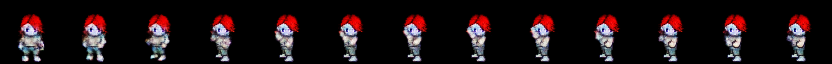} \\
    $O=3$ & \boximgc{}{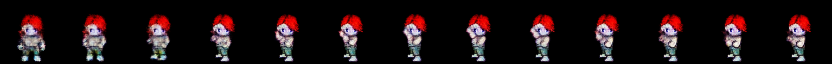} \\
    $O=4$ & \boximgc{}{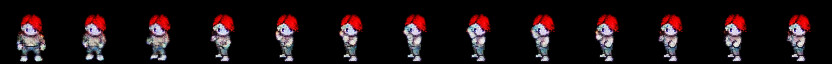} \\
    $c=1$ & \boximgc{}{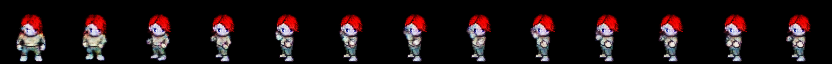} \\
    $c=5$ & \boximgc{}{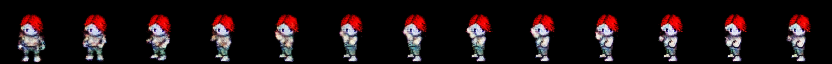} \\
    $c=7$ & \boximgc{}{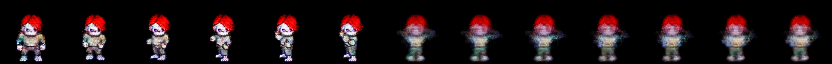} \\
    $c=9$ & \boximgc{}{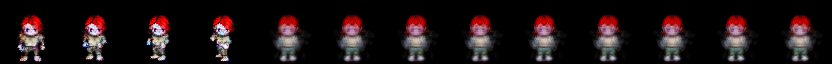}
  \end{tabular}
   \caption{LPC: examples of reenactment for the soft generalization scenario. Comparison with the baselines ($\beta$-TCVAE and dis-VAE), and ablation study on the chunk size ($c$), Blind Reenactment Loss ($\lambda$), and order of the model ($O$).}
   \label{fig:lpc_partial}
\end{figure*}

\begin{figure}[tb]
\centering
  \tiny
  \setlength{\subfigsz}{.88\linewidth}
  \setlength\tabcolsep{1.5pt}
  \begin{tabular}{rcc}
    Source  & \boximgc{.857}{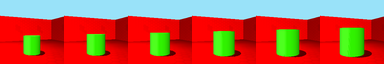} \\
    Driving & \boximgc{}{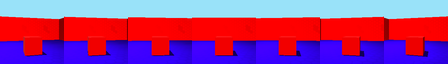} \\
    $\beta$-TCVAE & \boximgc{}{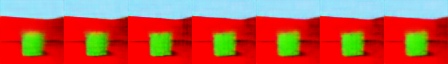} \\
    dis-VAE & \boximgc{}{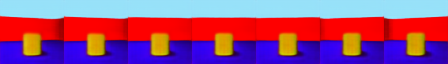} \\
    SVG & \boximgc{}{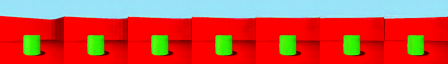} \\
    MTC-VAE & \boximgc{}{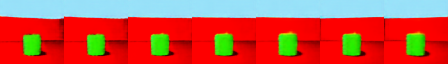} \\
    $\lambda=0$ & \boximgc{}{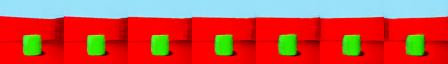} \\
    $O=1$ & \boximgc{}{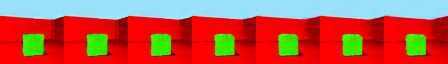} \\
    $O=3$ & \boximgc{}{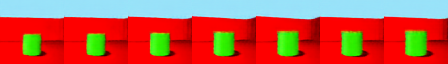} \\
    $O=4$ & \boximgc{}{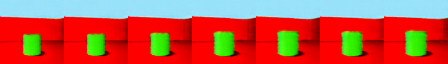} \\
    $c=1$ & \boximgc{}{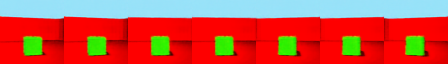} \\
    $c=3$ & \boximgc{}{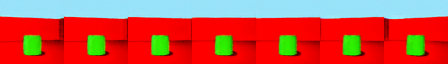} \\
    $c=7$ & \boximgc{}{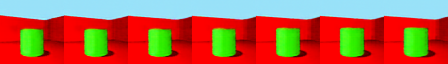} \\
    $c=9$ & \boximgc{}{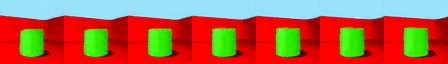}
  \end{tabular}
   \caption{3dShapes: examples of reenactment for appearance holdout. Comparison with the baselines ($\beta$-TCVAE and dis-VAE), and ablation study on the chunk size ($c$), Blind Reenactment Loss ($\lambda$), and order of the model ($O$).}
   \label{fig:3dShapes_appearance}
\end{figure}

\begin{figure}[tb]
\centering
  \tiny
  \setlength{\subfigsz}{.88\linewidth}
  \setlength\tabcolsep{1.5pt}
  \begin{tabular}{rcc}
    Source  & \boximgc{.25}{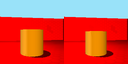} \\
    Driving & \boximgc{}{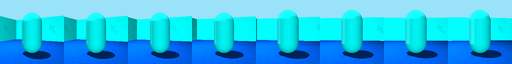} \\
    $\beta$-TCVAE & \boximgc{}{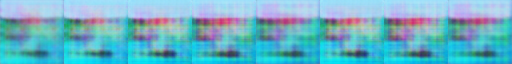} \\
    dis-VAE & \boximgc{}{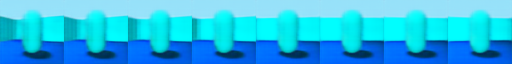} \\
    SVG & \boximgc{}{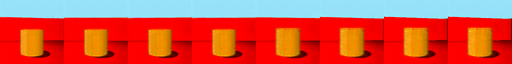} \\
    MTC-VAE & \boximgc{}{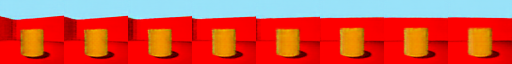} \\
    $\lambda=0$ & \boximgc{}{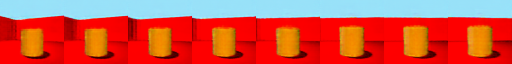} \\
    $O=1$ & \boximgc{}{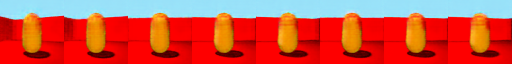} \\
    $O=3$ & \boximgc{}{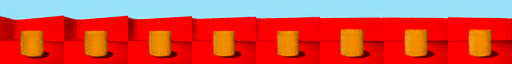} \\
    $O=4$ & \boximgc{}{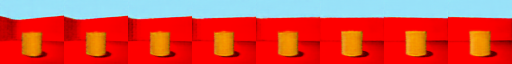} \\
    $c=1$ & \boximgc{}{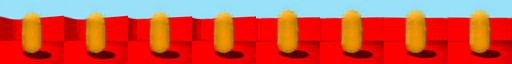} \\
    $c=3$ & \boximgc{}{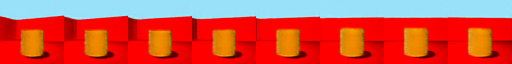} \\
    $c=7$ & \boximgc{}{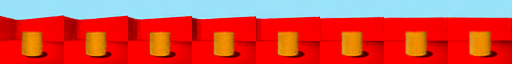} \\
    $c=9$ & \boximgc{}{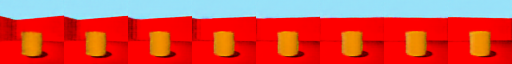}
  \end{tabular}
   \caption{3dShapes: examples of reenactment for motion holdout. Comparison with the baselines ($\beta$-TCVAE and dis-VAE), and ablation study on the chunk size ($c$), Blind Reenactment Loss ($\lambda$), and order of the model ($O$).}
   \label{fig:3dShapes_motion}
\end{figure}

\begin{figure}[tb]
\centering
  \tiny
  \setlength{\subfigsz}{.88\linewidth}
  \setlength\tabcolsep{1.5pt}
  \begin{tabular}{rcc}
    Source  & \boximgc{.667}{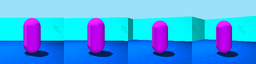} \\
    Driving & \boximgc{}{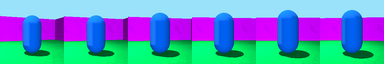} \\
    $\beta$-TCVAE & \boximgc{}{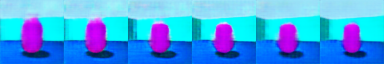} \\
    dis-VAE & \boximgc{}{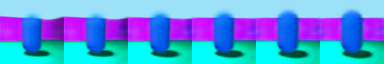} \\
    SVG & \boximgc{}{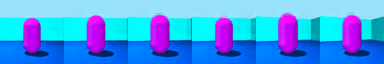} \\
    MTC-VAE & \boximgc{}{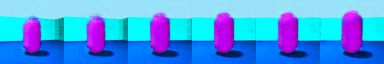} \\
    $\lambda=0$ & \boximgc{}{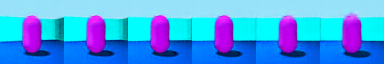} \\
    $O=1$ & \boximgc{}{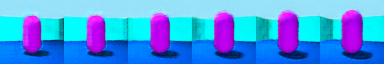} \\
    $O=3$ & \boximgc{}{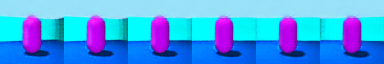} \\
    $O=4$ & \boximgc{}{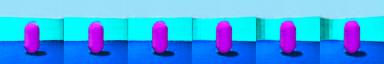} \\
    $c=1$ & \boximgc{}{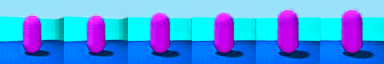} \\
    $c=3$ & \boximgc{}{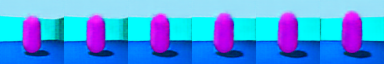} \\
    $c=7$ & \boximgc{}{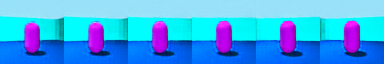} \\
    $c=9$ & \boximgc{}{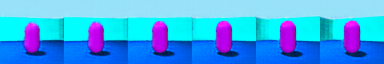}
  \end{tabular}
   \caption{3dShapes: examples of reenactment for the soft generalization scenario. Comparison with the baselines ($\beta$-TCVAE and dis-VAE), and ablation study on the chunk size ($c$), Blind Reenactment Loss ($\lambda$), and order of the model ($O$).}
   \label{fig:3dShapes_partial}
\end{figure}

\begin{figure*}
\centering
  \tiny
  \setlength{\subfigsz}{.97\linewidth}
  \setlength\tabcolsep{1.5pt}
  \begin{tabular}{rccc}
    Source & \boximgc{.04}{dSprites_s} \\
    Driving & \boximgc{}{dSprites_d} \\
    $\beta$-TCVAE & \boximgc{}{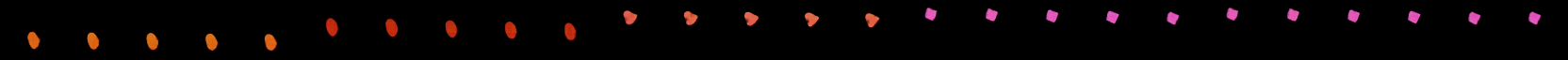} \\
    dis-VAE & \boximgc{}{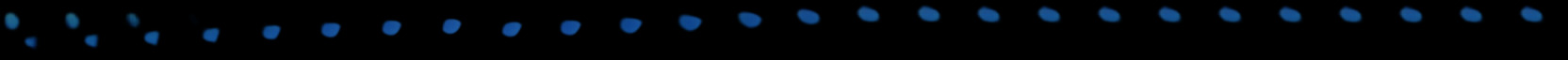} \\
    SVG & \boximgc{}{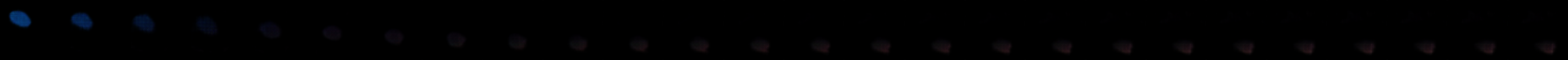} \\
    MTC-VAE & \boximgc{}{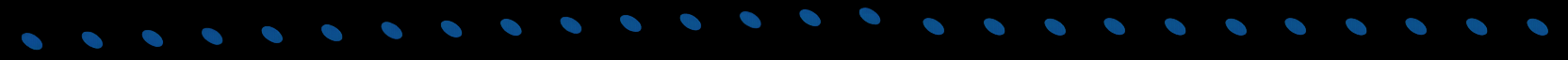} \\
    $O=1$ & \boximgc{}{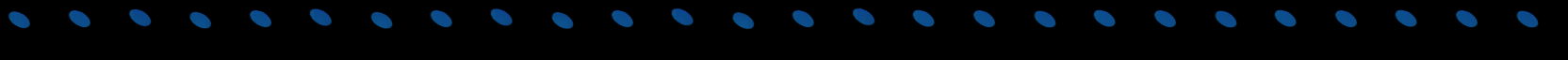} \\
    $O=3$ & \boximgc{}{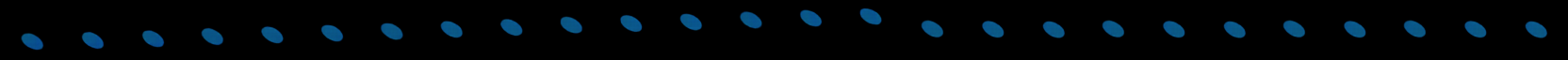} \\
    $O=4$ & \boximgc{}{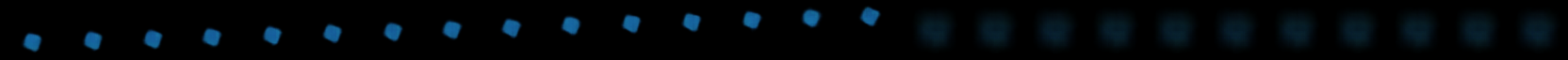} \\
    $c=1$ & \boximgc{}{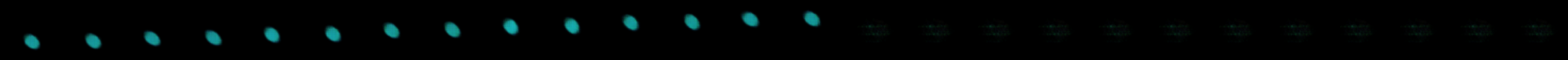} \\
    $c=5$ & \boximgc{}{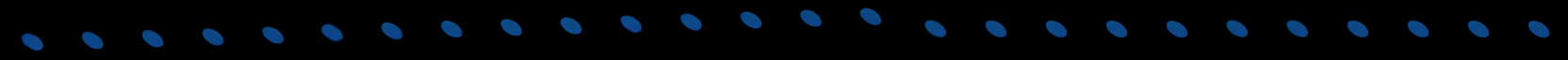} \\
    $c=7$ & \boximgc{}{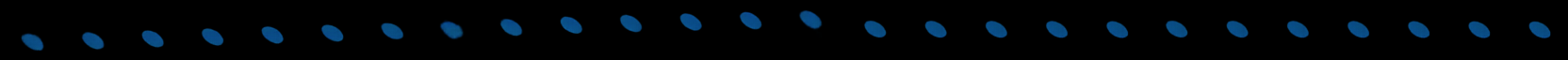} \\
    $c=9$ & \boximgc{}{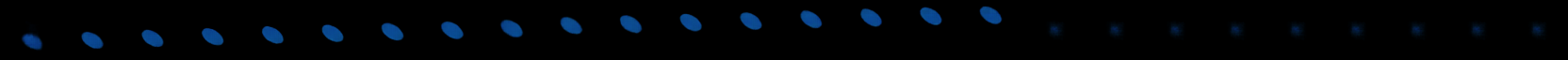}
  \end{tabular}
   \caption{dSprites: examples of reenactment for the soft generalization scenario. Comparison with the baselines ($\beta$-TCVAE and dis-VAE), and ablation study on the chunk size ($c$), Blind Reenactment Loss ($\lambda$), and order of the model ($O$).}
   \label{fig:dSprites_partial}
\end{figure*}

\begin{figure*}
\centering
  \tiny
  \setlength{\subfigsz}{.95\linewidth}
  \setlength\tabcolsep{1.5pt}
  \begin{tabular}{rc}
    Source  & \boximgc{.667}{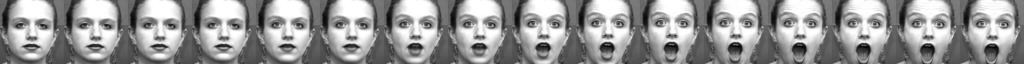} \\
    Driving & \boximgc{}{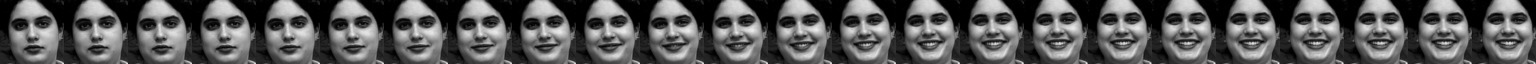} \\
    $\beta$-TCVAE & \boximgc{}{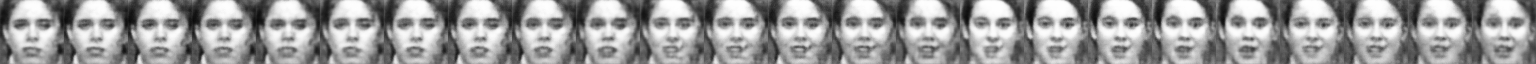} \\
    dis-VAE & \boximgc{}{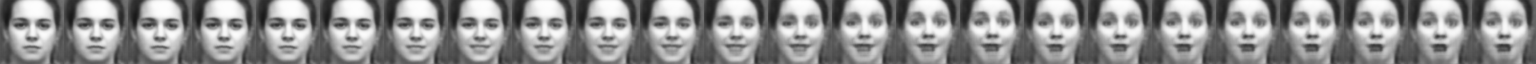} \\
    SVG & \boximgc{}{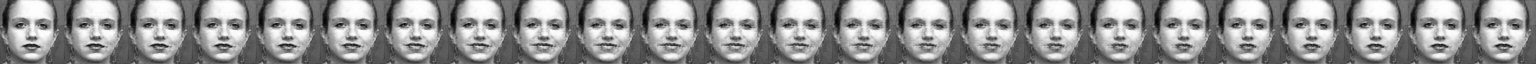} \\
    MTC-VAE & \boximgc{}{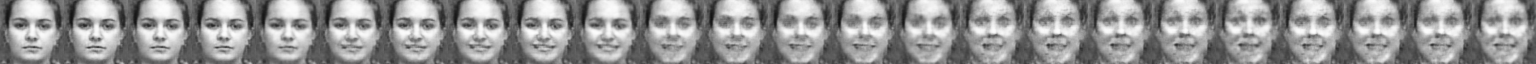} \\
    $\lambda=0$ & \boximgc{}{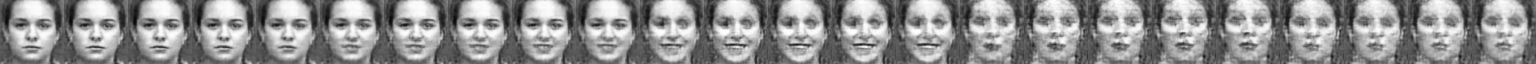} \\
    $O=1$ & \boximgc{}{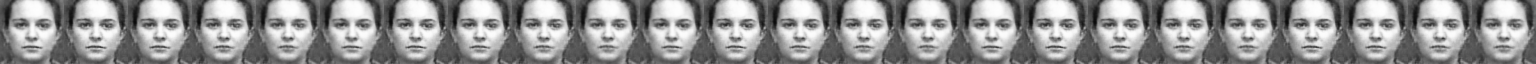} \\
    $O=3$ & \boximgc{}{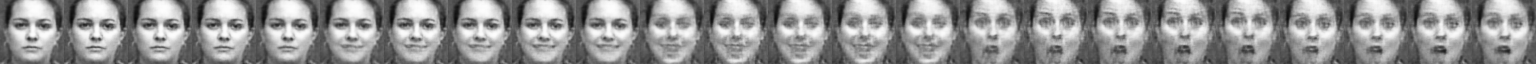} \\
    $O=4$ & \boximgc{}{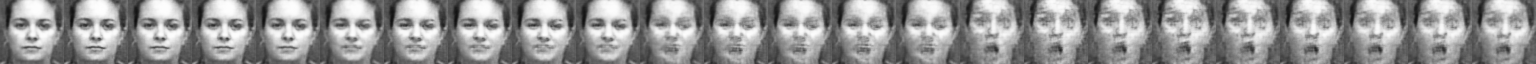} \\
    $c=1$ & \boximgc{}{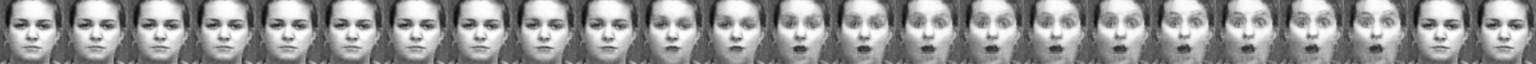} \\
    $c=3$ & \boximgc{}{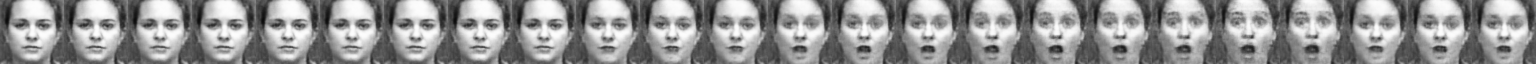} \\
    $c=7$ & \boximgc{}{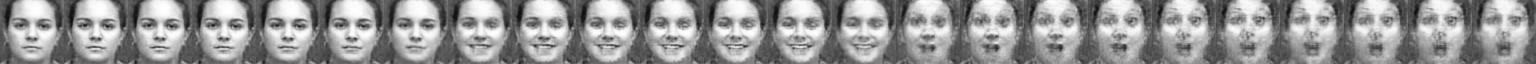} \\
    $c=9$ & \boximgc{}{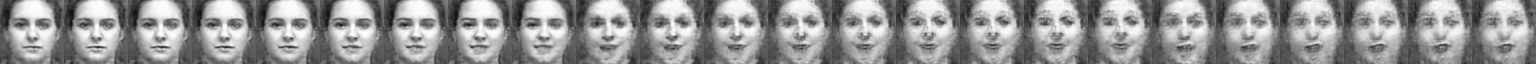}
  \end{tabular}
   \caption{CK+: examples of reenactment for appearance holdout. Comparison with the baselines ($\beta$-TCVAE and dis-VAE), and ablation study on the chunk size ($c$), Blind Reenactment Loss ($\lambda$), and order of the model ($O$).}
   \label{fig:CK_appearance}
\end{figure*}

\begin{figure*}
\centering
  \tiny
  \setlength{\subfigsz}{.95\linewidth}
  \setlength\tabcolsep{1.5pt}
  \begin{tabular}{rc}
    Source  & \boximgc{.372}{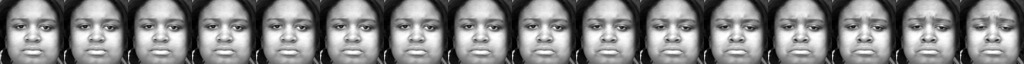} \\
    Driving & \boximgc{}{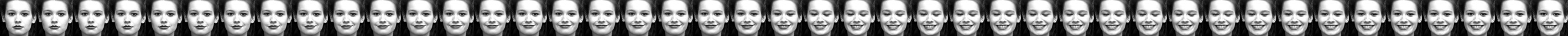} \\
    $\beta$-TCVAE & \boximgc{}{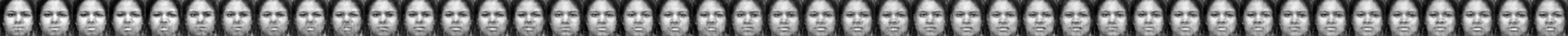} \\
    dis-VAE & \boximgc{}{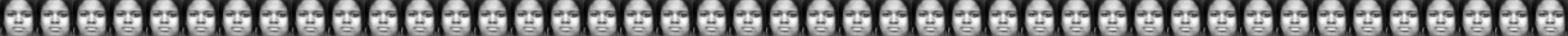} \\
    SVG & \boximgc{}{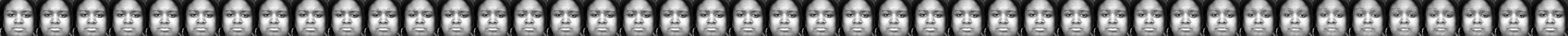} \\
    MTC-VAE & \boximgc{}{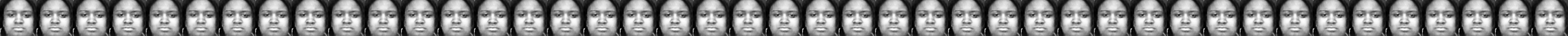} \\
    $\lambda=0$ & \boximgc{}{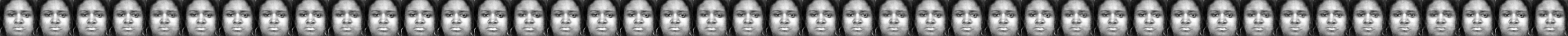} \\
    $O=1$ & \boximgc{}{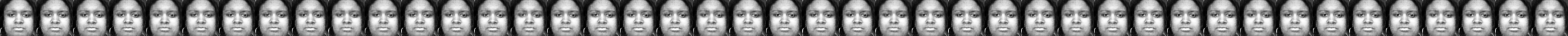} \\
    $O=3$ & \boximgc{}{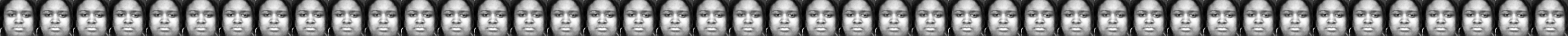} \\
    $O=4$ & \boximgc{}{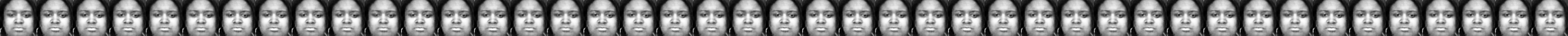} \\
    $c=1$ & \boximgc{}{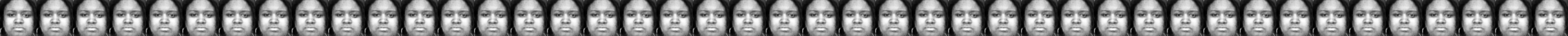} \\
    $c=3$ & \boximgc{}{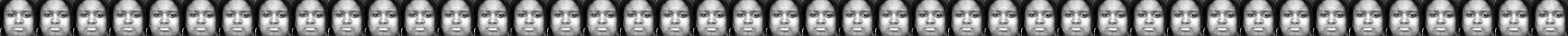} \\
    $c=7$ & \boximgc{}{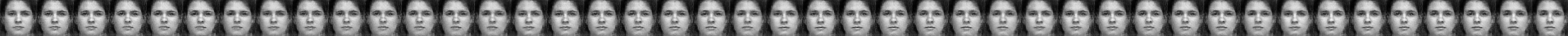} \\
    $c=9$ & \boximgc{}{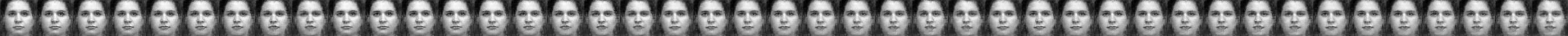}
  \end{tabular}
   \caption{CK+: examples of reenactment for motion holdout. Comparison with the baselines ($\beta$-TCVAE and dis-VAE), and ablation study on the chunk size ($c$), Blind Reenactment Loss ($\lambda$), and order of the model ($O$).}
   \label{fig:CK_motion}
\end{figure*}

\begin{figure*}
\centering
  \tiny
  \setlength{\subfigsz}{.95\linewidth}
  \setlength\tabcolsep{1.5pt}
  \begin{tabular}{rc}
    Source  & \boximgc{}{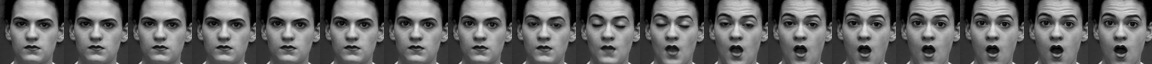} \\
    Driving & \boximgc{.889}{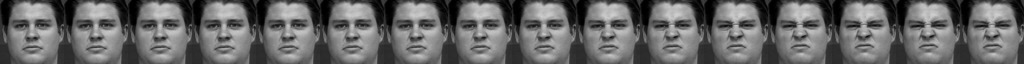} \\
    $\beta$-TCVAE & \boximgc{.889}{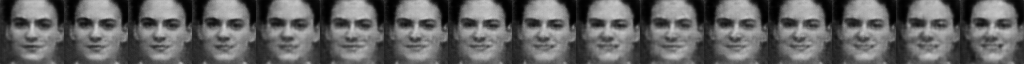} \\
    dis-VAE & \boximgc{.889}{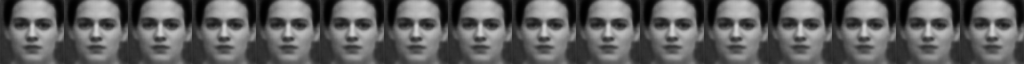} \\
    SVG & \boximgc{.889}{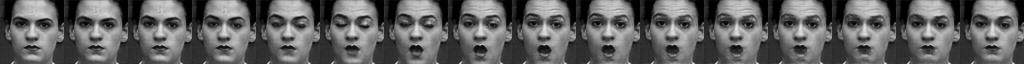} \\
    MTC-VAE & \boximgc{.889}{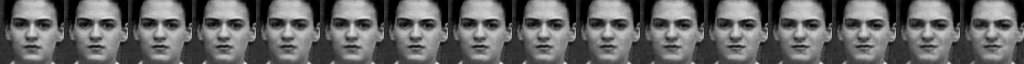} \\
    $\lambda=0$ & \boximgc{.889}{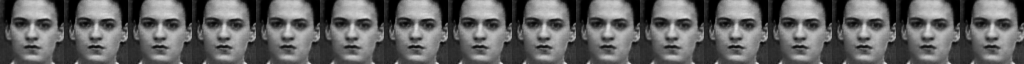} \\
    $O=1$ & \boximgc{.889}{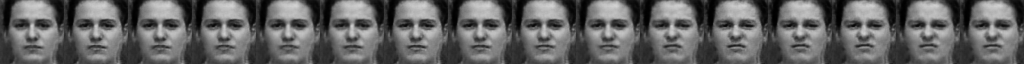} \\
    $O=3$ & \boximgc{.889}{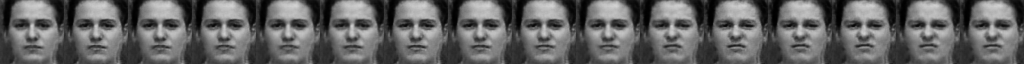} \\
    $O=4$ & \boximgc{.889}{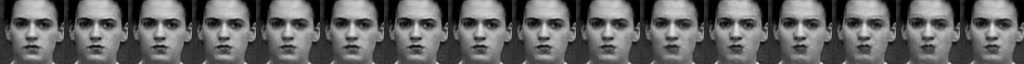} \\
    $c=1$ & \boximgc{.889}{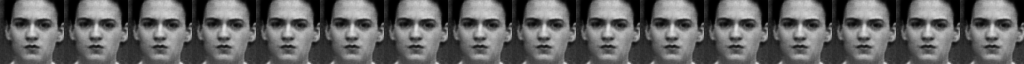} \\
    $c=3$ & \boximgc{.889}{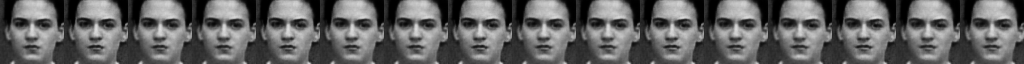} \\
    $c=7$ & \boximgc{.889}{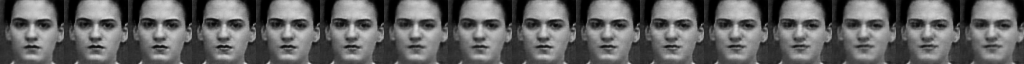} \\
    $c=9$ & \boximgc{.889}{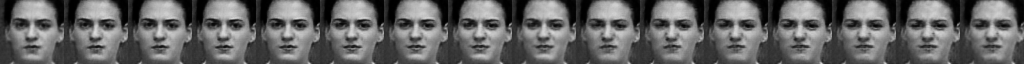}
  \end{tabular}
   \caption{CK+: examples of reenactment for the soft generalization scenario. Comparison with the baselines ($\beta$-TCVAE and dis-VAE), and ablation study on the chunk size ($c$), Blind Reenactment Loss ($\lambda$), and order of the model ($O$).}
   \label{fig:CK_partial}
\end{figure*}

\begin{figure*}
\centering
  \tiny
  \setlength{\subfigsz}{.99\linewidth}
  \setlength\tabcolsep{1.5pt}
  \begin{tabular}{rc}
    Source & \boximgc{.015}{MUG_s} \\
    Driving & \boximgc{.96}{MUG_d} \\
    $\beta$-TCVAE & \boximgc{.96}{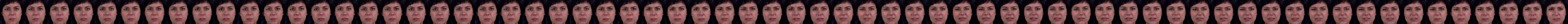} \\
    dis-VAE & \boximgc{.96}{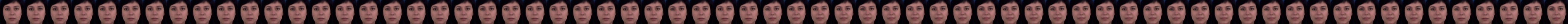} \\
    SVG & \boximgc{.96}{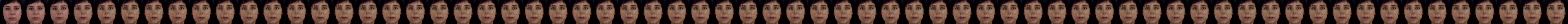} \\
    MTC-VAE & \boximgc{.96}{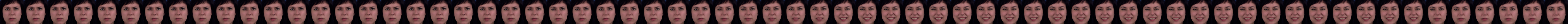} \\
    $\lambda=0$ & \boximgc{.96}{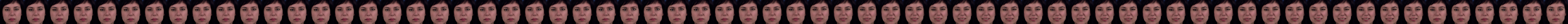} \\
    $O=1$ & \boximgc{.96}{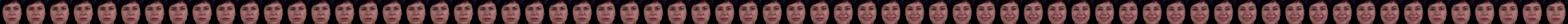} \\
    $O=2$ & \boximgc{.96}{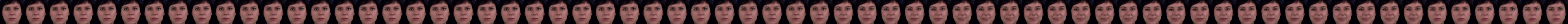} \\
    $O=4$ & \boximgc{.96}{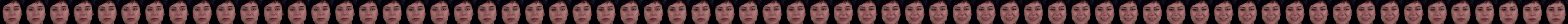} \\
    $c=1$ & \boximgc{.96}{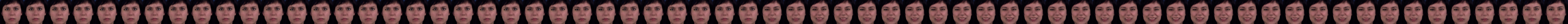} \\
    $c=3$ & \boximgc{.96}{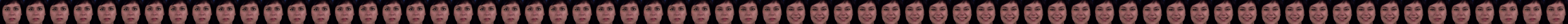} \\
    $c=7$ & \boximgc{.96}{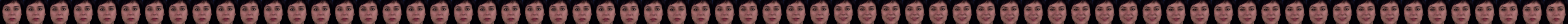} \\
    $c=9$ & \boximgc{.96}{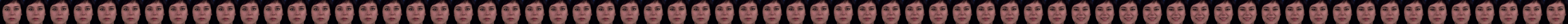}
  \end{tabular}
   \caption{MUG: examples of reenactment for the soft generalization scenario. Comparison with the baselines ($\beta$-TCVAE and dis-VAE), and ablation study on the chunk size ($c$), Blind Reenactment Loss ($\lambda$), and order of the model ($O$).}
   \label{fig:MUG_partial}
\end{figure*}

\begin{figure*}
\centering
  \tiny
  \setlength{\subfigsz}{.95\linewidth}
  \setlength\tabcolsep{1.5pt}
  \begin{tabular}{rcc}
    Source  & \boximgc{}{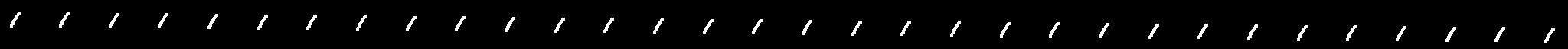} \\
    Driving & \boximgc{}{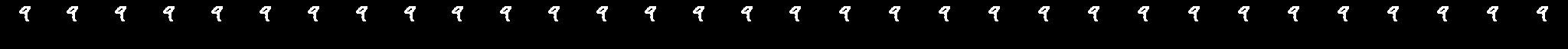} \\
    $\beta$-TCVAE & \boximgc{}{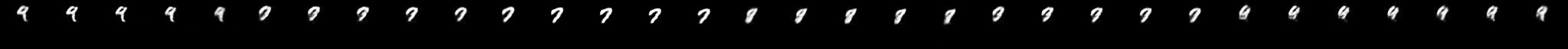} \\
    dis-VAE & \boximgc{}{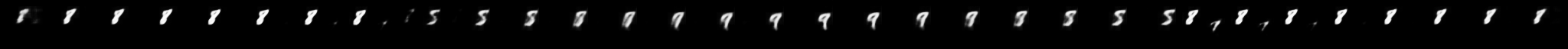} \\
    SVG & \boximgc{}{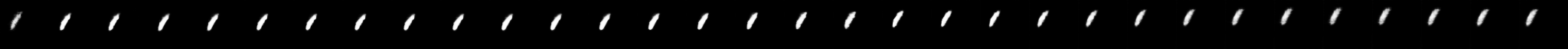} \\
    MTC-VAE & \boximgc{}{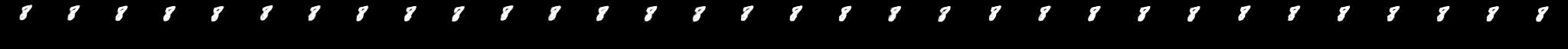} \\
    $\lambda=0$ & \boximgc{}{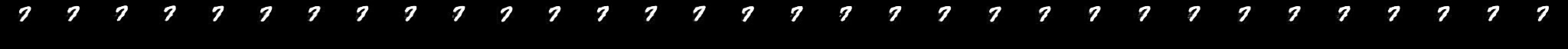} \\
    $O=1$ & \boximgc{}{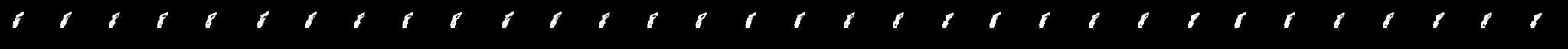} \\
    $O=3$ & \boximgc{}{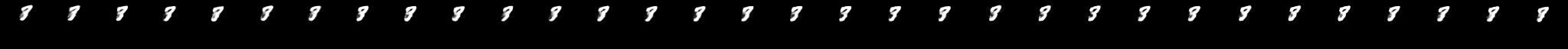} \\
    $O=4$ & \boximgc{}{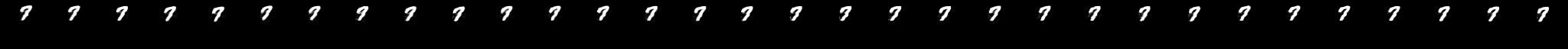} \\
    $c=1$ & \boximgc{}{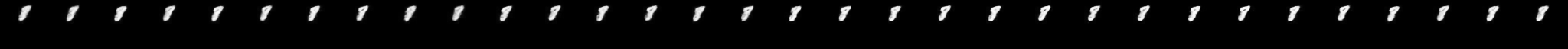} \\
    $c=3$ & \boximgc{}{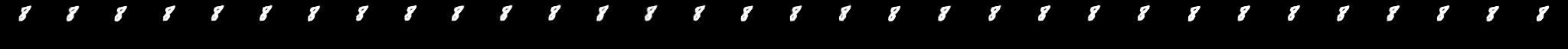} \\
    $c=7$ & \boximgc{}{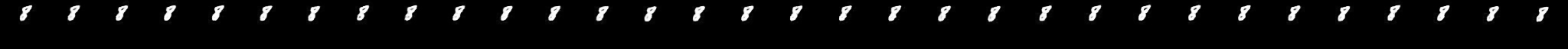} \\
    $c=9$ & \boximgc{}{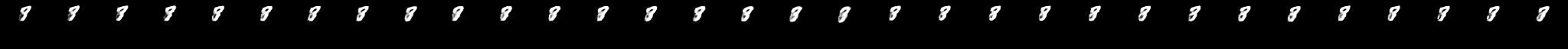}
  \end{tabular}
   \caption{MMNIST: examples of reenactment for appearance holdout. Comparison with the baselines ($\beta$-TCVAE and dis-VAE), and ablation study on the chunk size ($c$), Blind Reenactment Loss ($\lambda$), and order of the model ($O$).}
   \label{fig:MMNIST_appearance}
\end{figure*}

\begin{figure*}
\centering
  \tiny
  \setlength{\subfigsz}{.95\linewidth}
  \setlength\tabcolsep{1.5pt}
  \begin{tabular}{rcc}
    Source  & \boximgc{}{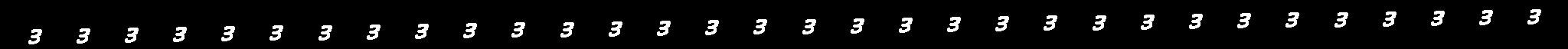} \\
    Driving & \boximgc{}{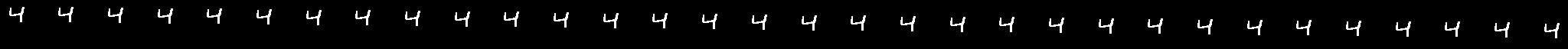} \\
    $\beta$-TCVAE & \boximgc{}{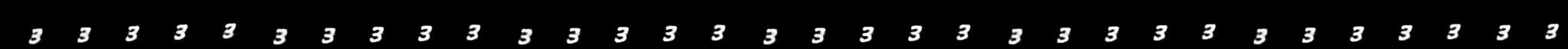} \\
    dis-VAE & \boximgc{}{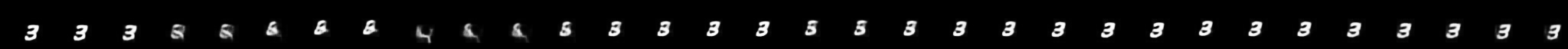} \\
    SVG & \boximgc{}{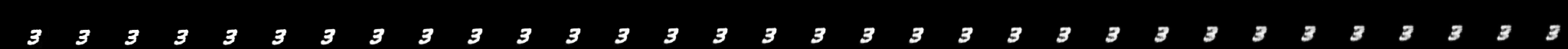} \\
    MTC-VAE & \boximgc{}{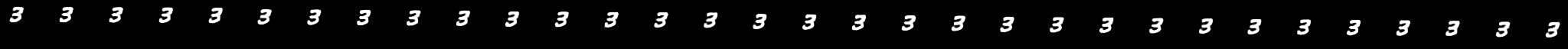} \\
    $\lambda=0$ & \boximgc{}{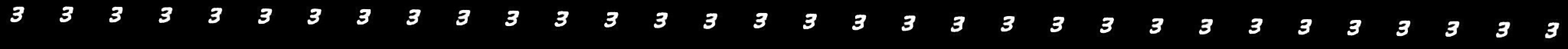} \\
    $O=1$ & \boximgc{}{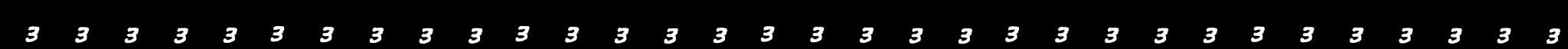} \\
    $O=3$ & \boximgc{}{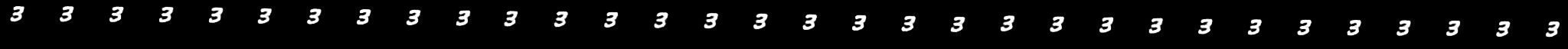} \\
    $O=4$ & \boximgc{}{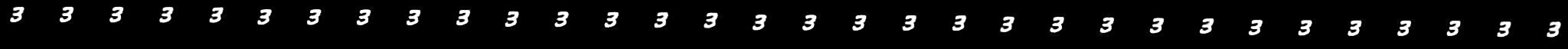} \\
    $c=1$ & \boximgc{}{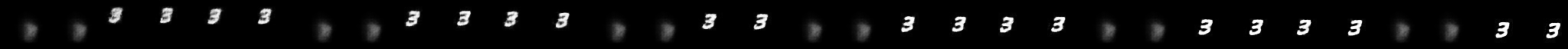} \\
    $c=3$ & \boximgc{}{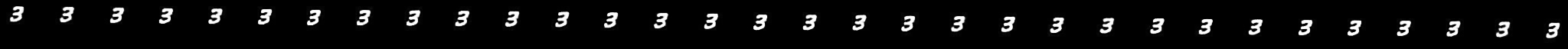} \\
    $c=7$ & \boximgc{}{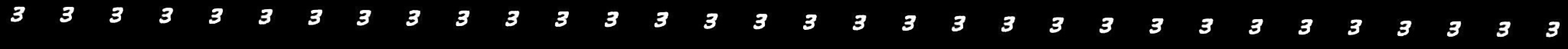} \\
    $c=9$ & \boximgc{}{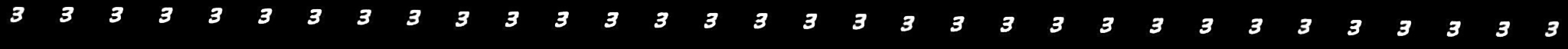}
  \end{tabular}
   \caption{MMNIST: examples of reenactment for motion holdout. Comparison with the baselines ($\beta$-TCVAE and dis-VAE), and ablation study on the chunk size ($c$), Blind Reenactment Loss ($\lambda$), and order of the model ($O$).}
   \label{fig:MMNIST_motion}
\end{figure*}

\begin{figure*}
\centering
  \tiny
  \setlength{\subfigsz}{.95\linewidth}
  \setlength\tabcolsep{1.5pt}
  \begin{tabular}{rcc}
    Source  & \boximgc{}{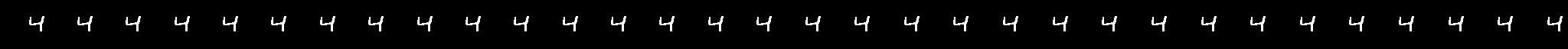} \\
    Driving & \boximgc{}{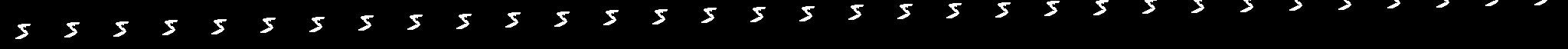} \\
    $\beta$-TCVAE & \boximgc{}{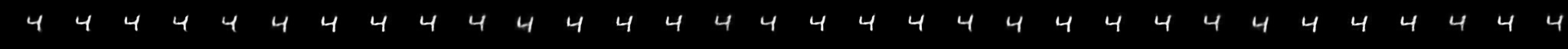} \\
    dis-VAE & \boximgc{}{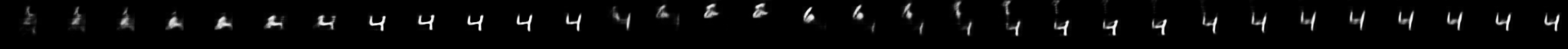} \\
    SVG & \boximgc{}{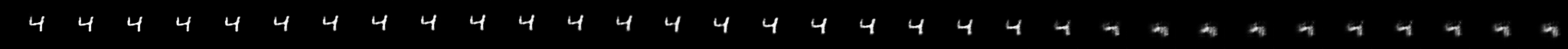} \\
    MTC-VAE & \boximgc{}{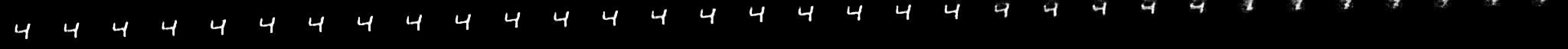} \\
    $\lambda=0$ & \boximgc{}{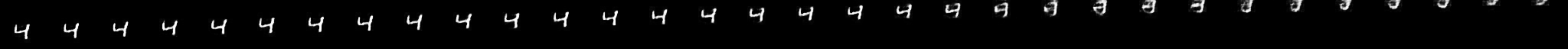} \\
    $O=1$ & \boximgc{}{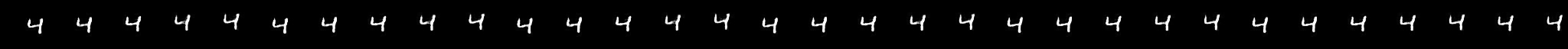} \\
    $O=3$ & \boximgc{}{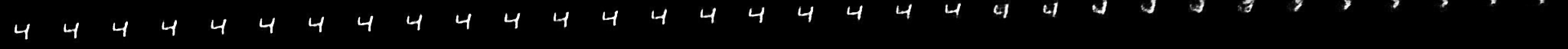} \\
    $O=4$ & \boximgc{}{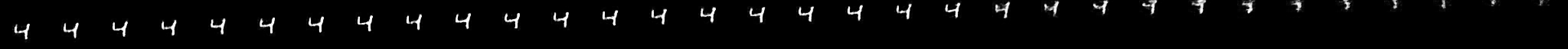} \\
    $c=1$ & \boximgc{}{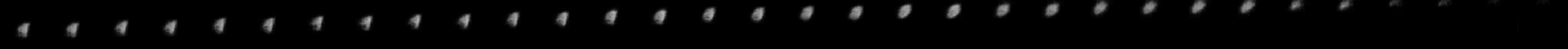} \\
    $c=3$ & \boximgc{}{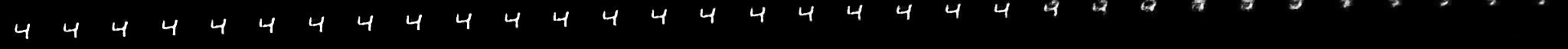} \\
    $c=7$ & \boximgc{}{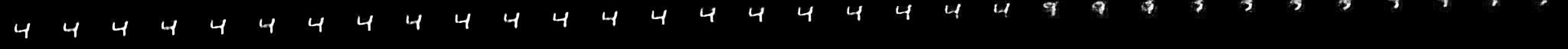} \\
    $c=9$ & \boximgc{}{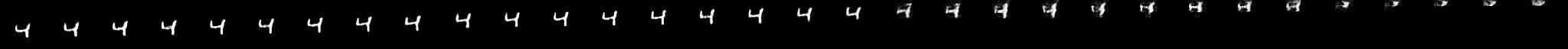}
  \end{tabular}
   \caption{MMNIST: examples of reenactment for the soft generalization scenario. Comparison with the baselines ($\beta$-TCVAE and dis-VAE), and ablation study on the chunk size ($c$), Blind Reenactment Loss ($\lambda$), and order of the model ($O$).}
   \label{fig:MMNIST_partial}
\end{figure*}

\printbibliography

\begin{IEEEbiography}[{\includegraphics[width=1in,height=1.1in,clip,keepaspectratio]{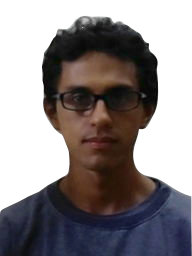}}]{Juan F. Hern\'{a}ndez Albarrac\'{i}n}
is a Ph.D. candidate at University of Campinas, Brazil. He has a bachelor degree in Computer Engineering from National University of Colombia (2014), and a M.Sc. degree in Computer Science from University of Campinas (2017). He has experience in Machine Learning and Computer Vision, focusing on evolutionary computing, deep learning, and generative models applied for image/video classification and synthesis.
\end{IEEEbiography}

\begin{IEEEbiography}[{\includegraphics[width=1in,height=1.1in,clip,keepaspectratio]{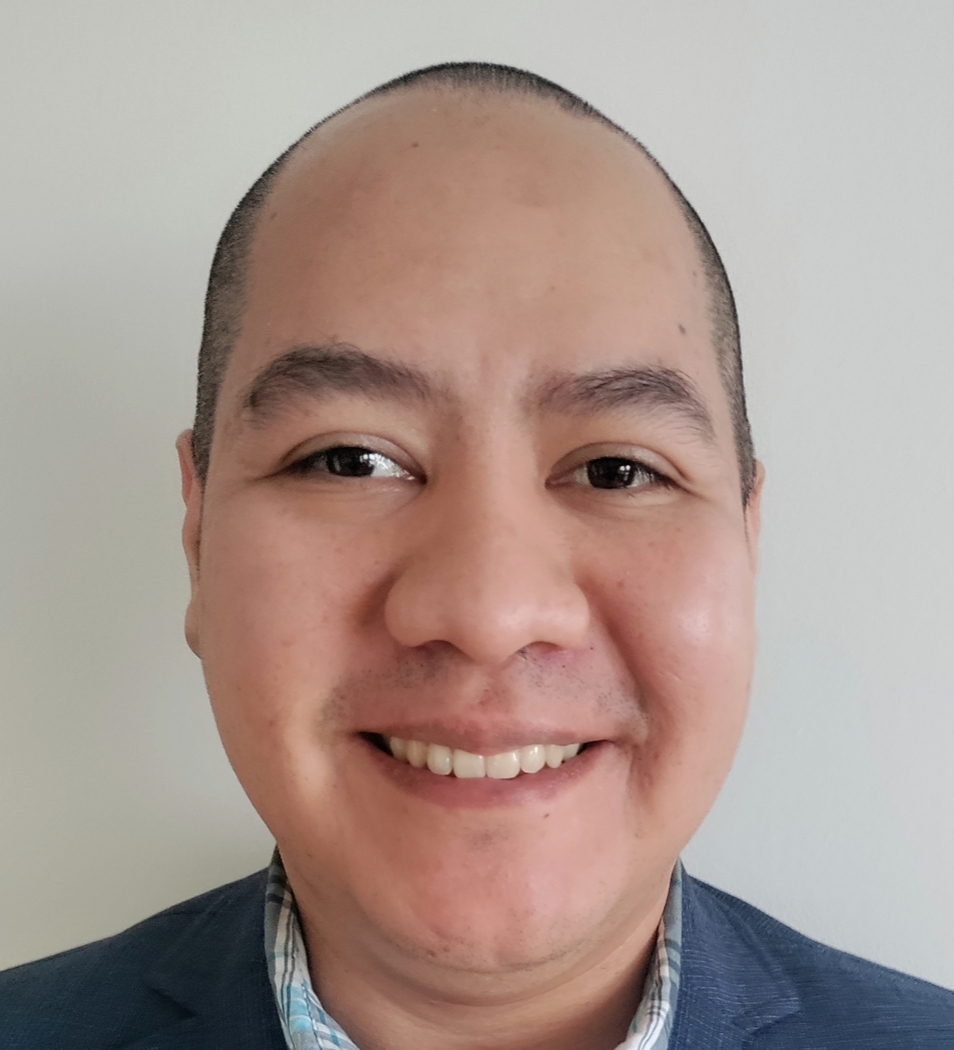}}]{Ad\'{i}n Ram\'{i}rez Rivera} (S'12, M'14, SM'21) received his B.Eng.\ degree in Computer Engineering from Universidad de San Carlos de Guatemala (USAC), Guatemala in 2009.  He completed his M.Sc.\ and Ph.D.\ degrees in Computer Engineering from Kyung Hee University, South Korea in 2013.  He is currently an Associate Professor at the Department of Informatics, University of Oslo, Norway.  His research interests are video understanding (including video classification, semantic segmentation, spatiotemporal feature modeling, and generation), and understanding and creating complex feature spaces.
\end{IEEEbiography}

\vfill

\end{document}